\documentclass{scrartcl}
\usepackage{natbib}
\usepackage{enumitem}
\usepackage{mathtools}

\usepackage{tikz}
\usetikzlibrary{external}
\usetikzlibrary{decorations.pathreplacing}
\usetikzlibrary{patterns}
\usetikzlibrary{fadings}
\usetikzlibrary{shapes.geometric}
\usetikzlibrary{angles,quotes}
\usetikzlibrary{math}
\usetikzlibrary{arrows}
\usetikzlibrary{shapes.multipart}
\usepackage{subcaption}
\usepackage{amsmath, tabularx}
\usepackage{amsfonts}
\usepackage{amssymb}
\usepackage{dsfont}
\usepackage[T1]{fontenc}
\usepackage[latin1]{inputenc}
\usepackage{epsfig}
\usepackage{graphicx}

\usepackage[boxed, lined]{algorithm2e}
\usepackage{float}
\usepackage{comment}

\newcommand{\bNULL}{\mathbf{0}}

\newcommand{\E}{\mathbf{E}}

\newcommand{\bk}{\mathbf{k}}

\newcommand{\bi}{\mathbf{i}}
\newcommand{\ba}{\mathbf{a}}

\newcommand{\bh}{\mathbf{h}}

\newcommand{\bx}{\mathbf{x}}
\newcommand{\bX}{\mathbf{X}}

\newcommand{\bw}{\mathbf{w}}
\newcommand{\bu}{\mathbf{u}}
\newcommand{\bv}{\mathbf{v}}
\newcommand{\bz}{\mathbf{z}}
\newcommand{\bM}{\mathbf{M}}

\newcommand{\D}{{\mathcal{D}}}
\newcommand{\A}{{\mathcal{A}}}

\newcommand{\F}{{\mathcal{F}}}
\newcommand{\G}{{\mathcal{G}}}

\newcommand{\Nu}{{\mathcal{N}}}

\newcommand{\N}{\mathbb{N}}

\newcommand{\R}{\mathbb{R}}
\newcommand{\Z}{\mathbb{Z}}
\newcommand{\Rd}{\mathbb{R}^d}
\usepackage{bbm}
\newcommand{\IND}{\mathbbm{1}}
\usepackage{bm}
\newcommand{\balpha}{\pmb{\alpha}}
\newcommand{\btheta}{\pmb{\theta}}

\newcommand{\beq}{\begin{eqnarray*}}

\newcommand{\eeq}{\end{eqnarray*}}

\newcommand{\beqm}{\begin{eqnarray}}

\newcommand{\eeqm}{\end{eqnarray}}

\newtheorem{theorem}{Theorem}

\newtheorem{lemma}{Lemma}
\newtheorem{definition}{Definition}

\newtheorem{remark}{Remark}

\let\oldremark\remark
\renewcommand{\remark}{\oldremark\normalfont}

\DeclareMathOperator*{\argmin}{arg\,min}
\DeclareMathOperator{\sgn}{sgn}
\DeclareMathOperator{\VC}{VCdim}
\newcommand{\EXP}{{\mathbf E}}
\newcommand{\PROB}{{\mathbf P}}

\renewcommand{\bf}{\normalfont \bfseries}
\renewcommand{\it}{\normalfont \itshape}
\usepackage{lmodern}
\allowdisplaybreaks
\usepackage[colorlinks=false, pdfborder={0 0 0}]{hyperref}
\hypersetup{
	colorlinks   = black, 
	urlcolor     = black,
	linkcolor    = black, 
	citecolor   = black
}
\begin{document}
	\newcounter{const}\setcounter{const}{0}
	\newcommand{\nconst}{\stepcounter{const}c_{\arabic{const}}}
	\newcommand{\const}{c_{\arabic{const}}}
	\newcommand{\mconst}{\addtocounter{const}{-1}c_{\arabic{const}}\stepcounter{const}}
	\newcommand{\mmconst}[1]{\addtocounter{const}{-#1}c_{\arabic{const}}\addtocounter{const}{#1}}
	\newcounter{ass}\setcounter{ass}{0}
	\newenvironment{assumption}[1][]{\refstepcounter{ass}\par\medskip\noindent%
	\textbf{Assumption~\arabic{ass}.} \rmfamily\itshape}{\medskip}
\newcounter{prop}\setcounter{prop}{0}
\newenvironment{property}[1][]{\refstepcounter{prop}\par\medskip\noindent%
	\textbf{(P\arabic{prop})} \rmfamily\itshape}{\medskip}
	\newcounter{rem}\setcounter{rem}{0}
	\newenvironment{remark-customized}[1][]{\refstepcounter{rem}\par\medskip\noindent%
	\textbf{Remark~\arabic{rem}.} \rmfamily}{\medskip}
	\allowdisplaybreaks
\renewcommand{\thefootnote}{\fnsymbol{footnote}}
\begin{center}
	
	{\LARGE \bf
		Analysis of convolutional neural network image classifiers in a rotationally symmetric model}
\footnote{
	Running title: {\it Rotationally symmetric image classification}}
\vspace{0.5cm}

	Michael Kohler\footnote{\label{note1}Funded by the Deutsche Forschungsgemeinschaft (DFG, German
	Research Foundation)  - Projektnummer 449102119.}
and Benjamin Walter$^{\ref{note1},}$\footnote{Corresponding author. Tel: +49-6 151-16-23386}
	\\

	{\it 
		Fachbereich Mathematik, Technische Universit\"at Darmstadt,
		Schlossgartenstr. 7, 64289 Darmstadt, Germany,
		email: kohler@mathematik.tu-darmstadt.de, 
		bwalter@mathematik.tu-darmstadt.de}

\end{center}
\vspace{0.5cm}

\begin{center}
	\today
\end{center}
\vspace{0.5cm}

\noindent
{\bf Abstract}\\
Convolutional neural network image classifiers are defined and the rate of convergence of the misclassification risk of the estimates towards the optimal misclassification risk is analyzed. 
Here we consider images as random variables with values in some functional space, where we only observe discrete samples as function values on some finite grid. %
 Under suitable structural and smoothness assumptions on the functional a posteriori probability, which includes some kind of symmetry against rotation of subparts of the input image, it is shown that least squares plug-in classifiers based on convolutional neural networks are able to circumvent the curse of dimensionality in binary image classification
 if we neglect a resolution-dependent error term. 
 The finite sample size behavior of the classifier is analyzed by applying it to simulated and real data.
\vspace*{0.2cm}

\noindent{\it AMS classification:} Primary 62G05; secondary 62G20.

\vspace*{0.2cm}

\noindent{\it Key words and phrases:}
Curse of dimensionality,
convolutional neural networks,
image classification,
rate of convergence.
\section{Introduction}
\label{se1}
In image classification, the task is to assign a given image to a class, where the class of the image depends on what kind of objects are represented on the image. For several years, the most successful methods in real-world applications are based on convolutional neural networks (CNNs), cf., e.g.,  \cite{He2016}, \cite{Goodfellow2016}, and \cite{Rawat2017}. 
For some image classification problems, it does not matter for the correct classification whether objects are rotated by arbitrary angles. This is the case, for example, in visual medical diagnosis applications, see, \cite{Veeling2018}, or in galaxy morphology prediction, see, \cite{Dieleman2015}, and further applications, see, e.g., \cite{Delchevalerie2021} and the literature cited therein.
A large number of papers demonstrate the empirical success of increasing complex network architectures, especially for image classification tasks with rotated objects, many architectures try to exploit this symmetry, e.g. by some kind of invariance to rotation, see, e.g., \cite{Delchevalerie2021}, \cite{Dieleman2015}, and \cite{Cohen2016}. However, a theoretical justification for this empirical success exists only partially, see, \cite{Rawat2017}. 
The aim of this article is, on the one hand, to introduce a statistical setting for image classification that includes the irrelevance of rotation of objects by arbitrary angles, and, on the other hand, to derive in this setting a rate of convergence of image classifiers based on CNNs, which is independent of the dimension of the input image.

\subsection{Image classification}
\label{se1.1}
In order to introduce the statistical setting for image classification, we describe idealized (random) images as $[0,1]$-valued functions on the cube
\[
C_{h}=\left[-\frac{h}{2},\frac{h}{2}\right]\times\left[-\frac{h}{2},\frac{h}{2}\right]\subset\R^2
\]
for  $h>0$.
The function value at position $(i,j)\in C_{h}$ describes the corresponding gray scale value and the width $h$ define the size of the image area. 
To obtain a suitable measurable space on these functions, we assume that they are continuous and denote
\[
[0,1]^{K}\coloneqq\{f:K\rightarrow[0,1]~:~f\text{ is a continuous function}\}
\] 
for all compact subsets $K\subset\R^2$. 
We can motivate the constraint that the function $f$ is continuous as follows: 
Both humans, due to their limited vision (cf., e.g., \cite{Gimelfarb2018}), and computers observe only digital discretized images, 
and for any discrete image with an arbitrary resolution it is possible to construct a continuous image such that the corresponding function is continuous (in practical applications, bicubic or bilinear interpolation can be used for this purpose, see \cite{Gonzalez2018}).
Since the space of all real-valued continuous functions on $C_h$ equipped with the metric induced by the maximum norm $\|\cdot\|_{\infty}$ defines a metric space, we obtain a measurable space %
$([0,1]^{C_{h}},\mathcal B([0,1]^{C_{h}}))$
with the corresponding Borel $\sigma$-algebra. Next we introduce our statistical setting for image classification: 
Let $(\Phi,Y)$, $(\Phi_1,Y_1)$, $\dots$, $(\Phi_n,Y_n)$
be independently and identically distributed random variables with values in $[0,1]^{C_1}\times\{0,1\}$.
Here the (random) image $\Phi$ has the (random) class $Y\in\{0,1\}$. 
In practice, we can only observe discrete images consisting of a finite number of pixels. To obtain discrete observations from our idealized images, we evaluate them on a corresponding finite grid. To obtain a corresponding grid, we divide the cube $C_1$ into $\lambda^2$ equal sized cubes and choose the grid points as the centers of the small cubes. Formally, this means that we define the grid $G_{\lambda}\subset C_{1}$ with resolution ${\lambda}\in\N$ by
\begin{equation}
	\label{se1eq1}
	G_{\lambda}=\left\{\left(\frac{i-\frac{1}{2}}{\lambda}-\frac{1}{2},\frac{j-\frac{1}{2}}{\lambda}-\frac{1}{2}\right)~:~i,j\in\{1,\dots,\lambda\}\right\}.
\end{equation}
The corresponding (continuous) function $g_{\lambda}:[0,1]^{C_1}\rightarrow[0,1]^{G_{\lambda}}$, which evaluates a idealized continuous image on the grid $G_{\lambda}$, is defined by
\[
g_{\lambda}(\phi)=\left(\phi\left(
\bu
\right)\right)_{\bu\in G_{\lambda}}\quad\big(\phi\in[0,1]^{C_1}\big),
\]
where for $[0,1]^{G_{\lambda}}$ we use the notation
\[
A^{I}=\{(a_i)_{i\in I} : a_i\in A~(i\in I)\}
\]
for a nonempty and finite index set $I$ and some $A\subseteq\R$. 
Based on the observations
\[
\D_n=\{(g_{\lambda}(\Phi_1),Y_1),...,(g_{\lambda}(\Phi_n),Y_n)\},
\]
we aim to construct a classifier
$
f_n=f_n(\cdot,\D_n):[0,1]^{G_{\lambda}}\rightarrow\{0,1\}
$
such that its misclassification risk
$
\PROB\{f_n(g_{\lambda}(\Phi))\neq Y|\D_n\}
$
is as small as possible. 
The misclassification risk is minimized by the so-called Bayes classifier, which is defined as
\[
f^*(\bx)=
\begin{cases}
	1&,\text{if }\eta^{({\lambda})}(\bx)>\frac{1}{2}\\
	0&,\text{ elsewhere},
\end{cases}
\]
where $\eta^{(\lambda)}$ is the a posteriori probability of class 1 for discrete images of resolution ${\lambda}$ given by
\begin{equation}
	\label{se1eq2}
	\eta^{({\lambda})}(\bx)=\PROB\{Y=1|g_{\lambda}(\Phi)=\bx\}\quad\left(\bx\in[0,1]^{G_{\lambda}}\right).
\end{equation}
Thus we have
\[
\min_{f:[0,1]^{G_{\lambda}}\rightarrow[0,1]}\PROB\{f(g_{\lambda}(\Phi))\neq Y\}=\PROB\{f^*(g_{\lambda}(\Phi))\neq Y\}
\]
(cf., e.g., Theorem 2.1 in \cite{Devroye1996}).
Since the a posteriori probability \eqref{se1eq2} is unknown in general 
we evaluate the statistical performance of our classifier $f_n$ by deriving an upper bound on
the expected misclassification risk of our classifier and the optimal misclassification risk, i.e. we want to derive an upper bound on
\begin{equation}
	\label{se1eq3}
	\begin{split}
		&\EXP\left\{\PROB\{f_n(g_{\lambda}(\Phi))\neq Y|\D_n\}-\min_{f:[0,1]^{G_{\lambda}}\rightarrow[0,1]}\PROB\{f(g_{\lambda}(\Phi))\neq Y\}\right\}\\
		&=\PROB\{f_n(g_{\lambda}(\Phi))\neq Y\}-\PROB\{f^*(g_{\lambda}(\Phi))\neq Y\}.
	\end{split}
\end{equation}
Here we use so-called plug-in classifiers, which are defined by
\[
f_n(\bx)=
\begin{cases}
	1&,\text{if }\eta_n(\bx)\geq\frac{1}{2}\\
	0&,\text{ elsewhere},
\end{cases}
\]
where 
$\eta_n(\cdot)=\eta_n(\cdot,\D_n):[0,1]^{G_{\lambda}}\rightarrow\R$
is an estimate of the a posteriori porbability \eqref{se1eq2}. 
To derive nontrivial of convergence for \eqref{se1eq3}, it is necessary to restrict the class of distributions of $(g_{\lambda}(\Phi),Y)$ (cf., \cite{Cover1968} and \cite{Devroye1982}). For this purpose, in
\cite{KoKrWa2020} they have introduced the hierarchical max-pooling model for the a posteriori probability of class 1 for discrete images \eqref{se1eq2} (see Definition \ref{de1} below), where they define a (random) image directly as a $[0,1]^{\{1,\dots,d_1\}\times\{1,\dots,d_2\}}$-valued random variable for some image dimensions $d_1,d_2\in\N$. In the hierarchical max-pooling model, the following two main ideas are used: The first idea is that the class of an image is determined by whether the image contains an object that is contained in a subpart of the image. The approach is then to estimate for all subparts of the image whether they contain the corresponding object or not. The probability that the image contains the object is then assumed to be the maximum of the probabilities of all subparts (see Definition \ref{de1} a)). The second idea is that the probabilities for the individual subparts are composed hierarchically by combining decisions from smaller subparts (see Definition \ref{de1} b)).
\begin{definition}
	\label{de1}
	Let $d_1,d_2\in\N$ with $d_1,d_2>1$ and $m:[0,1]^{\{1, \dots, d_1\} \times \{1, \dots, d_2\}} \rightarrow \R$.
	
	\noindent
	{\bf a)}
	We say that
	$m$
	satisfies a {\bf max-pooling model with index set}
	\[
	I \subseteq \{0, \dots, d_1-1\} \times \{0, \dots, d_2-1\},
	\]
	if there exist a function $f:[0,1]^{(1,1)+I} \rightarrow \R$ such that
	\[
	m(\bx)=
	\max_{
		(i,j) \in \Z^2 \, : \,
		(i,j)+I \subseteq \{1, \dots, d_1\} \times \{1, \dots, d_2\}
	}
	f\left(
	x_{(i,j)+I}
	\right)
	\quad
	(x \in [0,1]^{\{1, \dots, d_1\} \times \{1,
		\dots, d_2\}}).
	\]
	
	\noindent
	{\bf b)}
	Let $I=\{0, \dots, 2^l-1\} \times \{0, \dots, 2^l-1\}$
	for some $l \in \N$.
	We say that
	\[
	f:[0,1]^{\{1, \dots, 2^l\} \times \{1, \dots, 2^l\}} \rightarrow \R
	\]
	satisfies a
	{\bf hierarchical model of level $l$},
	if there exist functions
	\[
	g_{k,s}: \R^4 \rightarrow [0,1]
	\quad (k=1, \dots, l, s=1, \dots, 4^{l-k} )
	\]
	such that we have
	\[
	f=f_{l,1}
	\]
	for some
	$f_{k,s} :[0,1]^{\{1, \dots, 2^k\} \times \{1, \dots, 2^k\}} \rightarrow \R$ recursively defined by
	\begin{eqnarray*}
		f_{k,s}(\bx)&=&g_{k,s} \big(
		f_{k-1,4 \cdot (s-1)+1}(x_{
			\{1, \dots, 2^{k-1}\} \times \{1, \dots, 2^{k-1}\}
		})
		, \\
		&&
		\hspace*{1cm}
		f_{k-1,4 \cdot (s-1)+2}(x_{
			\{2^{k-1}+1, \dots, 2^k\} \times \{1, \dots, 2^{k-1}\}
		}), \\
		&&
		\hspace*{1cm}
		f_{k-1,4 \cdot (s-1)+3}(x_{
			\{1, \dots, 2^{k-1}\} \times \{2^{k-1}+1, \dots, 2^k\}
		}), \\
		&&
		\hspace*{1cm}
		f_{k-1,4 \cdot s}(x_{
			\{2^{k-1}+1, \dots, 2^k\} \times \{2^{k-1}+1, \dots, 2^k\}
		})
		\big)
		\\
		&&
		\hspace*{6cm}
		\left(
		x \in
		[0,1]^{
			\{
			1, \dots, 2^k
			\}
			\times
			\{ 1, \dots, 2^k
			\}
		}
		\right)
	\end{eqnarray*}
	for $k=2, \dots, l, s=1, \dots,4^{l-k}$,
	and
	\[
	f_{1,s}(
	x_{1,1},x_{1,2},x_{2,1},x_{2,2}
	)= g_{1,s}(x_{1,1},x_{1,2},x_{2,1},x_{2,2})
	\quad
	( x_{1,1},x_{1,2},x_{2,1},x_{2,2} \in [0,1])
	\]
	for $s=1, \dots, 4^{l-1}$.
	
	\noindent
	{\bf c)}
	We say that
	$m: [0,1]^{\{1, \dots, d_1\} \times \{1, \dots, d_2\}} \rightarrow \R$
	satisfies a {\bf hierarchical max-pooling model of level $l$
	} (where $2^l \leq \min\{ d_1,d_2\}$),
	if $m$ satisfies a max-pooling model with index set
	\[
	I=\{0, \dots, 2^l-1\} \times \{0, \dots 2^l-1\}
	\]
	and the function
	$f:[0,1]^{ \{ 1, \dots, 2^l \} \times \{1, \dots, 2^l\}} \rightarrow \R$ in the definition of this
	max-pooling model satisfies a hierarchical model
	with level $l$.
\end{definition}
In addition to these structural assumptions on the a posteriori probability, \cite{KoKrWa2020} also assume that the functions $g_{k,s}$ of the hierarchical model are $(p,C)$-smooth (for the definition of $(p,C)$-smoothness, see Section \ref{se1.4}). A drawback of the hierarchical max-pooling model, which is also used in \cite{KoLa2020} and in a generalized form in \cite{Wa2021}, is that it does not include some kind of 
symmetry against rotation
of subparts of the input image.
\subsection{Main results}
In this article we introdue a new model for the functional a posteriori probability 
\begin{equation}
	\label{apostid}
	\eta(\phi)=\PROB\{Y=1|\Phi=\phi\}\quad\Big(\phi\in[0,1]^{C_1}\Big)
\end{equation}
for continuous images. This allows us to integrate into our model both the ideas of the hierarchical max-pooling model and 
an assumption concerning the irrelevance of rotation of objects.
Assuming the new model for the functional a posteriori probability \eqref{apostid}, we show that least-squares plug-in CNN image classifiers (with ReLU activation function) achieve a rate of convergence of the expected difference of the misclassification risk of the classifier and the optimal misclassification risk \eqref{se1eq3} of
\[
\sqrt{\log(\lambda)\cdot(\log n)^4\cdot n^{-\frac{2\cdot p}{2\cdot p+4}}+\epsilon_{\lambda}}
\]
(up to some constant factor), where $\epsilon_{\lambda}$ is an error term depending on the image resolution and $p$ is a smoothness parameter of the a posteriori probability.
For a suitably small error term $\epsilon_{\lambda}$ and an appropriate and sufficiently large choice of the image resolution $\lambda$, \eqref{se1eq3} converges with rate 
\[n^{-\frac{2\cdot p}{2\cdot p+4}}\]
(up to some logarithmic factor). Hence, in this case, 
our CNN image classifiers 
are able to circumvent the curse of dimensionality assuming the new model for the functional a posteriori probability \eqref{apostid}.
\subsection{Discussion of related results}
A statistical theory for image classification using CNNs (with ReLU activation function) is considered in \cite{KoKrWa2020}, \cite{Wa2021}, and \cite{KoLa2020}. \cite{KoKrWa2020} and \cite{Wa2021} study plug-in CNN image classifiers learned by minimizing the squares loss, assuming generalizations of the hierarchical max-pooling model (see Definition \ref{de1}) for the a posteriori probability of class 1. The model in \cite{KoKrWa2020} consists of several hierarchical max-pooling models and the model in \cite{Wa2021} generalizes the hierarchical max-pooling model in the sense that the relative distances of hierarchically combined subparts are variable. In \cite{KoLa2020}, the hierarchical max-pooling model from Definition \ref{de1} is considered, where the CNN image classifiers minimize the cross-entropy loss. All three papers achieve a rate of convergence that is independent of the input image dimension.
The statistical performance of CNNs for classification problems where the data is assumed to have a low-dimensional geometric structure is studied in \cite{Liu2021}. Here as well, a dimension reduction is achieved while residual convolutional neural network architectures are used, i.e., convolutional neural networks containing skip layer connections.
\cite{Lin2019} obtained generalization bounds for CNN architectures in a setting of multiclass classification.
Classification problems using standard deep feedforward neural networks were analyzed in \cite{Kim2018}, \cite{Bos2021} and \cite{Hu2020}. %

Much more theoretical results exist in the context of regression estimation. \cite{Oono2019} use a similar residual CNN network architecture as \cite{Liu2021} and obtain estimation error rates that are optimal in the minimax sense. While they show that application-preferred architectures (especially in image classification applications) perform as well as standard feedforward neural networks, they do not identify situations in which CNN architectures outperform standard feedforward neural networks.
For standard deep feedforward neural networks, rate of convergence results with dimension reduction could be shown under the assumption that the regression function is a hierarchical composition of functions of small input dimension (cf., \cite{KoKr17}, \cite{Bauer2019}, \cite{SchmidtHieber2020}, \cite{KoLa2021}, \cite{Suzuki2019} and \cite{Langer2021}). \cite{KoKrLa2019} have shown that in case where the regression function has a low local dimensionality, sparse neural network estimates achieve a dimension reduction.
\cite{Imaizumi2019} obained generalization error rates for the estimation of regression functions with partitions having rather general smooth boundaries by neural networks.

Approximation results for CNNs were obtained by \cite{Zhou2020}, \cite{Petersen2020} and \cite{Yarotsky2018}. 
That the gradient descent finds the global minimum of the empirical risk with squares loss is shown for CNN architectures, e.g., in \cite{Du2018}. The networks used here are overparameterized. In \cite{KoKr2021}, it was shown that overparametrized deep neural networks minimizing the empirical $L_2$ risk do not, in general, generalize well.
\subsection{Notation}
\label{se1.4}
Throughout the paper, the following notation is used:
The sets of natural numbers, natural numbers including zero,
integers
and real numbers
are denoted by $\N$, $\N_0$, $\Z$ and $\R$, respectively.
For $\bx=(x_1,\dots,x_d)\in\R^d$ we denote the maximum norm by
\[
\|\bx\|_{\infty}=\max(|x_1|,\dots,|x_d|),
\]
and for $f:\R^d \rightarrow \R$
\[
\|f\|_\infty = \sup_{\bx \in \R^d} |f(\bx)|
\]
is its supremum norm, and the supremum norm of $f$
on a set $A \subseteq \R^d$ is denoted by
\[
\|f\|_{A,\infty} = \sup_{\bx \in A} |f(\bx)|.
\]
Let $p=q+s$ for some $q \in \N_0$ and $0< s \leq 1$.
A function $f:\R^d \rightarrow \R$ is called
$(p,C)$-smooth, if for every $\balpha=(\alpha_1, \dots, \alpha_d) \in
\N_0^d$
with $\sum_{j=1}^d \alpha_j = q$ the partial derivative
$\frac{
	\partial^q f
}{
	\partial x_1^{\alpha_1}
	\dots
	\partial x_d^{\alpha_d}
}$
exists and satisfies
\[
\left|
\frac{
	\partial^q f
}{
	\partial x_1^{\alpha_1}
	\dots
	\partial x_d^{\alpha_d}
}
(\bx)
-
\frac{
	\partial^q f
}{
	\partial x_1^{\alpha_1}
	\dots
	\partial x_d^{\alpha_d}
}
(\bz)
\right|
\leq
C
\cdot
\| \bx-\bz \|^s
\]
for all $\bx,\bz \in \R^d$.%

Let $\F$ be a set of functions $f:\Rd \rightarrow \R$,
let $\bx_1, \dots, \bx_n \in \Rd$ and set $\bx_1^n=(\bx_1,\dots,\bx_n)$.
A finite collection $f_1, \dots, f_N:\Rd \rightarrow \R$
is called an $\varepsilon$-- cover of $\F$ on $\bx_1^n$
if for any $f \in \F$ there exists  $i \in \{1, \dots, N\}$
such that
\[
\frac{1}{n} \sum_{k=1}^n |f(\bx_k)-f_i(\bx_k)| < \varepsilon.
\]
The $\varepsilon$--covering number of $\F$ on $\bx_1^n$
is the  size $N$ of the smallest $\varepsilon$--cover
of $\F$ on $\bx_1^n$ and is denoted by $\Nu_1(\varepsilon,\F,\bx_1^n)$.

For $z \in \R$ and $\beta>0$ we define
$T_\beta z = \max\{-\beta, \min\{\beta,z\}\}$. If $f:\R^d \rightarrow
\R$
is a function and $\F$ is a set of such functions, then we set
\[
(T_{\beta} f)(\bx)=
T_{\beta} \left( f(\bx) \right)
\quad \mbox{and} \quad
T_{\beta} \mathcal{F}
=
\left\{
T_{\beta} f
\quad : \quad
f \in \mathcal{F}
\right\}.
\]
Let $I$ be a nonempty and finite index set. For $\bx\in\R^d$ we use the notation
$\bx_I=(x_i)_{i\in I}$
and for $M\subset\R^d$ we define
$\bx+M=\{\bx+\bz : \bz\in M\}$.
\subsection{Outline of the paper}
In Section \ref{se2} the new model for the functional a posteriori probability is introduced and the CNN image classifiers used in this paper are defined in Section \ref{se3}. The main result is presented in Section \ref{se4} and proven in Section \ref{se6}. In Section \ref{se5} the finite sample size behavior of our classifier is analyzed by applying it to simulated and real data. 
\section{A rotationally symmetric hierarchical max-pooling model for the functional a posteriori probability}
\label{se2}
\begin{figure}[h]
	\centering
	\includegraphics[width=.29\textwidth]{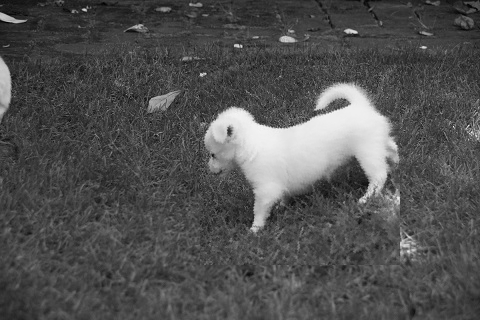}
	\includegraphics[width=.29\textwidth]{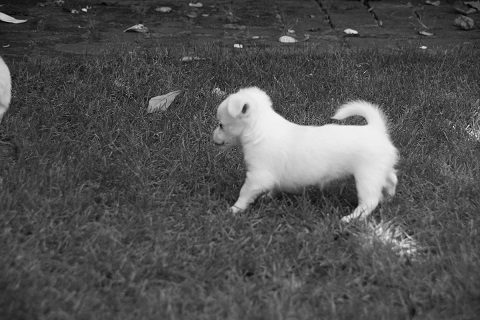}
	\includegraphics[width=.29\textwidth]{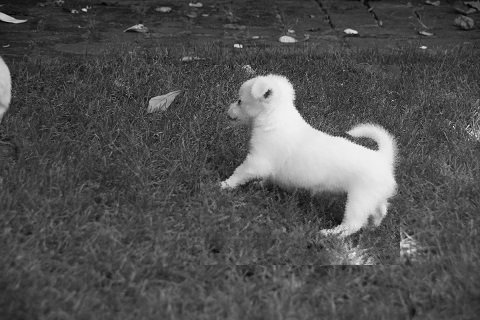}
	\caption{
		All three images are assigned to the class `dog'.
	}
	\label{fig1}
\end{figure}

We aim to extend the hierarchical max-pooling model from \cite{KoKrWa2020} (see Definition \ref{de1}) so that it becomes more realistic for practical applications of image classification. 
We do this by introducing a model for the functional a posteriori probability $\eta(\phi)=\PROB\{Y=1|\Phi=\phi\}$.
Here we are able to introduce some kind of symmetry against rotation of subparts of the input image. 
In order to rotate a subpart of an image, we define the function $rot^{(\alpha)}:\R^2\rightarrow\R^2$ given by
\[
rot^{(\alpha)}(\bx)=
\left(
\begin{matrix}
	\cos(\alpha) & -\sin(\alpha)\\
	\sin(\alpha) & \cos(\alpha)
\end{matrix}
\right)
\cdot\bx
\quad(\bx\in\R^2)
\]
which rotates its input through an angle $\alpha\in[0,2\pi]$ about the origin $\bNULL\in\R^2$. %
Furthermore, we define the translation function $\tau_{\bv}:\R^2\rightarrow\R^2$ with translation vector $\bv\in\R^2$ by
\[
\tau_{\bv}(\bx)=\bx+\bv\quad\big(\bx\in\R^2\big).
\]
Besides the ideas of the hierarchical max-pooling model from \cite{KoKrWa2020}, we want to integrate the following idea into our model:
We consider an image classification problem, where rotated objects correspond to each other, i.e., when asking whether an image contains a particular object, it does not matter for the correct classification whether the corresponding object is shown in some rotated position (cf., Figure \ref{fig1}).
We solve this problem by assuming that there is a function 
into which we can insert differently rotated subparts of an image (this function corresponds to the function $f:[0,1]^{C_h}\rightarrow[0,1]$ in part a) of the definition below). 
For a given subpart, the function estimates the probability whether the subpart contains a specific object. 
We then estimate the probability whether a subpart contains the object rotated by an arbitrary angle as follows: We rotate the subpart through different angles and estimate for each angle by the above function whether the subpart contains the object. %
The probability that the subpart contains the object rotated by an arbitrary angle is then assumed to be the maximum of the estimated probabilities for the various rotated subparts.

\begin{figure}[h]
\centering
\includegraphics{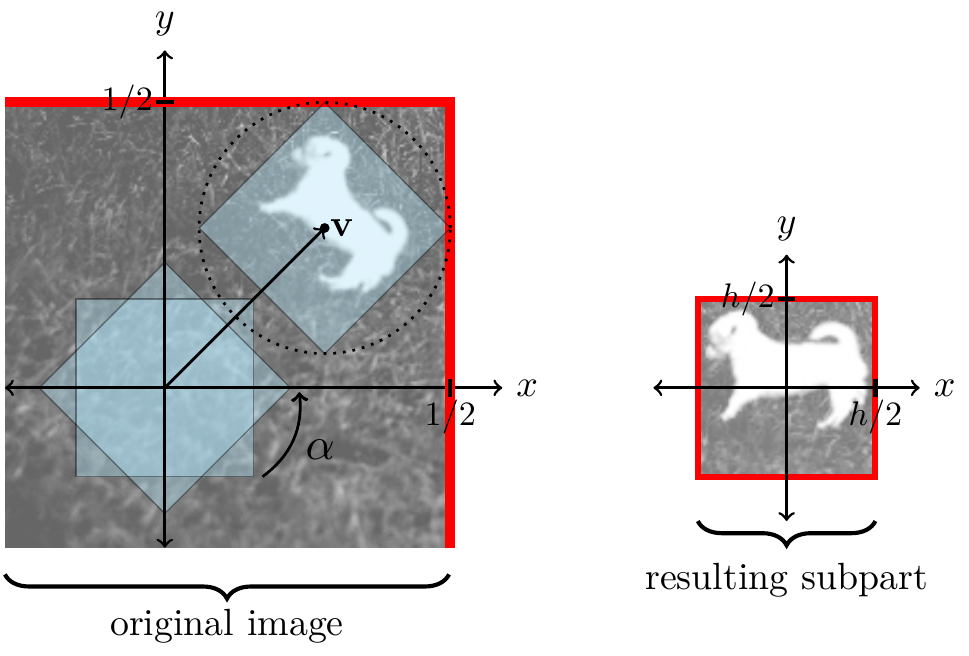}
	\caption{Illustration of an image $\phi$ and a subpart of the image, which is given by $\phi\circ \tau_{\bv}\circ rot^{(\alpha)}\big|_{C_h}$ as used in Definition \ref{de2} a).}
	\label{fig2}
\end{figure}

In the following definition we consider subparts of images $\phi\in[0,1]^{C_1}$. The subparts will have the form of possibly rotated cubes $C_h$ of side length $h>0$, which are subsets of $C_1$.
A subpart of the image $\phi\in[0,1]^{C_1}$ with side length $h$ rotated by an angle $\alpha\in\R$ and located at position $\bv$ is given by the function %
\[
\phi\circ \tau_{\bv}\circ rot^{(\alpha)}\big|_{C_h}\in[0,1]^{C_h},
\]
where we require $h\leq1/\sqrt{2}$ and $\bv\in[-1/2+h/\sqrt{2},1/2-h/\sqrt{2}]^2$ to ensure that the function $\tau_{\bv}\circ rot^{(\alpha)}\big|_{C_{h}}$ maps into the image area $C_1$ for all angles $\alpha\in[0,2\pi]$ (for an illustration see Figure \ref{fig2}).
A non-rotated subpart with side length $0<h'\leq h$ of an image $\phi\in[0,1]^{C_h}$ is then given by $\phi\circ \tau_{\bv}\big|_{C_{h'}}\in[0,1]^{C_{h'}}$
for some $\bv\in\R^2$ with $\bv+C_{h'}\subseteq C_h$.
\begin{definition}
	\label{de2}
	Let $m:[0,1]^{C_1}\rightarrow[0,1]$.

\noindent
{\bf a)}
Let $0<h\leq 1/\sqrt{2}$ and let
\begin{equation}
	\label{def1eq1}
	{h}/{\sqrt{2}}\leq b\leq{1}/{2}.
\end{equation} 
We say that $m$ satisfies a {\bf rotationally symmetric max-pooling model of width $h$ and border distance $b$}, if there exist a function $f:[0,1]^{C_{h}}\rightarrow[0,1]$ such that
\[
m(\phi)=\sup_{\bv\in\left[-(\frac{1}{2}-b),\frac{1}{2}-b\right]^2}\sup_{\substack{\alpha\in[0,2\pi]}}
f\left(\phi\circ \tau_{\bv}\circ rot^{(\alpha)}\big|_{C_{h}}\right)
\quad
\big(\phi\in [0,1]^{C_1}\big).
\]

\noindent
{\bf b)}
Let $l\in\N$ and $h>0$ and define $h_k=h/2^{l-k}$ for $k\in\Z$.
We say that
$f:[0,1]^{C_h}\rightarrow[0,1]$
satisfies a {\bf hierarchical model of level $l$}, if there exist functions
\[
g_{k,s}:\R^4\rightarrow[0,1]\quad(k=1,\dots,l,s=1,\dots,4^{l-k})
\]
and functions 
\begin{equation}
	\label{f0s}
	f_{0,s}:[0,1]^{C_{h_0}}\rightarrow[0,1]\quad(s=1,\dots,4^{l})
\end{equation}
such that we have
\[
f=f_{l,1}
\]
for some 
$
f_{k,s}:[0,1]^{C_{h_k}}\rightarrow\R
$
recursively defined by
\begin{align*}
	&f_{k,s}(\phi)
	=g_{k,s}\Big(
	f_{k-1,4\cdot(s-1)+1}\big(
	\phi\circ \tau_{(-h_{k-2},-h_{k-2})}\big|_{C_{h_{k-1}}}\big),
	\\
	&\hspace{1.5cm}
	f_{k-1,4\cdot(s-1)+2}\big(
	\phi\circ \tau_{(h_{k-2},-h_{k-2})}\big|_{C_{h_{k-1}}}\big),
	\\
	&\hspace{1.5cm}
	f_{k-1,4\cdot(s-1)+3}\big(
	\phi\circ \tau_{(-h_{k-2},h_{k-2})}\big|_{C_{h_{k-1}}}\big),
	\\
	&\hspace{1.5cm}
	f_{k-1,4\cdot s}\big(
	\phi\circ \tau_{(h_{k-2},h_{k-2})}\big|_{C_{h_{k-1}}}
	\big)
	\Big)
	\\
	&
	\hspace{7cm}\big(\phi\in[0,1]^{C_{h_k}}\big)
\end{align*}
for $k=1,\dots,l$ and $s=1,\dots,4^{l-k}$.

\noindent
{\bf c)} We say that $m$ satisfies a {\bf rotationally symmetric hierarchical max-pooling model of level $l$, width $h$ and border distance $b$}, if $m$ satisfies a rotationally symmetric max-pooling model with width $h$ and border distance $b$,
and the function $f:[0,1]^{C_h}\rightarrow[0,1]$ in the definition of this rotationally symmetric max-pooling model satisfies a hierarchical model of level $l$.

\noindent
{\bf d)}
Let $p=q+s$ for some $q \in \N_0$ and $s \in (0,1]$, and let $C>0$.
We say that a  hierarchical model is $(p,C)$--smooth
if all functions $g_{k,s}$ in its definition are $(p,C)$--smooth.
\end{definition}
\begin{remark-customized}
Condition \eqref{def1eq1} for the border distance ensures that the considered subparts do not extend beyond the border of the image area and that the set of centers $\bv$ of the subparts is not empty.%
\end{remark-customized}
\section{Convolutional neural network image classifiers}
\label{se3}
In this section, we define the CNN architecture that we will use in this paper. Our network architecture consists of $t\in\N$ convolutional neural networks computed in parallel,
followed by a fully connected standard feedforward neural network. %
Each of the $t$ convolutional neural networks consists of $L\in\N$ convolutional layers, a linear layer and a global max-pooling layer.
As activation function we use the ReLU function $\sigma:\R\rightarrow\R$, which is given by $\sigma(x)=\max\{x,0\}$.

\begin{figure}[h]
	\centering
\includegraphics{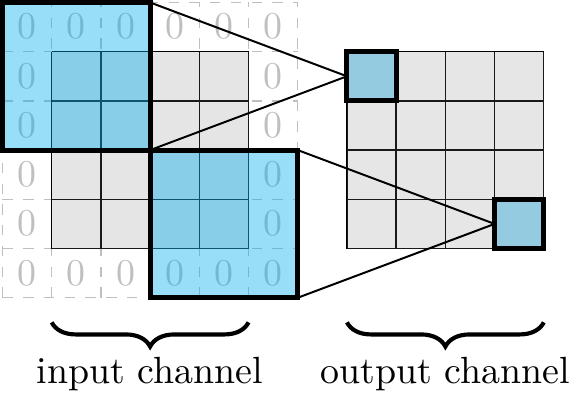}
	\caption{Example of symmetric zero padding for $M_r=3$ and $\lambda=4$.}
	\label{fig4}
\end{figure}

In the $r$-th convolutional layer we have $k_r\in\N$ channels and use filters of size $M_r\in\N$, where the global max-pooling layer computes the output of the convolutional neural network by a linear layer and by the computation of the maximum over (almost) all neurons of the output of the linear layer (the set of neurons whose maximum is computed depends on an output bound $B\in\N_0$). %
Our convolutional neural network architecture depends on a weight vector (so-called filters)
\[
\bw
=
\left(
w_{i,j,s_1,s_2}^{(r)}
\right)_{
	1 \leq i,j \leq M_r, s_1 \in \{1, \dots, k_{r-1}\}, s_2 \in \{1, \dots, k_r\},
	r \in \{1, \dots,L \}
},
\]
bias weights
\[
\bw_{bias}
=
\left(
w_{s_2}^{(r)}
\right)_{
	s_2 \in \{1, \dots, k_r\},
	r \in \{1, \dots,L\}
},
\]
and output weights
\[
\bw_{out}=\big(w_{s}\big)_{s\in\{1,\dots,k_L\}}.
\]
The output of the convolutional neural network is given by a real-valued function on $[0,1]^{G_{\lambda}}$ of the form
\begin{equation}
	\label{cnn2}
	\begin{split}
		f^{(B)}_{\bw, \bw_{bias},\bw_{out}}(\bx)
		&=
		\max \Bigg\{
		\sum_{s_2=1}^{k_L}
		w_{s_2}\cdot
		o_{(i,j),s_2}^{(L)} \,
		: \,
		(i,j)\in\{1+B,\dots,\lambda-B\}^2
		\Bigg\},
	\end{split}
\end{equation}
which depends on some output bound $B\in\{0,\dots,\lfloor(\lambda-1)/2\rfloor\}$, and where $o_{(i,j),s_2}^{(L)}$ is
the output of the last convolutional layer, which is
recursively defined
as follows:

We start with
\[
o_{(i,j),1 }^{(0)} = x_{\left(\frac{i-1/2}{\lambda}-\frac{1}{2},\frac{j-1/2}{\lambda}-\frac{1}{2}\right)}
\quad \mbox{for }
(i,j)\in \{1, \dots, \lambda\}^2
\]
and define recursively
\begin{equation}
	\label{conv1}
	o_{(i,j),s_2}^{(r)}
	=
	\sigma \Bigg(
	\sum_{s_1=1}^{k_{r-1}}
	\sum_{\substack{t_1,t_2 \in \{1, \dots, M_r\}\\i+t_1-\lceil M_r/2\rceil\in\{1,\dots,\lambda\}\\j+t_2-\lceil M_r/2\rceil\in\{1,\dots,\lambda\}}}
	w_{t_1,t_2,s_1,s_2}^{(r)}
	\cdot
	o_{(i+t_1-\lceil M_r/2\rceil,j+t_2-\lceil M_r/2\rceil),s_1}^{(r-1)}
	+
	w_{s_2}^{(r)}
	\Bigg)
\end{equation}
for $(i,j)\in\{1,\dots,\lambda\}^2$, $s_2\in\{1,\dots,k_r\}$
and
$r \in \{1, \dots, L\}$. 
For $\bk=(k_1,\dots,k_L)$ and $\bM=(M_1,\dots,M_L)$ we introduce the function class
\[
\F_{L,\bk,\bM,B}^{CNN}=\left\{f: f~\text{ is of the form \eqref{cnn2}}\right\}.
\]
In definition \eqref{conv1} we use a so-called zero padding, which ensures that the size of a channel is the same as in the previous layer. 
For odd filter sizes $M_r$ we obtain a symmetric zero padding as illustrated in Figure \ref{fig4}.

A fully connected standard feedforward neural network $g_{net}:\R^t\rightarrow\R$ with ReLU activation function, $L_{net}\in\N_0$ hidden layers and $k_r$ neurons in the $r$-th layer $(r=1,\dots,L_{net})$ is defined by
\begin{equation}
	\label{FNN}
	g_{net}(\bx) = \sum_{i=1}^{k_{L_{net}}} w_{i}^{(L_{net})}g_i^{(L_{net})}(\bx) + w_{0}^{(L_{net})}
\end{equation}
for some output weights $w_0^{(L_{net})},\dots,w_{k_{L_{net}}}^{(L_{net})}\in\R$,
where $g_i^{(L_{net})}$ is recursively defined by
\[ 
g_i^{(r)}(\bx) = \sigma\left(\sum_{j=1}^{k_{r-1}} w_{i,j}^{(r-1)} g_j^{(r-1)}(\bx) + w_{i,0}^{(r-1)} \right)
\]
for $w_{i,0}^{(r-1)},\dots,w^{(r-1)}_{i,k_{r-1}}\in\R$,
$i \in \{1,\dots,r_{net}\}$,
$r \in \{1, \dots, L_{net}\}$, $k_0=t$
and
\[
g_i^{(0)}(\bx) =x_i
\]
for $i=1,\dots,k_0$.
We define the class of fully connected standard feedforward neural networks with $L_{net}$ layers and $r_{net}\in\N$ neurons per layer by
\begin{equation}
	\label{FNNclass}
	\G_{t}(L_{net},r_{net})=\left\{g_{net}~:~g_{net}\text{ is of the form \eqref{FNN} with }k_1=\dots=k_{L_{net}}=r_{net}\right\}.
\end{equation}

Our overall convolutional neural network architecture is then defined by
\begin{equation*}
	\F_{\btheta}^{CNN}=\left\{f(\bx)=g_{net}(f_1(\bx),\dots,f_t(\bx)) : f_1,\dots,f_t\in\F_{L,\bk,\bM,B}^{CNN},~g_{net}\in\G_{t}(L_{net},r_{net})\right\}
\end{equation*}
for a parameter vector $\btheta=(t,L,\bk,\bM,B,L_{net},r_{net})$.

We define the least squares estimate of
$\eta^{(\lambda)}(\bx)=\EXP\{Y=1|g_{\lambda}(\Phi)=\bx\}$
by
\begin{equation}
	\eta_n = \argmin_{f \in \F_{\btheta}^{CNN}}
	\frac{1}{n} \sum_{i=1}^n |Y_i - f(g_{\lambda}(\Phi_i))|^2
	\label{minp}
\end{equation}
and define our classifier $f_n$ by
\[
f_n(\bx)=
\begin{cases}
	1, & \mbox{if } \eta_n(\bx) \geq \frac{1}{2} \\
	0, & \mbox{elsewhere}.
\end{cases}
\]
For simplicity, we assume that the minimum of the empirical $L_2$ risk \eqref{minp} exists. 
If this is not the case, our result also holds for an estimator whose empirical $L_2$ risk is close enough to the infimum. 
\section{Main result}
\label{se4}
In the sequel, let $\lambda\in\N$ be the resolution of the observed images defined as in Section 1.2, i.e., the discretized quadratic images consist of $\lambda^2$ pixels.
Futhermore, we assume that the functional a posteriori probability $\eta(\phi)=\PROB\{Y=1|\Phi=\phi\}$
satisfies a $(p,C)$-smooth rotationally symmetric hierarchical max-pooling model of level $l$ and width $h$. 
Before presenting the main result, we introduce two further assumptions on the a posteriori probability $\eta$.
In order to formulate these assumptions we need the following notation:
For a subset $A\subseteq\R^2$ let 
$1\big|_{A}:A\rightarrow\R$ denote the constant function with value one.
Let
$f_{0,s}:[0,1]^{C_{h_0}}\rightarrow[0,1]$ $(s=1,\dots,4^l)$ be the functions from the hierarchical model of $\eta$, where $h_0=h/2^l$.
We will use the assumptions below
to approximate a rotationally symmetric hierarchical max-pooling model by a convolutional neural network.
The first assumption is a smoothness assumption on the functions $f_{0,s}$ if we apply them to constant images.
\begin{assumption}
	\label{ass1}
	For all $s\in\{1,\dots,4^l\}$ there exist a $(p,C)$-smooth function $g_{0,s}:\R\rightarrow[0,1]$
	such that
	\[
	g_{0,s}(x)=f_{0,s}\left(x\cdot1\big|_{C_{h_0}}\right)
	\]
	holds for all $x\in[0,1]$.
\end{assumption}

\begin{figure}[h]
	\centering
\includegraphics{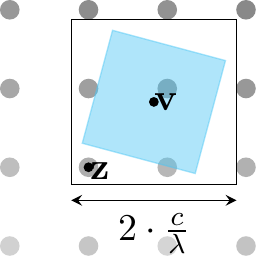}
	\caption{Illustration of a subpart with center $\bv$ and a point $\bz$ %
		as in Assumption \ref{ass2}, where we choosed $c=1.05$ and $h_0=(c\cdot\sqrt{2})/\lambda$. In the background one can see possible pixel values on the corresponding grid $G_{\lambda}\subset C_1$.}
	\label{fig5}
\end{figure}

\noindent
In the second assumption we bound the error that occurs if we replace the input of the function $f_{0,s}$, which is an possibly rotated subpart of an image $\phi\in[0,1]^{C_1}$ (cf., Definition \ref{de2}), by a constant image whose gray scale value is chosen from the local neighborhood of the corresponding subpart. The size of the subpart, as well as the size of the neighborhood of the subpart, depends on the resolution $\lambda$, as shown in Figure \ref{fig5}.
\begin{assumption}
	\label{ass2}
There exists a measurable $A\subset[0,1]^{C_1}$ with $P_{\Phi}(A)=1$, $\epsilon_{\lambda}\in[0,1]$ and a scaling factor $c>1$
with $h_0\leq\min\{(c\cdot\sqrt{2})/\lambda,1/\sqrt{2}\}$
such that for all $\phi\in A$,  $\bv\in[h_0/\sqrt{2}-1/2,1/2-h_0/\sqrt{2}]^2$, $\alpha\in[0,2\pi]$, and $s\in\{1,\dots,4^l\}$:
	\begin{equation*}
		\sup_{\substack{\bz\in C_1~:~\|\bv-\bz\|_{\infty}\leq\frac{c}{\lambda}}}\Bigg|f_{0,s}\Big(\underbrace{\phi\circ \tau_{\bv}\circ rot^{(\alpha)}\big|_{C_{h_0}}}_{\text{subpart of }\phi\text{ with center }\bv}\Big)-f_{0,s}\Big(\phi(\bz)\cdot1\big|_{C_{h_0}}\Big)\Bigg|\leq\epsilon_{\lambda}.
	\end{equation*}
\end{assumption}
\begin{remark-customized}
Note that $\phi\circ \tau_{\bv}\circ rot^{(\alpha)}\big|_{C_{h_0}}$ is a subpart of $\phi$ with center $\bv$ and width $h_0$ rotated by $\alpha$ as illustrated in Figure \ref{fig2}. So we apply $f_{0,s}$ to an arbitrary subpart of $\phi$ with center $\bv$ and let $\bz$ be choosen from the neighborhood of $\bv$. The condition $h_0\leq(c\cdot\sqrt{2})/\lambda$ ensures that the subpart of width $h_0$ is contained in the corresponding neighborhood. As illustrated in Figure \ref{fig5}, for a small scaling factor $c$, we consider subparts whose size approximately corresponds to the resolution. 
\end{remark-customized}

\noindent
To motivate that Assumption \ref{ass2} seems realistic for some small $\epsilon_{\lambda}\in[0,1]$, we consider the following example: %
Suppose that $A\subset[0,1]^{C_1}$ is defined as bilinear interpolations of all $\bx\in[0,1]^{G_{\lambda_{max}}}$ for some $\lambda_{max}\in\N$.
Furthermore, let us choose $\lambda$ much larger than $\lambda_{max}$. 
If we now consider for an arbitrary image coordinate a neighborhood whose width is upper bounded by $1/\lambda$, the gray scale values in this neighborhood differ only slightly. 
Therefore, we could replace a subpart contained in such a neighborhood with a corresponding constant image without changing the individual pixel values substantially.
\begin{theorem}
	\label{th1}
	Let $n\in\N\setminus\{1\}$ and $l\in\N$, choose $\lambda\in\N$ with

\noindent\begin{minipage}{0.25\textwidth}
	\begin{equation}
		\label{th1eq2}
		\lambda\geq2^l+2\cdot l-1,
	\end{equation} 
\end{minipage}%
\begin{minipage}{0.37\textwidth}\centering
	\begin{equation}
		~\text{let}~~0<h\leq\frac{2^l}{\sqrt{2}\cdot\lambda},
		\label{th1eq1}
	\end{equation}
\end{minipage}%
\begin{minipage}{0.37\textwidth}
	\begin{equation}
		\label{th1eq3}
		~\text{set}~~
		b=\frac{{2^l+2\cdot l-1}}{2\cdot\lambda},
	\end{equation}
\end{minipage}\vskip1em

\noindent
and let $p\in[1,\infty)$.
	Let 
	$(\Phi,Y)$, $(\Phi_1,Y_1)$, ...,$(\Phi_n,Y_n)$ 
	be independent and identically distributed $[0,1]^{C_1}\times\{0,1\}$-valued random variables. Assume that the functional a posteriori probability $\eta(\phi)=\PROB\{Y=1|\Phi=\phi\}$ satisfies a $(p,C)$-smooth rotationally symmetric hierarchical max-pooling model of level $l$, width $h$ and border distance $b$.
Furthermore, assume Assumption \ref{ass1} for $(p,C)$-smooth functions $\{g_{0,s}\}_{s=1,\dots,4^l}$ %
and Assumption \ref{ass2} for some $\epsilon_{\lambda}\in[0,1]$, some measurable $A\subset[0,1]^{C_1}$ and some scaling 
factor $c>1$.

Choose $L_{n}=\lceil \nconst\cdot n^{2/(2p+4)}\rceil$ for some sufficiently large constant $\const>0$, set
\[
L=\frac{4^{l+1}-1}{3}\cdot(L_{n}+1),\quad t=\left\lceil\frac{2^{l-1/2}\cdot\pi}{c-1
}\right\rceil,\quad B=2^{l-1}+l-1, \quad L_{net}=\lceil\log_2 t\rceil,
\]
$r_{net}=3\cdot t$
and
$k_r=5\cdot4^{l-1}+\nconst$
$(r=1,\dots,L)$ for $\const>0$ sufficiently large,  
and for $k=0,\dots,l$ set
\[
M_r=\IND_{\{k>1\}}\cdot2^{k-1}+3\quad\quad\Bigg(r={\sum_{i=0}^{k-1}4^{l-i}\cdot(L_{n}+1)}+1,\dots,\sum_{i=0}^{k}4^{l-i}\cdot(L_{n}+1)\Bigg),
\]
where we define the empty sum as zero.
Define $f_n$ as in Section \ref{se3}. Then we have
\begin{equation}
	\label{th1eq4}
	\begin{split}
		&\PROB\{f_n(g_{\lambda}(\Phi))\neq Y\}-\min_{f:[0,1]^{G_{\lambda}}\rightarrow[0,1]}\PROB\{f(g_{\lambda}(\Phi))\neq Y\}\\
		&\leq \nconst\cdot\sqrt{\log(\lambda)\cdot(\log n)^4\cdot n^{-\frac{2\cdot p}{2\cdot p+4}}+\epsilon_{\lambda}}
	\end{split}
\end{equation}
for some constant $\const>0$ which does not depend on $\lambda$ and $n$.
\end{theorem}
\begin{remark-customized}
	The constant $\const$ in \eqref{th1eq4} depends polynomially on $2^l$. Therefore the resolution $\lambda$ occurs logarithmically in \eqref{th1eq4} only
	in the case where $2^l\ll\lambda$, which leads to small widths $h$ (cf., equation \eqref{th1eq1}). 
	If we assume that there exists a sufficiently small resolution $\lambda_n$ such that further $\epsilon_{\lambda_n}\leq\nconst\cdot n^{-{2p}/{(p+4)}}$ for some constant $\const>0$, 
	we obtain a rate
	\[n^{-\frac{p}{2\cdot p+4}}\] 
	(up to some logarithmic factor) in Theorem \ref{th1}. Hence, under this assumption and an appropriate choice of $\lambda$, our CNN image classifier is able to circumvent the curse of dimensionality in case that the a posteriori probability satisfies a $(p,C)$-smooth rotationally symmetric hierarchical max-pooling model.
\end{remark-customized}

\begin{remark-customized}
	\label{re4}
	In our approximation result of Lemma \ref{le1}, we can choose the function $f_{CNN}\in\F_{\btheta}^{CNN}$ such that its $t$ CNNs, which are computed in parallel, share the same weights. More precisely, we can choose $f_{CNN}$ such that each filter of any layer corresponds to a rotated filter in the same layer in a CNN computed in parallel (the weights only have different positions within the filters). Therefore, with an appropriate restriction to our function class $\F_{\btheta}^{CNN}$ so that the weights of the $t$ CNNs are shared, one could improve the rate of convergence in Theorem \ref{th1} by a constant factor. In some image classification applications where rotated objects correspond to each other, such a constraint increases the performance, see, e.g., \cite{Marcos2016}, \cite{Dieleman2015}, \cite{Wu2015}, and \cite{Vives2017}. Our theoretical analysis therefore supports the use of such additional weight sharing, in addition to the weight sharing of the convolutional operation, and provides a theoretical indication of why such CNN architectures have better generalization properties.
\end{remark-customized}

\begin{figure}[h]
\centering
		\begin{minipage}{.35\textwidth}
			\centering
\includegraphics{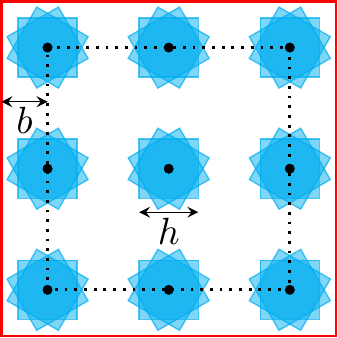}
		\end{minipage}
		\begin{minipage}{.35\textwidth}
			\centering
\includegraphics{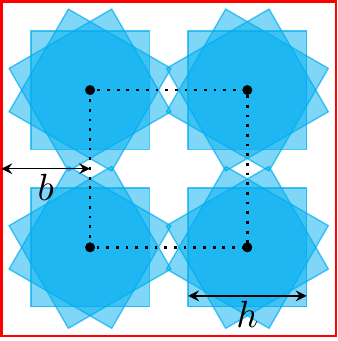}
		\end{minipage}
	\caption{The figure shows possible subparts of width $h$ for the rotationally symmetric hierarchical max-pooling model used in Theorem \ref{th1}.  On both sides we consider an example in which we have $\lambda=2^9$ and $h=2^l/(\sqrt{2}\cdot\lambda)$, where on the left hand side we have chosen $l=7$ and on the right hand side $l=8$.}
	\label{fig6}
\end{figure}

\begin{remark-customized}
Condition \eqref{th1eq2} ensures that the border distance $b$ defined as in \eqref{th1eq3} remains less than or equal to $1/2$ 
and that the width $h$ satisfies $h\leq1/\sqrt{2}$ (cf., equation \eqref{th1eq1}). Moreover, condition \eqref{th1eq2} ensures that $h/\sqrt{2}\leq b$.
In the case of maximum width $h=2^l/(\sqrt{2}\cdot\lambda)$ and 
for large $l$, we get close to the minimum border distance $h/\sqrt{2}$, since
\[
b=\frac{{2^l+2\cdot l-1}}{2\cdot\lambda}=\frac{h}{\sqrt{2}}\cdot\underbrace{\frac{2^{l}+2\cdot l-1}{2^{l}}}_{\approx1}.
\]
Condition \eqref{th1eq2} and choice \eqref{th1eq3} are therefore no real limitations on our model and we obtain, as we have shown in Figure \ref{fig6} for applications, reasonable border distances $b$ and widths $h$ of the subparts.
\end{remark-customized}
\begin{remark-customized}
	Some of the network parameters depend on the rotationally symmetric hierarchical max-pooling model. In applications, these network parameters can be chosen in a data-dependent way, e.g., by using the splitting of the sample technique as used in the next section.
\end{remark-customized}
\section{Application to simulated and real data}
\label{se5}
In this section, we study the finite sample size behavior of our CNN image classifier introduced in Section \ref{se3} by applying it to synthetic and real image data sets. Furthermore, we introduce three other CNN architectures that we can motivate from our theory and compare the performance of all four image classifiers. The three alternative CNN image classifiers are also defined as least-squares plug-in classifiers.

We denote the function class introduced in Section \ref{se3} by 
$
\F_1=\F_{\btheta}^{CNN}
$
for a parameter vector $\btheta=(t,L,\bk,\bM,B,L_{net},r_{net})$.
For the first alternative CNN architecture, we replace the fully connected feedforward neural network by simply computing the maximum over the outputs of the $t$ convolutional neural networks:
\begin{equation*}
	\F_2=\left\{f(\bx)=\max\{f_1(\bx),\dots,f_t(\bx)\} : f_1,\dots,f_t\in\F_{L,\bk,\bM,B}^{CNN}\right\}.
\end{equation*}
Following the proof of Theorem \ref{th1}, it is easy to see that the corresponding least squares plug-in image classifier over this function class, achieve the same rate of convergence as in Theorem \ref{th1}.
Our second alternative approach is inspired by the observation from Remark \ref{re4}. Here we follow, e.g., \cite{Dieleman2015} or \cite{Vives2017} by applying the same CNN to multiple rotated versions of the input image and then compute the overall output as the maximum of the individual outputs. We rotate the input image by $90^{\circ}$, $180^{\circ}$, and $270^{\circ}$, since multiples of $90^{\circ}$ rotations map the grid $G_{\lambda}$ onto itself. Because it does not matter whether we rotate the input feature maps of a convolutional layer and then inversely rotate the output feature maps, or whether we rotate the corresponding filters, this architecture corresponds in our case to an architecture that has shared rotated filters (for an illustration and a more detailed explanation, see \cite{Dieleman2016}).
The rotation function 
$rot_{90^{\circ}}:[0,1]^{G_{\lambda}}\rightarrow[0,1]^{G_{\lambda}}$
which rotates a discretized image with resolution $\lambda\in\N$ by $90^{\circ}$ is given by
\begin{equation*}
	\big(rot_{90^{\circ}}(\bx)\big)_{\left(\frac{i-1/2}{\lambda}-\frac{1}{2},\frac{j-1/2}{\lambda}-\frac{1}{2}\right)}
	=x_{\left(\frac{\lambda-j+1-1/2}{\lambda}-\frac{1}{2},\frac{i-1/2}{\lambda}-\frac{1}{2}\right)}\quad\left(\bx\in[0,1]^{G_{\lambda}}\right)
\end{equation*}
for all $i,j\in\{1,\dots,\lambda\}$ and our function class is defined by 
\begin{align*}
	&\F_3
	=\Big\{f(\bx)=\max\{g(\bx),g(rot_{90^{\circ}}(\bx)),\dots,g(\underbrace{rot_{90^{\circ}}\circ\dots\circ rot_{90^{\circ}}}_{3\text{ times}}(\bx))\} : g\in\F_2\Big\}.
\end{align*}
For our third alternative network architecture, we extend the idea from the function class $\F_3$ by first rotating an input image by all angles of the discretization 
\[
\{\alpha_1,\dots,\alpha_t\}=\left\{\frac{2\pi}{t}\cdot0,\frac{2\pi}{t}\cdot1,\dots,\frac{2\pi}{t}\cdot(t-1)\right\}
\]
of $[0,2\pi)$ for some $t\in\N$. The corresponding function class is defined by
\begin{align*}
	&\F_4
	=\Big\{f(\bx)=\max\{g(f_{rot}^{(\alpha_1)}(\bx)),g(f_{rot}^{(\alpha_2)}(\bx)),\dots,g(f_{rot}^{(\alpha_t)}(\bx))\} : g\in\F_{L,\bk,\bM,B}^{CNN}\Big\},
\end{align*}
where we use a nearest neighbor interpolation for the rotation function $f_{rot}^{(\alpha_i)}$, which we define and explain in detail in Section A.2 of the supplement.  

In our first application, we apply our CNN image classifiers to simulated synthetic image datasets. A synthetic image dataset consists of finitely many realizations \[\D_N=\{(\bx_1,y_1),\dots,(\bx_N,y_N)\}\]
of a $[0,1]^{G_{\lambda}}\times\{0,1\}$-valued random variable $(\bX,Y)$. Here, as in Section \ref{se1}, $\lambda\in\N$ denotes the resolution of the images and the value of $Y$ denotes the class of the image. In our first example, we use the values $\lambda=32$ and $\lambda=64$. The images of both classes contain three randomly rotated geometric objects each, where images of class 0 contain three squares. The images of class 1 also contain three squares, although at least one of the squares is missing exactly one quarter (see Figure 6). For a detailed explanation of the creation of the image data sets, see Section A.1 in the supplement.
\begin{figure}[h]
	\centering
	\includegraphics[width=.1\textwidth]{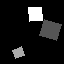}
	\includegraphics[width=.1\textwidth]{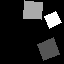}
	\includegraphics[width=.1\textwidth]{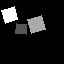}
	\includegraphics[width=.1\textwidth]{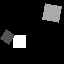}
	\includegraphics[width=.1\textwidth]{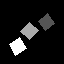}
	\includegraphics[width=.1\textwidth]{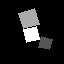}
	\includegraphics[width=.1\textwidth]{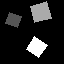}
	\includegraphics[width=.1\textwidth]{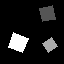}
	\\[\smallskipamount]
	\includegraphics[width=.1\textwidth]{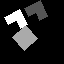}
	\includegraphics[width=.1\textwidth]{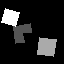}
	\includegraphics[width=.1\textwidth]{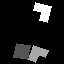}
	\includegraphics[width=.1\textwidth]{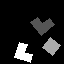}
	\includegraphics[width=.1\textwidth]{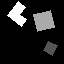}
	\includegraphics[width=.1\textwidth]{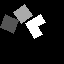}
	\includegraphics[width=.1\textwidth]{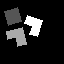}
	\includegraphics[width=.1\textwidth]{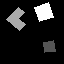}
	\caption{Some random images as realizations of the random variable $\bX$, where the first row show images of class 0 and the lower row show images of class 1.}
	\label{fig7}
\end{figure}

Since our image classifiers depend on parameters that influence their performance, we select them in a data-dependent manner by splitting our training data $\D_{n}$ into a learning set of size $n_l=\lfloor4/5\cdot n\rfloor$ and a validation set of size $n_v=n-n_l$. We then train our classifiers with different choices of parameter combinations on the learning set and choose the parameter combination that minimizes the empirical misclassification risk on the validation set. Finally, we train our classifier with the best parameter combination on the entire training set $\D_n$. 
For all four network architectures, we adaptively choose the parameters $l\in\{2,3\}$, $k\in\{2,4\}$ and $L_n\in\{1,2\}$, %
where the network parameters are then given by $L=L_{n}\cdot l$, $\bk=(k,\dots,k)$, $\bM=(M_1,\dots,M_L)$, $B=2^{l-1}-(l-1)$
with filter sizes $M_1,\dots,M_L$ defined by
\[
M_{(r-1)\cdot L_n+1},\dots,M_{r\cdot L_n}=\IND_{\{r>2\}}\cdot2^{r-2}+3\quad(r=1,\dots,l)
\]
(note that the choice of layers and filter sizes is a simplification contrary to the choice in Theorem \ref{th1}).
To make the comparison of the three network architectures fairer, i.e., to avoid that the network architectures $\F_3$ and $\F_4$ are able to learn more angles, we adaptively choose $t\in\{4,8\}$ for the function classes $\F_1$ and $\F_2$, $t\in\{1,2\}$ for the function class $\F_3$ and $t=8$ for the function class $\F_4$. In particular, $\F_3$ depends on $t$, since the function class $\F_2$ depends on $t$.
For the function class $\F_1$ we additionally set $L_{net}=\lceil\log_2 t\rceil$ and $r_{net}=3\cdot t$. 
In our example, we consider $n=200$ and $n=400$, using the \textit{Adam} method of the Python library \textit{Keras} for the least-squares minimization problem \eqref{minp}. For the implementation of the four network architectures, we also use the \textit{Keras} library.

The performance of each estimate is measured by its empirical misclassification risk
\begin{equation}
	\epsilon_{N}(f_n)=\frac{1}{N}\sum_{k=1}^{N}\IND_{\{f_n(\bx_{n+k})\neq y_{n+k}\}}%
	\label{se5eq1}
\end{equation}
where $f_n$ is the corresponding plug-in image classifier based on the training data and  $(\bx_{n+1},y_{n+1}),\dots,(\bx_{n+N},y_{n+N})$ 
are newly generated independent realizations of the random variable $(\bX,Y)$. In our example we choose $N=10^4$.
Since our estimates and the corresponding errors \eqref{se5eq1} depend on randomly chosen data, we compute the classifiers and their errors \eqref{se5eq1} on 20 independently generated data sets $\D_{n+N}$.
Table \ref{table1} lists the median and interquartile range (IQR) of all runs.
\begin{table}[h]
	\centering
	\begin{tabular}{ccccc}
		\hline
		 &  \multicolumn{2}{c}{$\lambda=32$} &  \multicolumn{2}{c}{$\lambda=64$}  \\
		\hline
		 & $n=200$ & $n=400$  & $n=200$ & $n=400$ \\
		\hline
		\textit{approach} & median (IQR) & median (IQR) & median (IQR) & median (IQR) \\
		\hline
		$\F_1$ & 0.3972 (0.0998) & 0.2139 (0.1553) & 0.4044 (0.1379) & 0.2850 (0.3038) \\
		$\F_2$ & 0.3926 (0.0728)  & 0.2312 (0.0768) & 0.2013 (0.2668)  & 0.0768 (0.0351) \\
		$\F_3$ & \textbf{0.1247} (0.0786)  & {0.0610 (0.0322)} & \textbf{0.0476} (0.0263)  & {0.0209 (0.0114)}  \\
		$\F_4$ & {0.1386 (0.0862)} & \textbf{0.0357} (0.0301) & {0.0521 (0.0666)}   & \textbf{0.0206} (0.0154) \\
		\hline
	\end{tabular}
	\caption{Median and interquartile range of the empirical misclassification risk $\epsilon_N(f_n)$.}
	\label{table1}
\end{table}
We observe that the two classifiers using the architectures $\F_3$ and $\F_4$ outperform the two classifiers that do not include additional weight sharing, which supports Remark 4. 
In two out of four cases, the classifier with architecture $\F_4$ performs best.
Moreover, the fourth classifier has the largest relative improvement with increasing sample size,
which could be an indicator of a better rate of convergence.
We also observe that a larger resolution leads to a better performance, which suggests that the error term $\epsilon_{\lambda}$ from Assumption \ref{ass2} is small for large resolutions.

\begin{figure}[h]
	\centering
	\includegraphics[width=.1\textwidth]{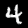}
	\includegraphics[width=.1\textwidth]{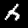}
	\includegraphics[width=.1\textwidth]{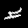}
	\includegraphics[width=.1\textwidth]{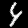}
	\includegraphics[width=.1\textwidth]{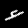}
	\includegraphics[width=.1\textwidth]{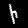}
	\includegraphics[width=.1\textwidth]{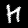}
	\includegraphics[width=.1\textwidth]{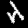}
	\\[\smallskipamount]
	\includegraphics[width=.1\textwidth]{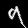}
	\includegraphics[width=.1\textwidth]{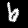}
	\includegraphics[width=.1\textwidth]{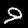}
	\includegraphics[width=.1\textwidth]{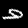}
	\includegraphics[width=.1\textwidth]{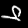}
	\includegraphics[width=.1\textwidth]{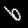}
	\includegraphics[width=.1\textwidth]{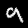}
	\includegraphics[width=.1\textwidth]{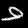}
	\caption{The first row show some images of the fours and the lower row show images of the nines of the MNIST-rot data set.}
	\label{figure8}
\end{figure}

In our second application, we test our CNN image classifiers on real images. Here we use the classes `4' and `9' of the MNIST-rot dataset (\cite{Larochelle2007}), which contains images of handwritten digits. The digits are randomly rotated by angles from $[0,2\pi)$ (see Figure \ref{figure8}).
The resulting data set consists of $2,400$ training images and $N=10,000$ test images of resolution $\lambda=28$. Out of the 2,400 
training images, we randomly select $n/2$ training images per class and evaluate our classifiers using the corresponding $N$ test images. We choose the parameters of our CNN image classfiers as above. The median and interquartile range (IQR) of the empirical misclassification risk \eqref{se5eq1} of 20 runs are presented in Table \ref{table2}.
\begin{table}[h]
	\centering
	\begin{tabular}{ccc}
		\hline
		 &  \multicolumn{2}{c}{$\lambda=28$} \\
		\hline
		 & $n=200$ & $n=400$ \\ %
		\hline
		\textit{approach} & median (IQR) & median (IQR)\\
		\hline
		$\F_1$ & 0.2965 (0.0669) & 0.2123 (0.0492) \\ %
		$\F_2$ & 0.3201 (0.0482)  & 0.2153 (0.0421) \\ %
		$\F_3$ & 0.1627 (0.0577)  & 0.1106 (0.0397) \\ %
		$\F_4$ & \textbf{0.1169} (0.0397)  & \textbf{0.0771} (0.0246) \\
		\hline
	\end{tabular}
	\caption{Median and interquartile range of the empirical misclassification risk $\epsilon_N(f_n)$ based on the corresponding subsets of the MNIST-rot data set.}
	\label{table2}
\end{table}
We observe that the classifier using the function class $\F_4$ outperforms the other classfiers.
\section{Proofs}
\label{se6}
\subsection{An approximation result}
In this subsection, we show that a rotationally symmetric hierarchical max-pooling model can be approximated by a convolutional neural network. 
\begin{lemma}
\label{le1}
	Let $n,l,\lambda\in\N$ with $(2^l+2\cdot l-1)\leq\lambda$. Let $0<h\leq2^l/(\sqrt{2}\cdot\lambda)$, set $b=({2^l+2\cdot l-1})/({2\cdot\lambda})$
	and let $p\in[1,\infty)$.
	Let $\eta:[0,1]^{C_1}\rightarrow[0,1]$ be a function that satisfies a $(p,C)$-smooth rotationally symmetric hierarchical max-pooling model of level $l$, width $h$ and border distance $b$.
	Furthermore, assume Assumption \ref{ass1} for $(p,C)$-smooth functions $\{g_{0,s}\}_{s=1,\dots,4^l}$ and Assumption \ref{ass2} for some $\epsilon_{\lambda}\in[0,1]$, some measurable  $A\subset[0,1]^{C_1}$ and $c>1$. Choose the parameters $L_{n}$ and $\btheta=(t,L,\bk,\bM,B,L_{net},r_{net})$ as in Theorem \ref{th1}.
	Then there exist some $f_{CNN}\in\F_{\btheta}^{CNN}$ such that %
	\begin{align*}
		\left|f_{CNN}(g_{\lambda}(\phi))-\eta(\phi)\right|^2\leq \nconst\cdot\left(n^{-\frac{2\cdot p}{2\cdot p+4}}+\epsilon_{\lambda}^2\right)
	\end{align*}
	holds for all $\phi\in A$ and some constant $\const>0$ which does not depend on $\lambda$ and $n$.
\end{lemma}
We will prove Lemma \ref{le1} at the end of this subsection and first present some auxiliary results.
First we show that the rotationally symmetric max-pooling model can be approximated by the discretized hierarchical max-pooling model introduced in the following definition. This new model is similar to the hierarchical max-pooling model of \cite{KoKrWa2020} (see Definition \ref{de1}) with the main difference that the positions of the hierarchically combined subparts are variable. Throughout this subsection we will use the following notation:
For $k\in\N_0$ and $\lambda\in\N$ we define the index set
\begin{equation*}
		I^{(k)}
	=
	\left\{-\frac{\lceil2^{k-1}\rceil+k-1}{\lambda},\dots,\frac{-1}{\lambda},0,\frac{1}{\lambda},\dots,\frac{\lceil2^{k-1}\rceil+k-1}{\lambda}\right\}^2\subset\R^2,
\end{equation*}
where we have
$I^{(0)}=\{0\}\times\{0\}$.
\begin{definition}
	\label{de3}
	Let $\lambda,l,d\in\N$ with $2^l+2\cdot l-1\leq\lambda$. 
	
	\noindent
	{\bf a)}
	We say that $\bar{\eta}:[0,1]^{G_{\lambda}}\rightarrow\R$ satisfies a {\bf discretized max-pooling model} {\bf of order $d$}
	if there exist functions $\bar{f}^{(i)}:[0,1]^{I^{(l)}}\rightarrow\R$ for $i\in\{1,\dots,d\}$ such that
	\[
	\bar{\eta}(\bx)=\max_{\bu\in G_{\lambda}~:~\bu+I^{(l)}\subseteq G_{\lambda}}\max_{i\in\{1,\dots,d\}}\bar{f}^{(i)}(\bx_{\bu+I^{(l)}}).
	\]
	
	\noindent
	{\bf b)} 
	We say that 
	$\bar{f}:[0,1]^{I^{(l)}}\rightarrow\R$
	satisfies a {\bf discretized hierarchical model of level $l$ with functions $\{\bar{g}_{k,s}\}_{k\in\{0,\dots,l\},s\in\{1,\dots,4^{l-k}\}}$}, where
	\[
	\bar{g}_{k,s}:\R^{4}\rightarrow\R_+\quad\big(k=1,\dots,l,s=1,\dots,4^{l-k}\big)
	\]
	and
	\[
	\bar{g}_{0,s}:[0,1]\rightarrow\R_+\quad\big(s=1,\dots,4^l\big),
	\]
	if there exist grid points
	\[
\bi_{k,s}\in\left\{-\frac{\lfloor2^{k-1}\rfloor+1}{\lambda},\dots,0,\dots,\frac{\lfloor2^{k-1}\rfloor+1}{\lambda}\right\}^2\quad\big(k=0,\dots,l-1,s=1,\dots,4^{l-k}\big)%
\]
	such that we have
	\[
	\bar{f}=\bar{f}_{l,1}
	\]
	for some $\bar{f}_{k,s}:[0,1]^{I^{(k)}}\rightarrow\R$ recursively defined by
	\begin{align*}
		\bar{f}_{k,s}(\bx)=\bar{g}_{k,s}\Big(&\bar{f}_{k-1,4\cdot(s-1)+1}(\bx_{\bi_{k-1,4\cdot(s-1)+1}+I^{(k-1)}}),
		\bar{f}_{k-1,4\cdot(s-1)+2}(\bx_{\bi_{k-1,4\cdot(s-1)+2}+I^{(k-1)}}),\\
		&\bar{f}_{k-1,4\cdot(s-1)+3}(\bx_{\bi_{k-1,4\cdot(s-1)+3}+I^{(k-1)}}),
		\bar{f}_{k-1,4\cdot s}(\bx_{\bi_{k-1,4\cdot s}+I^{(k-1)}})\Big)
	\end{align*}
	for $k=1,\dots,l$ and $s=1,\dots,4^{l-k}$
	and
	\[
	\bar{f}_{0,s}(x)=\bar{g}_{0,s}(x)
	\]
	for $s=1,\dots,4^l$.
	
	\noindent
	{\bf c)}
		We say that
	$\bar{\eta}: [0,1]^{G_{\lambda}} \rightarrow \R$
	satisfies a {\bf discretized hierarchical max-pooling model of level $l$ and order $d$
	 with functions} $\big\{\bar{g}^{(i)}_{k,s}\big\}_{i\in\{1,\dots,d\},k\in\{0,\dots,l\},s\in\{1,\dots,4^{l-k}\}}$,
	if $\bar{\eta}$ satisfies a discretized max-pooling model of order $d$
	and the functions
	$\bar{f}^{(i)}:[0,1]^{I^{(l)}} \rightarrow \R$ in the definition of this discretized
	max-pooling model satisfy a discretized hierarchical model
	of level $l$ with functions $\big\{\bar{g}^{(i)}_{k,s}\big\}_{k\in\{0,\dots,l\},s\in\{1,\dots,4^{l-k}\}}$ for all $i\in\{1,\dots,d\}$.
\end{definition}
We now show that we can approximate the rotationally symmetric hierarchical max-pooling model by a discretized hierarchical max-pooling model if the functions $\bar{g}_{k,s}^{(i)}$ from the discretized model correspond to the functions $g_{k,s}$ from the continuous model.
\begin{lemma} 
	\label{le2}
Let $\lambda,l\in\N$ with $2^l+2\cdot l-1\leq\lambda$, and set 
$b=({2^l+2\cdot l-1})/({2\cdot\lambda})$.
Furthermore, let $0<h\leq2^l/(\sqrt{2}\cdot\lambda)$ and set $h_k=h/2^{l-k}$ for $k\in\Z$.
We assume that $\eta:[0,1]^{C_1}\rightarrow\R$ satisfies a rotationally symmetric max-pooling model of level $l$,
width $h$, and border distance $b$ given by the functions
\[
g_{k,s}:\R^4\rightarrow[0,1]\quad(k=1,\dots,l,s=1,\dots,4^{l-k})
\]
and functions 
\[
f_{0,s}:[0,1]^{C_{h_0}}\rightarrow[0,1]\quad(s=1,\dots,4^l),
\]
and
let the functions $f_{k,s}:[0,1]^{C_{h_k}}\rightarrow[0,1]$ $(k=1,\dots,l,s=1,\dots,4^{l-k})$ defined as in Definition \ref{de2}. Moreover, we assume that all restrictions $g_{k,s}\big|_{[0,1]^4}:[0,1]^4\rightarrow[0,1]$ are Lipschitz continous regarding the maximum metric with Lipschitz constant $L>0$ and 
that Assumption \ref{ass2} is satisfied 
for some $\epsilon_{\lambda}\in[0,1]$, some measurable $A\subset[0,1]^{C_1}$ and $c>1$.
Then there exist a discretized hierarchical max-pooling model $\bar{\eta}:[0,1]^{G_{\lambda}}\rightarrow\R$ of level $l$ and order 
\begin{equation}
	\label{l2eq1}
	d=\left\lceil\frac{2^{l-1/2}\cdot\pi}{c-1
	}\right\rceil
\end{equation}
 with functions $\{\bar{g}_{k,s}^{(i)}\}$, where
 \[
 \bar{g}^{(i)}_{k,s}=g_{k,s}\quad\big(i=1,\dots,d,k=0,\dots,l,s=1,\dots,4^{l-k}\big)
 \]
with $g_{0,s}(x)=f_{0,s}(x\cdot1\big|_{C_{h_0}})$ $(x\in[0,1])$ for $s=1,\dots,4^l$
such that
\[
|\bar{\eta}(g_{\lambda}(\phi))-\eta(\phi)|\leq L^{l}\cdot\epsilon_{\lambda}\quad\big(\phi\in A\big).
\]
\end{lemma}
\begin{remark-customized}
For $p\in[1,\infty)$, the Lipschitz continuity of the restrictions $g_{k,s}\big|_{[0,1]^4}$ is a consequence of the $(p,C)$-smoothness of the functions $g_{k,s}$.
\end{remark-customized}

\noindent
{\bf Proof.}
In the proof we use that for $n\in\N$, $a_1,\dots,a_n,b_1,\dots,b_n\in\R$ it holds that
\begin{equation}
	\label{ple1eq1}
	\left|\max_{i=1,\dots,n}a_i-\max_{i=1,\dots,n}b_i\right|\leq\max_{i=1,\dots,n}|a_i-b_i|,
\end{equation}
which follows from the fact that
in case $a_j=\max_{i=1,\dots,n}a_i\geq\max_{i=1,\dots,n}b_i$ (which we can assume w.l.o.g.) we have
\[
\left|\max_{i=1,\dots,n}a_i-\max_{i=1,\dots,n}b_i\right|=a_j-\max_{i=1,\dots,n}b_i\leq a_j-b_j\leq\max_{i=1,\dots,n}|a_i-b_i|.
\]
Before we completely define the discretized hierarchical max-pooling model $\bar{\eta}$, i.e., before we define the corresponding grid points, we will bound $|\bar{\eta}(g_{\lambda}(\phi))-\eta(\phi)|$ using equation \eqref{ple1eq1}.
Therefore we define the grid $G=\{\bu\in G_{\lambda}~:~\bu+I^{(l)}\subseteq G_{\lambda}\}$ and the cubes
\[
P_{\bu}=\Bigg(\bu+\left[-\frac{1}{2\lambda},\frac{1}{2\lambda}\right]^2\Bigg)\cap\left[-\frac{1}{2}+b,\frac{1}{2}-b\right]^2
\quad\big(\bu\in G\big)
\]
such that the definitions of $G_{\lambda}$, $I^{(l)}$ and $b$ yield
\begin{equation}
	\label{ple1eq2}
	\begin{split}
\bigcup_{\bu\in G}P_{\bu}&=
\bigcup_{\bu\in G_{\lambda}~:~\bu+I^{(l)}\subseteq G_{\lambda}}
\left(\bu+\left[-\frac{1}{2\lambda},\frac{1}{2\lambda}\right]^2\right)\cap\left[-\frac{1}{2}+b,\frac{1}{2}-b\right]^2
\\
&=
\bigcup
\Bigg\{\bu+\left[-\frac{1}{2\lambda},\frac{1}{2\lambda}\right]^2~:~
\bu\in
\left\{-\frac{1}{2}+\frac{2^{l-1}+l-\frac{1}{2}}{\lambda},\dots,\frac{1}{2}-\frac{2^{l-1}+l-\frac{1}{2}}{\lambda}\right\}^2
\Bigg\}\\
&\quad\cap\left[-\frac{1}{2}+\frac{2^{l-1}+l-\frac{1}{2}}{\lambda},\frac{1}{2}-\frac{2^{l-1}+l-\frac{1}{2}}{\lambda}\right]^2
\\
&=\left[-\frac{1}{2}+\frac{2^{l-1}+l-\frac{1}{2}}{\lambda},\frac{1}{2}-\frac{2^{l-1}+l-\frac{1}{2}}{\lambda}\right]^2
\\
&=\left[-\frac{1}{2}+b,\frac{1}{2}-b\right]^2
	\end{split}
\end{equation}
Furthermore, definition \eqref{l2eq1} allows us to cover $[0,2\pi]$ 
by intervals $\{\Theta_i\}_{i=1,\dots,d}$ of side length $(c-1)/(2^{l-3/2})$ with centers $\{\alpha_i\}_{i=1,\dots,d}$.
Then, for $\phi\in A$ and $\bx\coloneqq g_{\lambda}(\phi)$ inequality \eqref{ple1eq1} and equation \eqref{ple1eq2} imply
\begin{align*}
	&|\bar{\eta}(\bx)-\eta(\phi)|\\
	&=\left|\max_{\bu\in G_{\lambda}~:~\bu+I^{(l)}\subseteq G_{\lambda}}\max_{i\in\{1,\dots,d\}}\bar{f}^{(i)}_{l,1}(\bx_{\bu+I^{(l)}})-\sup_{\bv\in\left[-\frac{1}{2}+b,\frac{1}{2}-b\right]^2}\sup_{\alpha\in[0,2\pi]}f_{l,1}(\phi\circ \tau_{\bv}\circ rot^{(\alpha)}\big|_{C_h})\right|\\
	&=\left|\max_{\bu\in G}\max_{i\in\{1,\dots,d\}}\bar{f}^{(i)}_{l,1}(\bx_{\bu+I^{(l)}})-\max_{\bu\in G}\sup_{\bv\in P_{\bu}}\max_{i\in\{1,\dots,d\}}\sup_{\alpha\in\Theta_i}f_{l,1}(\phi\circ \tau_{\bv}\circ rot^{(\alpha)}\big|_{C_h})\right|\\
	&\leq\max_{\bu\in G}\left|\max_{i\in\{1,\dots,d\}}\bar{f}^{(i)}_{l,1}(\bx_{\bu+I^{(l)}})-\sup_{\bv\in P_{\bu}}\max_{i\in\{1,\dots,d\}}\sup_{\alpha\in\Theta_i}f_{l,1}(\phi\circ \tau_{\bv}\circ rot^{(\alpha)}\big|_{C_h})\right|\\
	&\leq\max_{\bu\in G}\sup_{\bv\in P_{\bu}}\max_{i\in\{1,\dots,d\}}
	\sup_{\alpha\in\Theta_i}
	\left|\bar{f}^{(i)}_{l,1}(\bx_{\bu+I^{(l)}})-f_{l,1}(\phi\circ \tau_{\bv}\circ rot^{(\alpha)}\big|_{C_h})\right|.
\end{align*}
It suffices now to show that for all $i\in\{1,\dots,d\}$ there exist 
grid points $\bi^{(i)}_{k,s}$ ($k=0,\dots,l-1$, $s=1,\dots,4^{l-k}$) %
of $\bar{f}^{(i)}_{l,1}$, such that
\begin{equation}
	\label{ple1eq4}
	\left|\bar{f}^{(i)}_{l,1}(\bx_{\bu+I^{(l)}})-f_{l,1}(\phi\circ \tau_{\bv}\circ rot^{(\alpha)}\big|_{C_h})\right|\leq L^l\cdot\epsilon_{\lambda}
\end{equation}
for all $\bu\in G$, $\bv\in P_{\bu}$, $i\in\{1,\dots,d\}$ and $\alpha\in\Theta_i$. 

\begin{figure}[h]
	\centering
	\begin{minipage}{.45\textwidth}
		\centering
\includegraphics{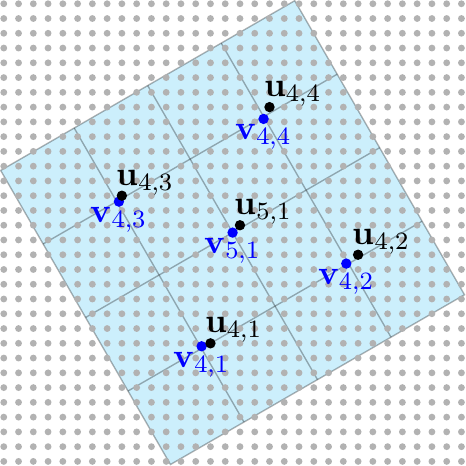}
	\end{minipage}
	\begin{minipage}{.45\textwidth}
		\centering
\includegraphics{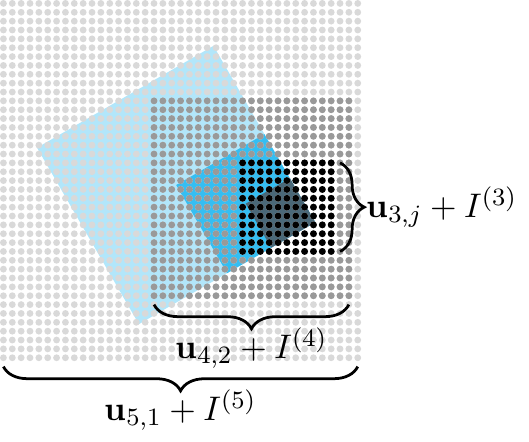}
	\end{minipage}
	\caption{
		On the left hand side $\bv_{k,s}$ and $\bu_{k,s}$ are shown as used in the proof of Lemma \ref{le2}, while on the right hand side one can see the corresponding grids, where $j=4\cdot(2-1)+2=6$. We choosed $\alpha=\pi/6$, $\lambda=100$ and $h=2^5/(\sqrt{2}\cdot\lambda)$.}
	\label{fig9}
\end{figure}

To show this let $\bu\in G$, $\bv\in P_{\bu}$, $i\in\{1,\dots,d\}$ and $\alpha\in\Theta_i$ be fixed for the remainder of the proof.
The idea is to construct the grid points $\bi^{(i)}_{k,s}$, which do not depend on $\bu$, $\bv$ and $\alpha$, such that we are able to prove equation \eqref{ple1eq4} by showing via induction on $k$ that
\begin{equation}
	\label{ple1eq5}
	\left|\bar{f}_{k,s}^{(i)}(\bx_{\bu_{k,s}+I^{(k)}})-f_{k,s}(\phi\circ \tau_{\bv_{k,s}}\circ rot^{(\alpha)}\big|_{C_{h_k}})\right|\leq L^{k}\cdot\epsilon_{\lambda}
\end{equation}
for all $k=0,\dots,l$ and $s=1,\dots,4^{l-k}$ where we set $\bu_{l,1}=\bu$ and $\bv_{l,1}=\bv$, and
\begin{equation}
	\label{ple1eq6}
	\bu_{k-1,4\cdot(s-1)+j}=\bu_{k,s}+\bi^{(i)}_{k-1,4\cdot(s-1)+j}\text{ and }\bv_{k-1,4\cdot(s-1)+j}=\bv_{k,s}+rot^{(\alpha)}\left(\bh_{k-2}^{(j)}\right)
\end{equation}
for $k=1,\dots,l$, $s=1,\dots,4^{l-k}$ and $j=1,\dots,4$ with
\begin{equation*}
	\begin{split}
		&\bh_{k-2}^{(1)}=(-h_{k-2},-h_{k-2}),\\
		&\bh_{k-2}^{(3)}=(-h_{k-2},h_{k-2}),
	\end{split}
	\quad
	\begin{split}
		&\bh_{k-2}^{(2)}=(h_{k-2},-h_{k-2}),\\
		&\bh_{k-2}^{(4)}=(h_{k-2},h_{k-2}).
	\end{split}
\end{equation*}
The rest of the proof is organized in four steps. \textit{In the first step,} we define the grid points $\bi_{k,s}^{(i)}$ and show that they are well-defined according to Definition \ref{de3} b). \textit{In the second step}, we show that $\bu_{k,s}$ is `close' to $\bv_{k,s}$ (see Figure \ref{fig9} for an example). 
\textit{In the third step}, using Assumption \ref{ass2}, we show that equation \eqref{ple1eq5} holds for $k=0$ and \textit{the fourth step }corresponds to the induction step for the proof of equation \eqref{ple1eq5}.

\textit{Step 1}:
First, we consider a subpart of width $h$ rotated around the origin by the angle $\alpha_i$, where $\alpha_i$ is defined as the center of the interval $\Theta_i$. Analogous to the definition of $\bv_{k,s}$, we divide the subpart into smaller and smaller subparts and choose the points $\bz_{z,k}^{(i)}$ as the centers of these subparts. The idea is that $\bz_{k,s}^{(i)}$ is then `close' to $\bv_{k,s}-\bv$, as we will see in the \textit{second step}.
We set $\bz_{l,1}^{(i)}=(0,0)$ and recursively define 
\[
\bz_{k-1,4\cdot(s-1)+j}^{(i)}=\bz_{k,s}^{(i)}+rot^{(\alpha_i)}\left(\bh_{k-2}^{(j)}\right)
\]
for $k=1,\dots,l$, $s=1,\dots,4^{l-k}$ and $j=1,\dots,4$. Since $\bi_{k,s}^{(i)}$ are supposed to be grid points we choose
\begin{equation}
	\label{pl2eqdefbarz}
	\bar{\bz}_{k,s}^{(i)}\in\argmin_{\bz\in I^{(l)}}\|\bz-\bz_{k,s}^{(i)}\|_{\infty},
	\quad
	\big(k=0,\dots,l,s=1,\dots,4^{l-k}\big)
\end{equation}
and define
\[
\bi^{(i)}_{k-1,4\cdot(s-1)+j}=\bar{\bz}_{k-1,4\cdot(s-1)+j}^{(i)}-\bar{\bz}_{k,s}^{(i)}\quad\big(k=1,\dots,l,s=1,\dots,4^{l-k},j=1,\dots,4\big).
\]
To show that the grid points $\bi^{(i)}_{k,s}$ are well-defined according to Definition \ref{de3} b) we use that $h\leq2^l/(\sqrt{2}\cdot\lambda)$ and get
\begin{equation}
	\label{rotb}
	\left\|rot^{(\beta)}\left(\bh_{k-2}^{(j)}\right)\right\|_{\infty}\leq\sqrt{2}\cdot h_{k-2}=\frac{\sqrt{2}\cdot h}{2^{l-(k-2)}}=\frac{2^{k-2}}{\lambda}
\end{equation}
for $k=1,\dots,l$, $j=1,\dots,4$ and an arbitrary angle $\beta\in[0,2\pi]$
and therefore we have
\begin{align*}
	\|\bz_{k-1,4\cdot(s-1)+j}^{(i)}\|_{\infty}&\leq\|\bz_{k,s}^{(i)}\|_{\infty}+\|rot^{(\alpha_i)}\left(\bh_{k-2}^{(j)}\right)\|_{\infty}
	\leq\|\bz_{k,s}^{(i)}\|_{\infty}+\frac{2^{k-2}}{\lambda}
\end{align*}
for $k=1,\dots,l$, $s=1,\dots,4^{l-k}$ and $j=1,\dots,4$.
Since $\bz_{l,1}=(0,0)$ we then have
\begin{equation*}
	\begin{split}
		\|\bz_{k,s}^{(i)}\|_{\infty}\leq\sum_{j=k+1}^{l}\frac{2^{j-2}}{\lambda}
		=\frac{1}{2\cdot\lambda}\left(\sum_{j=0}^{l-1}{2^{j}}-\sum_{j=0}^{k-1}{2^{j}}\right)=\frac{2^{l}-2^k}{2\cdot\lambda}
	\end{split}
\end{equation*}
and due to \eqref{pl2eqdefbarz} and the definition of $I^{(l)}$ we get
\begin{equation}
	\label{ple1eq9}
	\|\bz_{k,s}^{(i)}-\bar{\bz}_{k,s}^{(i)}\|_{\infty}\leq\frac{1}{2\cdot\lambda}
\end{equation}
for $k=0,\dots,l$, $s=1,\dots,4^{l-k}$.
By using the triangle inequality, inequality \eqref{ple1eq9} and inequality \eqref{rotb} we obtain
\begin{align*}
	&\|\bi^{(i)}_{k-1,4\cdot(s-1)+j}\|_{\infty}\\
	&=\|\bar{\bz}_{k,s}^{(i)}-\bar{\bz}^{(i)}_{k-1,4\cdot(s-1)+j}\|_{\infty}\\
	&\leq
	\|\bar{\bz}^{(i)}_{k,s}-\bz_{k,s}^{(i)}\|_{\infty}+\|\bz_{k,s}^{(i)}-\bz_{k-1,4\cdot(s-1)+j}^{(i)}\|_{\infty}+\|\bz_{k-1,4\cdot(s-1)+j}^{(i)}-\bar{\bz}_{k-1,4\cdot(s-1)+j}^{(i)}\|_{\infty}\\
	&\leq\frac{1}{2\cdot\lambda}+\|rot^{(\alpha_i)}(\bh_{k-2}^{(j)})\|_{\infty}+\frac{1}{2\cdot\lambda}\\
	&\leq\frac{2^{k-2}+1}{\lambda}
\end{align*}
for $k=1,\dots,l$, $s=1,\dots,4^{l-k}$ and $j=1,\dots,4$,
which together with the fact that $\bi^{(i)}_{k,s}$ is a vector of integer multiples of $1/\lambda$ implies
\[
\bi^{(i)}_{k,s}\in\left\{-\frac{\lfloor2^{k-1}\rfloor+1}{\lambda},\dots,0,\dots,\frac{\lfloor2^{k-1}\rfloor+1}{\lambda}\right\}^2\quad\big(k=0,\dots,l-1,s=1,\dots,4^{l-k}\big).
\]

\textit{Step 2}:
For $k=1,\dots,l$, $s=1,\dots,4^{l-k}$ and $j=1,\dots,4$ we have
\begin{align*}
	&\|\bz_{k-1,4\cdot(s-1)+j}^{(i)}-(\bv_{k-1,4\cdot(s-1)+j}-\bv)\|_{\infty}\\
	&\leq\|\bz_{k,s}^{(i)}-(\bv_{k,s}-\bv)\|_{\infty}+\left\|rot^{(\alpha_i)}\left(\bh_{k-2}^{(j)}\right)-rot^{(\alpha)}\left(\bh_{k-2}^{(j)}\right)\right\|_{\infty}\\
	&=\|\bz_{k,s}^{(i)}-(\bv_{k,s}-\bv)\|_{\infty}+\left\|
	\left(
	\begin{matrix}
		\cos(\alpha_i)-\cos(\alpha) & \sin(\alpha)-\sin(\alpha_i)\\
		\sin(\alpha)-\sin(\alpha_i) & \cos(\alpha)-\cos(\alpha_i)
	\end{matrix}
	\right)
	\bh_{k-2}^{(j)}\right\|_{\infty}\\
	&\leq\|\bz_{k,s}^{(i)}-(\bv_{k,s}-\bv)\|_{\infty}+2\cdot h_{k-2}\cdot\max\{|\sin(\alpha)-\sin(\alpha_i)|,|\cos(\alpha)-\cos(\alpha_i)|\}\\
	&\leq\|\bz_{k,s}^{(i)}-(\bv_{k,s}-\bv)\|_{\infty}+h_{k-1}\cdot|\alpha-\alpha_i|\\
	&\leq\|\bz_{k,s}^{(i)}-(\bv_{k,s}-\bv)\|_{\infty}+\frac{2^{k-1}}{\sqrt{2}\cdot\lambda}\cdot\frac{\sqrt{2}\cdot(c-1)}{2^l},
\end{align*}
which together with $\bz_{l,1}^{(i)}=\bv_{l,1}-\bv=\bNULL$ implies
\begin{equation}
	\label{ple1eq10}
	\|\bz_{k,s}^{(i)}-(\bv_{k,s}-\bv)\|_{\infty}
	\leq \frac{c-1}{2^l\cdot\lambda}\cdot\sum_{i=k}^{l-1}2^{i}
	=\frac{(c-1)\cdot(2^{l}-2^{k})}{\lambda\cdot2^l}
	<\frac{c-1}{\lambda}
\end{equation}
for $k=0,\dots,l$ and $s=1,\dots,4^{l-k}$.
Furthermore, we 
have 
\begin{equation}
	\label{ple1eq11}
	\bu_{k,s}=\bu+\bar{\bz}_{k,s}^{(i)}
\end{equation}
for $k=0,\dots,l$, since $\bar{\bz}^{(i)}_{l,1}=(0,0)$ and 
\[
\bu_{k-1,4\cdot(s-1)+j}=\bu_{k,s}+\bi^{(i)}_{k-1,4\cdot(s-1)+j}=\bu_{k,s}+\bar{\bz}^{(i)}_{k-1,4\cdot(s-1)+j}-\bar{\bz}^{(i)}_{k,s}
\]
for $k=1,\dots,l$, $s=1,\dots,4^{l-k}$ and $j=1,\dots,4$. Inequalities \eqref{ple1eq9}, \eqref{ple1eq10} and \eqref{ple1eq11} imply
\begin{equation}
	\label{ple1eq12}
	\begin{split}
		\|\bu_{k,s}-\bv_{k,s}\|_{\infty}
		&=\|\bu-\bv+\bar{\bz}_{k,s}^{(i)}-{\bz}_{k,s}^{(i)}+{\bz}_{k,s}^{(i)}-\bv_{k,s}+\bv\|_{\infty}
		\\
		&\leq\|\bu-\bv\|_{\infty}+\|\bar{\bz}_{k,s}^{(i)}-\bz_{k,s}^{(i)}\|_{\infty}+\left\|\bz_{k,s}^{(i)}-(\bv_{k,s}-\bv)\right\|_{\infty}\\
		&\leq\frac{1}{2\cdot\lambda}+\frac{1}{2\cdot\lambda}+\frac{c-1}{\lambda}\\
		&=\frac{c}{\lambda}
	\end{split}
\end{equation}
for all $k=0,\dots,l$ and $s=1,\dots,4^{l-k}$.

\textit{Step 3}:
To use Assumption \ref{ass2}, we first show that $\bv_{0,s}\in[h_0/\sqrt{2}-1/2,1/2-h_0/\sqrt{2}]^2$ for all $s=1,\dots,4^l$.
By using inequality \eqref{rotb} we get
\begin{align*}
	\|\bv_{k-1,4\cdot(s-1)+j}-\bv\|_{\infty}&\leq\|\bv_{k,s}-\bv\|_{\infty}+\|rot^{(\alpha)}\left(\bh_{k-2}^{(j)}\right)\|_{\infty}
	\leq\|\bv_{k,s}-\bv\|_{\infty}+\frac{2^{k-2}}{\lambda}
\end{align*}
for $k=1,\dots,l$, $s=1,\dots,4^{l-k}$ and $j=1,\dots,4$, which together with $\bv_{l,1}=\bv$ implies
\begin{equation}
	\label{le1eqvks}
	\|\bv_{k,s}-\bv\|_{\infty}
	\leq
	\sum_{j=k+1}^{l}\frac{2^{j-2}}{\lambda}
	=\frac{1}{2\cdot\lambda}\left(\sum_{j=0}^{l-1}{2^{j}}-\sum_{j=0}^{k-1}{2^{j}}\right)
	=\frac{2^l-2^k}{2\cdot\lambda}
\end{equation}
for $k=0,\dots,l$ and $s=1,\dots,4^{l-k}$. By using inequality \eqref{le1eqvks}, $\bv\in[-1/2+b,1/2-b]^2$  and $h_0\leq1/(\sqrt{2}\cdot\lambda)$ we get
\begin{equation}
	\label{ple1eq13}
	\begin{split}
		\|\bv_{0,s}\|_{\infty}&\leq\|\bv\|_{\infty}+\|\bv_{0,s}-\bv\|_{\infty}\\
		&{\leq}\frac{1}{2}-b+\frac{2^l-1}{2\cdot\lambda}\\
		&{\leq}\frac{1}{2}-\frac{2^{l}+2\cdot l-1}{2\cdot\lambda}+\frac{2^l-1}{2\cdot\lambda}\\
		&=\frac{1}{2}-\frac{l}{\lambda}\\
		&\leq\frac{1}{2}-\frac{1/(\sqrt{2}\cdot\lambda)}{\sqrt{2}}\\
		&\leq\frac{1}{2}-\frac{h_0}{\sqrt{2}}
	\end{split}
\end{equation}
for $s=1,\dots,4^{l}$. 
By using Assumption \ref{ass2}, \eqref{ple1eq12} and \eqref{ple1eq13} we obtain
\begin{align*}
	&\left|\bar{f}^{(i)}_{0,s}(\bx_{\bu_{0,s}+I^{(0)}})-f_{0,s}(\phi\circ \tau_{\bv_{0,s}}\circ rot^{(\alpha)}\big|_{C_{h_0}})\right|\\
	&=\left|g_{0,s}(x_{\bu_{0,s}})-f_{0,s}(\phi\circ \tau_{\bv_{0,s}}\circ rot^{(\alpha)}\big|_{C_{h_0}})\right|\\
	&=\left|f_{0,s}(\phi({\bu_{0,s}})\cdot1_{C_{h_0}})-f_{0,s}(\phi\circ \tau_{\bv_{0,s}}\circ rot^{(\alpha)}\big|_{C_{h_0}})\right|\\
	&\leq\epsilon_{\lambda}
\end{align*}
for $s=1,\dots,4^{l}$.

\textit{Step 4}:
Now we assume that \eqref{ple1eq5} holds for some $k\in\{0,\dots,l-1\}$ and all $s\in\{1,\dots,4^{l-k}\}$.
Because of the Lipschitz assumption on the functions $g_{k,s}$, definition \eqref{ple1eq6}, the linearity of the function $rot^{( \alpha)}$ and the induction hypothesis \eqref{ple1eq5}, we conclude that
\begin{align*}
	&\left|\bar{f}^{(i)}_{k+1,s}(\bx_{\bu_{k+1,s}+I^{(k+1)}})-f_{k+1,s}(\phi\circ \tau_{\bv_{k+1,s}}\circ rot^{(\alpha)}\big|_{C_{h_{k+1}}})\right|\\
	&=\Big|g_{k+1,s}\Big(\bar{f}^{(i)}_{k-1,4\cdot(s-1)+1}(\bx_{\bu_{k+1,s}+\bi^{(i)}_{k,4\cdot(s-1)+1}+I^{(k)}}),
	\bar{f}^{(i)}_{k,4\cdot(s-1)+2}(\bx_{\bu_{k+1,s}+\bi^{(i)}_{k,4\cdot(s-1)+2}+I^{(k)}}),\\
	&\hspace{1.5cm}\bar{f}^{(i)}_{k,4\cdot(s-1)+3}(\bx_{\bu_{k+1,s}+\bi^{(i)}_{k,4\cdot(s-1)+3}+I^{(k)}}),
	\bar{f}^{(i)}_{k,4\cdot s}(\bx_{\bu_{k+1,s}+\bi^{(i)}_{k,4\cdot s}+I^{(k)}})\Big)\\
	&\quad-g_{k+1,s}\Big(f_{k,4\cdot(s-1)+1}(\phi\circ \tau_{\bv_{k+1,s}}\circ rot^{(\alpha)}\circ \tau_{(-h_{k-1},-h_{k-1})}\big|_{C_{h_{k}}}),\\
	&\hspace{1.5cm}f_{k,4\cdot(s-1)+2}(\phi\circ \tau_{\bv_{k+1,s}}\circ rot^{(\alpha)}\circ \tau_{(h_{k-1},-h_{k-1})}\big|_{C_{h_{k}}}),\\
	&\hspace{1.5cm}f_{k,4\cdot(s-1)+3}(\phi\circ \tau_{\bv_{k+1,s}}\circ rot^{(\alpha)}\circ \tau_{(-h_{k-1},h_{k-1})}\big|_{C_{h_{k}}}),\\
	&\hspace{1.5cm}f_{k,4\cdot s}(\phi\circ \tau_{\bv_{k+1,s}}\circ rot^{(\alpha)}\circ \tau_{(h_{k-1},h_{k-1})}\big|_{C_{h_{k}}})\Big)
	\Big|\\
	&=\Big|g_{k+1,s}\Big(\bar{f}^{(i)}_{k,4\cdot(s-1)+1}(\bx_{\bu_{k,4\cdot(s-1)+1}+I^{(k)}}),
	\bar{f}^{(i)}_{k,4\cdot(s-1)+2}(\bx_{\bu_{k,4\cdot(s-1)+2}+I^{(k)}}),\\
	&\hspace{1.5cm}\bar{f}^{(i)}_{k,4\cdot(s-1)+3}(\bx_{\bu_{k,4\cdot(s-1)+3}+I^{(k)}}),
	\bar{f}^{(i)}_{k,4\cdot s}(\bx_{\bu_{k,4\cdot s}+I^{(k)}})\Big)\\
	&\quad-g_{k+1,s}\Big(f_{k,4\cdot(s-1)+1}(\phi\circ \tau_{\bv_{k+1,s}}\circ \tau_{rot^{(\alpha)}(\bh_{k-1}^{(1)})}\circ rot^{(\alpha)}\big|_{C_{h_{k}}}),\\
	&\hspace{1.5cm}f_{k,4\cdot(s-1)+2}(\phi\circ \tau_{\bv_{k+1,s}}\circ \tau_{rot^{(\alpha)}(\bh_{k-1}^{(2)})}\circ rot^{(\alpha)}\big|_{C_{h_{k}}}),\\
	&\hspace{1.5cm}f_{k,4\cdot(s-1)+3}(\phi\circ \tau_{\bv_{k+1,s}}\circ \tau_{rot^{(\alpha)}(\bh_{k-1}^{(3)})}\circ rot^{(\alpha)}\big|_{C_{h_{k}}}),\\
	&\hspace{1.5cm}f_{k,4\cdot s}(\phi\circ \tau_{\bv_{k+1,s}}\circ \tau_{rot^{(\alpha)}(\bh_{k-1}^{(4)})}\circ rot^{(\alpha)}\big|_{C_{h_{k}}})\Big)
	\Big|\\
	&\leq L\cdot\max_{j\in\{1,\dots,4\}}\Big|\bar{f}^{(i)}_{k,4\cdot(s-1)+j}(\bx_{\bu_{k,4\cdot(s-1)+j}+I^{(k)}})\\
	&\hspace{2.5cm}
	-f_{k,4\cdot(s-1)+j}(\phi\circ \tau_{\bv_{k,4\cdot(s-1)+j}}\circ rot^{(\alpha)}\big|_{C_{h_{k}}})\Big|\\
	&\leq L^{k+1}\cdot\epsilon_{\lambda}
\end{align*}
for all $s\in\{1,\dots,4^{l-(k+1)}\}$.
\hfill $\Box$

\noindent
Now, we show how to bound the error that occurs once the functions ${g}_{k,s}^{(i)}$ in the discretized hierarchical max-pooling model are replaced by approximations $\bar{g}_{k,s}^{(i)}$.
The result is similar to Lemma 4 from \cite{KoKrWa2020} for the generalized hierarchical max-pooling model.
\begin{lemma}
\label{le3}
	Let $\lambda,l,t\in\N$ with $2^l+2\cdot l-1\leq\lambda$, and let
	\[
	g_{k,s}^{(i)}:\R^4\rightarrow[0,1],~\bar{g}_{k,s}^{(i)}:\R^4\rightarrow\R_+
	\quad\big(i=1,\dots,t,k=1,\dots,l,s=1,\dots,4^{l-k}\big),
	\]
	and
	\[
	g_{0,s}^{(i)}:[0,1]\rightarrow[0,1],~\bar{g}_{0,s}^{(i)}:[0,1]\rightarrow[0,2]\quad\big(i=1,\dots,t,s=1,\dots,4^{l}\big)
	\]
	be functions such that the restrictions $\{g_{k,s}^{(i)}\big|_{[0,2]^4}\}_{i=1,\dots,t,k=1,\dots,l,s=1,\dots,4^{l-k}}$ are Lipschitz continuous (with respect to the maximum metric) with Lipschitz constant $C>0$ and
	\[
	\left\|\bar{g}_{k,s}^{(i)}\right\|_{[0,2]^4,\infty}\leq2\quad\big(i=1,\dots,t,k=1,\dots,l,s=1,\dots,4^{l-k}\big).
	\]
Let $\eta:[0,1]^{G_{\lambda}}\rightarrow\R$ be a function that satisfies a discretized hierarchical max-pooling model of level $l$ and order $t$ with functions $g_{k,s}^{(i)}$
	and $\bar{\eta}:[0,1]^{G_{\lambda}}\rightarrow\R$ be a function that satisfies a discretized hierarchical max-pooling model of level $l$ and order $t$ with functions $\bar{g}_{k,s}^{(i)}$.
Furthermore, we assume that the two discretized hierarchical max-pooling models have the same grid points
$\{\bi_{k,s}^{(i)}\}$.
	Then for any $\bx\in[0,1]^{G_{\lambda}}$ it holds:
	\begin{align*}
		&|{\eta}(\bx)-\bar{\eta}(\bx)|
		\\&
		\leq(C+1)^l\cdot
		\max_{\substack{i\in\{1,\dots,t\},j\in\{1,\dots,4^l\},\\k\in\{1,\dots,l\},s\in\{1,\dots,4^{l-k}\}}}
		\left\{\|g^{(i)}_{0,j}-\bar{g}^{(i)}_{0,j}\|_{[0,1],\infty},\|g_{k,s}^{(i)}-\bar{g}^{(i)}_{k,s}\|_{[0,2]^4,\infty}\right\}.
	\end{align*}
\end{lemma}
{\bf Proof.}
The result follows by applying the triangle inequality and further straightforward
standard techniques. For the sake of completeness a complete proof is given in the
supplement.
\hfill $\Box$

\noindent
Next, we show that we can compute a discretized hierarchical max-pooling model by a convolutional neural network if the functions $\bar{g}^{(i)}_{k,s}$ correspond to standard feedforward neural networks.
\begin{lemma}
	\label{le4}
	Let $\lambda,l,t\in\N$ with $2^l+2\cdot l-1\leq\lambda$.
		For $L_{net},r_{net}\in\N$ let
			\[
		{g}^{(i)}_{net,k,s}\in\G_4(L_{net},r_{net})\quad\big(i=1,\dots,t,k=1,\dots,l, s=1,\dots,4^{l-k}\big)
		\]
		and
		\[
		{g}^{(i)}_{net,0,s}\in\G_1(L_{net},r_{net})\quad\big(i=1,\dots,t,s=1,\dots,4^{l}\big).
		\] Assume that the function
	$\bar{\eta}:[0,1]^{G_{\lambda}}\rightarrow\R$ satisfies a discretized max-pooling model of level $l$ and order $t$ with functions $\{\bar{g}^{(i)}_{k,s}\},$ 
	where we set
	\[
	\bar{g}^{(i)}_{k,s}=\sigma\circ{g}^{(i)}_{net,k,s}\quad\big(i=1,\dots,t,k=0,\dots,l,s=1,\dots,4^{l-k}\big).
	\]
	Set $B=2^{l-1}+(l-1)$, $L_t=\lceil\log_2 t\rceil$, $r_t=3\cdot t$, $k_r=5\cdot4^{l-1}+r_{net}$ for $r=1,\dots,L$,
	\[
	L=\frac{4^{l+1}-1}{3}\cdot(L_{net}+1),
	\]
	and for $k=0,\dots,l$ set
	\[
	M_r=\IND_{\{k>1\}}\cdot2^{k-1}+3\quad\Bigg(r={\sum_{i=0}^{k-1}4^{l-i}\cdot(L_{net}+1)}+1,\dots,\sum_{i=0}^{k}4^{l-i}\cdot(L_{net}+1)\Bigg),
	\]
	where we define the empty sum as zero.
	Then there exist some $f_{CNN}\in\F_{\btheta}^{CNN}$ with $\btheta=(t,L,\bk,\bM,B,L_t,r_t)$ such that
	\[
	\bar{\eta}(\bx)=f_{CNN}(\bx)
	\]
	holds for all $\bx\in[0,1]^{G_{\lambda}}$.
\end{lemma}
{\bf Proof.}
The proof is similar to the proof of Lemma 5 from \cite{KoKrWa2020} and can be found in the supplement.
\hfill $\Box$

\noindent
{\bf Proof of Lemma \ref{le1}.}
Let $\bar{\eta}$ be the discretized hierarchical max-pooling model of level $l$ and order $t$ which is given by the functions
$\{\bar{g}_{k,s}^{(i)}\}$ and grid points $\{\bi_{k,s}^{(i)}\}$
from Lemma \ref{le2} (due to Assumption \ref{ass1}, the functions $\{\bar{g}_{0,s}^{(i)}\}$ have $(p,C)$-smooth extensions on $\R$),
such that
\begin{equation}
	\label{pth1eq1}
	\left|\eta(\phi)-\bar{\eta}(g_{\lambda}(\phi))\right|\leq\nconst\cdot\epsilon_{\lambda}.
\end{equation}
for all $\phi\in A$ and some constant $\const>0$.
Furthermore, let $g_{net,0,s}^{(i)}\in\G_1(L_n,r_{net})$ %
and $g_{net,k,s}^{(i)}\in\G_4(L_n,r_{net})$ $(k>0)$ %
be the standard feedforward neural networks from \cite{KoLa2021} (cf., Lemma 7 from the supplement) which satisfy
\[
\left\|\bar{g}^{(i)}_{k,s}-\sigma\circ g_{net,k,s}^{(i)}\right\|_{[0,2]^4,\infty}
\leq\left\|\bar{g}^{(i)}_{k,s}-g_{net,k,s}^{(i)}\right\|_{[0,2]^4,\infty}
\leq \nconst\cdot L_{n}^{-\frac{2\cdot p}{4}}\leq \nconst\cdot n^{-\frac{p}{2\cdot p+4}}
\]
for $i=1,\dots,t$, $k=1,\dots,l$, $s=1,\dots,4^{l-k}$ and some constants $\mconst,\const>0$ and
\[
\left\|\bar{g}_{0,s}^{(i)}-\sigma\circ g_{net,0,s}^{(i)}\right\|_{[0,1],\infty}
\leq\left\|\bar{g}_{0,s}^{(i)}-g_{net,0,s}^{(i)}\right\|_{[0,1],\infty}
\leq\nconst\cdot L_{n}^{-{2\cdot p}}\leq\nconst\cdot n^{-\frac{p}{2\cdot p+1}},
\]
for $i=1,\dots,t$, $s=1,\dots,4^{l}$ and some constants $\mconst,\const>0$,
where we choose $c_1$ in the definition of $L_{n}$ sufficiently large such that the triangle inequality and the fact that the functions $\bar{g}_{k,s}^{(i)}$ are $[0,1]$-valued imply
\[
\left\|\sigma\circ g_{net,k,s}^{(i)}\right\|_{[0,2]^4,\infty}
\leq\|\bar{g}_{k,s}^{(i)}\|_{[0,2]^4,\infty}+\left\|\bar{g}^{(i)}_{k,s}-\sigma\circ g_{net,k,s}^{(i)}\right\|_{[0,2]^4,\infty}
\leq
1+\mmconst{3}\cdot L_n^{-\frac{2\cdot p}{4}}
\leq2
\]
for all $k=1,\dots,l$ and $s=1,\dots,4^{l-k}$
and
\[
\left\|\sigma\circ g_{net,0,s}^{(i)}\right\|_{[0,1],\infty}
\leq\left\|\bar{g}_{0,s}^{(i)}\right\|_{[0,1],\infty}+\left\|\bar{g}_{0,s}^{(i)}-\sigma\circ g_{net,0,s}^{(i)}\right\|_{[0,1],\infty}
\leq1+\mmconst{1}\cdot L_n^{-2\cdot p}
\leq2
\]
for all $s=1,\dots,4^l$.
Next we define the convolutional neural network ${f}_{CNN}\in\F^{CNN}$ by using Lemma \ref{le4} such that ${f}_{CNN}$ satisfies a discretized hierarchical max-pooling model which is given by
the functions $\{\sigma\circ g_{net,k,s}^{(i)}\}$ and grid points $\{\bi_{k,s}^{(i)}\}$.
By using $(a+b)^2\leq 2a^2+2b^2$, inequality \eqref{pth1eq1} and Lemma \ref{le3} we get
\begin{align*}
	&\left|f_{CNN}(g_{\lambda}(\phi))-\eta(\phi)\right|^2\\
	&\leq2\cdot\left|f_{CNN}(g_{\lambda}(\phi))-\bar{\eta}(g_{\lambda}(\phi))\right|^2+2\cdot\left|\bar{\eta}(g_{\lambda}(\phi))-\eta(\phi)\right|^2\\
		&\leq \nconst\cdot\Big(\max_{k\in\{1,\dots,l\},s\in\{1,\dots,4^{l-k}\},j\in\{1,\dots,4^l\},i\in\{1,\dots,t\}}\Big\{\|\sigma\circ g_{net,0,j}^{(i)}-\bar{g}^{(i)}_{0,j}\|_{[0,2],\infty},
	\\&\hspace{3cm}
	\|\sigma\circ g_{net,k,s}^{(i)}-\bar{g}_{k,s}^{(i)}\|_{[0,2]^4,\infty}\Big\}\Big)^2+2\cdot\mmconst{5}^2\cdot\epsilon_{\lambda}^2\\
	&\leq \nconst\cdot\left(n^{-\frac{2\cdot p}{2\cdot p+4}}+\epsilon_{\lambda}^2\right)
\end{align*}
for some constants $\mconst,\const>0$ which does not depend on $\lambda$ and $n$.
\hfill $\Box$
\subsection{Proof of Theorem \ref{th1}}
We denote $\F\coloneqq\F_{\btheta}^{CNN}$ and choose $\nconst>0$ so large that $\const\cdot\log n\geq2$ holds (cf., Lemma 10 from the supplement). Then $z\geq1/2$ holds if and only if $T_{\const\cdot\log n}z\geq1/2$, and consequently we have
\[
f_{n}(\bx)=
\begin{cases}
	1&,\text{ if }T_{\const\cdot\log n}\eta_n(\bx)\geq\frac{1}{2}\\
	0&,\text{ elsewhere}.
\end{cases}
\]
Because of Lemma 5 from the supplement we have
\begin{align*}
	&\PROB\{f_n(g_{\lambda}(\Phi))\neq Y\}-\min_{f:[0,1]^{G_{\lambda}}\rightarrow[0,1]}\PROB\{f(g_{\lambda}(\Phi))\neq Y\}
	\\&
	\leq 2\cdot\sqrt{\EXP\left\{\int|T_{\const\cdot\log n}\eta_n(\bx)-\eta^{(\lambda)}(\bx)|^2\PROB_{g_{\lambda}(\Phi)}(d\bx)\right\}}
\end{align*}
and hence it suffices to show
\[
\EXP\left\{\int|T_{\const\cdot\log n}\eta_n(\bx)-\eta^{(\lambda)}(\bx)|^2\PROB_{g_{\lambda}(\Phi)}(d\bx)\right\}\leq \nconst\cdot\Big(\log(\lambda)\cdot(\log n)^4\cdot n^{-\frac{2\cdot p}{2\cdot p+4}}+\epsilon_{\lambda}^2\Big)
\]
for some constant $\const>0$.
By Lemma 6 from the supplement we have
\begin{align*}
	&\EXP\left\{\int|T_{\mconst\cdot\log n}\eta_n(\bx)-\eta^{(\lambda)}(\bx)|^2\PROB_{g_{\lambda}(\Phi)}(d\bx)\right\}
	\\
	&\leq \frac{\nconst \cdot (\log n)^2 \cdot \sup_{\bx_1^n} \left(\log\left(
		\mathcal{N}_1 \left(\frac{1}{n\cdot \mmconst{2}\cdot\log(n)},  T_{\mmconst{2}\cdot\log(n)} \mathcal{F}, \bx_1^n\right)
		\right)+1\right)}{n}\notag\\
	&\quad + 2 \cdot \inf_{f \in \mathcal{F}} \int |f(\bx)-\eta^{(\lambda)}(\bx)|^2 {\PROB}_{g_{\lambda}(\Phi)} (d\bx)
\end{align*}
for some constant $\const>0$.
For the first term Lemma 10 from the supplement implies
\begin{align*}
&\frac{\const\cdot (\log n)^2 \cdot \sup_{\bx_1^n} \left(\log\left(
	\mathcal{N}_1 \left(\frac{1}{n\cdot \mmconst{2}\cdot\log(n)},  T_{\mmconst{2}\cdot\log(n)} \mathcal{F}, \bx_1^n\right)
	\right)+1\right)}{n}\notag\\
&\leq
\frac{\nconst\cdot L^2 \cdot \log(L)\cdot\log(\lambda) \cdot
	{(\log n)}^3}
{n}\\
&\leq
\nconst\cdot\log(\lambda) \cdot
	{(\log n)}^4\cdot n^{-\frac{2\cdot p}{2\cdot p+4}}.
\end{align*}
for some constants $\mconst,\const>0$.
Next we derive a bound on the approximation error
\[
 \inf_{f \in \mathcal{F}} \int |f(\bx)-\eta^{(\lambda)}(\bx)|^2 {\PROB}_{g_{\lambda}(\Phi)} (d\bx).
\]
By using 
the fact that the a posteriori probability $\eta$ minimizes the $L_2$ risk  (w.r.t. the random vector $(\Phi,Y)$), 
$\PROB_{\Phi}(A)=1$ and Lemma \ref{le1},
we get%
\begin{align*}
	\inf_{f \in \mathcal{F}} \int |f(\bx)-\eta^{(\lambda)}(\bx)|^2 {\PROB}_{g_{\lambda}(\Phi)} (d\bx)
	&\leq\int |\bar{f}(\bx)-\eta^{(\lambda)}(\bx)|^2 {\PROB}_{g_{\lambda}(\Phi)} (d\bx)\\
	&=\EXP\left\{|\bar{f}(g_{\lambda}(\Phi))-Y|^2\right\}-\EXP\left\{|\eta^{(\lambda)}(g_{\lambda}(\Phi))-Y|^2\right\}\\
	&\leq\EXP\left\{|\bar{f}(g_{\lambda}(\Phi))-Y|^2\right\}-\EXP\left\{|\eta(\Phi)-Y|^2\right\}\\
	&=\int_{A}|\bar{f}(g_{\lambda}(\phi))-\eta(\phi)|^2{\PROB}_{\Phi} (d\phi)\\
	&\leq \nconst\cdot\Big(n^{-\frac{2\cdot p}{2\cdot p+4}}+\epsilon_{\lambda}\Big)
\end{align*}
for $\bar{f}\in\F$ chosen as in Lemma \ref{le1} and some constant $\const>0$. Summarizing the above results, the proof is complete.
\hfill $\Box$
\setcitestyle{numbers} 
\bibliographystyle{kohler}
\bibliography{Literatur}

\begin{thebibliography}{44}
\providecommand{\natexlab}[1]{#1}

\bibitem[{Anthony and Bartlett(1999)}]{Anthony1999}
Anthony, M., and Bartlett, P.~L. (1999).
\newblock \emph{Neural Network Learning: Theoretical Foundations}.
\newblock Cambridge University Press, Cambridge.

\bibitem[{Bagirov, Clausen and Kohler(2009)}]{Bagirov2009}
Bagirov, A.~M., Clausen, C., and Kohler, M. (2009).
\newblock Estimation of a Regression Function by Maxima of Minima of Linear
  Functions.
\newblock \emph{IEEE Transactions on Information Theory}, \textbf{55}, pp.
  833--845.

\bibitem[{Bartlett et~al.(2019)Bartlett, Harvey, Liaw and
  Mehrabian}]{Bartlett2019}
Bartlett, P.~L., Harvey, N., Liaw, C., and Mehrabian, A. (2019).
\newblock Nearly-tight VC-dimension and Pseudodimension Bounds for Piecewise
  Linear Neural Networks.
\newblock \emph{Journal of Machine Learning Research}, \textbf{20}, pp. 1--17.

\bibitem[{Bauer and Kohler(2019)}]{Bauer2019}
Bauer, B., and Kohler, M. (2019).
\newblock On deep learning as a remedy for the curse of dimensionality in
  nonparametric regression.
\newblock \emph{Annals of Statistics}, \textbf{47}, pp. 2261--2285.

\bibitem[{Bos and Schmidt-Hieber(2021)}]{Bos2021}
Bos, T., and Schmidt-Hieber, J. (2021).
\newblock Convergence rates of deep ReLU networks for multiclass
  classification.
\newblock {a}rXiv: 2108.00969.

\bibitem[{Cabrera-Vives et~al.(2017)Cabrera-Vives, Reyes, F{\"o}rster,
  Est{\'e}vez and Maureira}]{Vives2017}
Cabrera-Vives, G., Reyes, I., F{\"o}rster, F., Est{\'e}vez, P.~A., and
  Maureira, J.~C. (2017).
\newblock Deep-HiTS: Rotation Invariant Convolutional Neural Network for
  Transient Detection.
\newblock {a}rXiv: 1701.00458.

\bibitem[{Cohen and Welling(2016)}]{Cohen2016}
Cohen, T.~S., and Welling, M. (2016).
\newblock Group Equivariant Convolutional Networks.
\newblock \emph{International Conference on Machine Learning (ICML)},
  \textbf{48}, pp. 2990--2999.

\bibitem[{Cover(1968)}]{Cover1968}
Cover, T.~M. (1968).
\newblock Rates of convergence of nearest neighbor procedures.
\newblock \emph{Proceedings of the Hawaii International Conference on Systems
  Siences}, pp. 413--415. Honolulu, HI.

\bibitem[{Cvetkovski(2012)}]{Cvetkovski2012}
Cvetkovski, Z. (2012).
\newblock \emph{Inequalities: Theorems, Techniques and Selected Problems}.
\newblock Springer, Berlin, Heidelberg.

\bibitem[{Delchevalerie et~al.(2021)Delchevalerie, Bibal, Frenay and
  Mayer}]{Delchevalerie2021}
Delchevalerie, V., Bibal, A., Frenay, B., and Mayer, A. (2021).
\newblock Achieving Rotational Invariance with Bessel-Convolutional Neural
  Networks.
\newblock \emph{Advances in Neural Information Processing Systems}.

\bibitem[{Devroye(1982)}]{Devroye1982}
Devroye, L. (1982).
\newblock Necessary and sufficient conditions for the pointwise convergence of
  nearest neighbor regression function estimates.
\newblock \emph{Zeitschrift f{\"{u}}r Wahrscheinlichkeitstheorie und verwandte
  Gebiete}, \textbf{61}, pp. 467--481.

\bibitem[{Devroye, Gy{\"{o}}rfi and Lugosi(1996)}]{Devroye1996}
Devroye, L., Gy{\"{o}}rfi, L., and Lugosi, G. (1996).
\newblock \emph{A Probabilistic Theory of Pattern Recognition}.
\newblock Springer, New York.

\bibitem[{Dieleman, De~Fauw and Kavukcuoglu(2016)}]{Dieleman2016}
Dieleman, S., De~Fauw, J., and Kavukcuoglu, K. (2016).
\newblock Exploiting Cyclic Symmetry in Convolutional Neural Networks.
\newblock \emph{Proceedings of the 33rd International Conference on
  International Conference on Machine Learning}, \textbf{48}, pp. 1889--1898.

\bibitem[{Dieleman, Willett and Dambre(2015)}]{Dieleman2015}
Dieleman, S., Willett, K.~W., and Dambre, J. (2015).
\newblock Rotation-invariant convolutional neural networks for galaxy
  morphology prediction.
\newblock \emph{Monthly Notices of the Royal Astronomical Society},
  \textbf{450}, pp. 1441--1459.

\bibitem[{Du et~al.(2018)Du, Lee, Li, Wang and Zhai}]{Du2018}
Du, S.~S., Lee, J.~D., Li, H., Wang, L., and Zhai, X. (2018).
\newblock Gradient Descent Finds Global Minima of Deep Neural Networks.
\newblock {a}rXiv: 1811.03804.

\bibitem[{Gimel'farb and Delmas(2018)}]{Gimelfarb2018}
Gimel'farb, G., and Delmas, P. (2018).
\newblock \emph{Image Processing And Analysis: A Primer}.
\newblock World Scientific.

\bibitem[{Gonzalez and Woods(2018)}]{Gonzalez2018}
Gonzalez, R.~C., and Woods, R.~E. (2018).
\newblock \emph{Digital Image Processing}.
\newblock Pearson.

\bibitem[{Goodfellow, Bengio and Courville(2016)}]{Goodfellow2016}
Goodfellow, I., Bengio, Y., and Courville, A. (2016).
\newblock \emph{Deep Learning}.
\newblock {MIT} Press, London.

\bibitem[{Gy{\"{o}}rfi et~al.(2002)Gy{\"{o}}rfi, Kohler, Krzyzak and
  Walk}]{Gyoerfi2002}
Gy{\"{o}}rfi, L., Kohler, M., Krzyzak, A., and Walk, H. (2002).
\newblock \emph{A Distribution-Free Theory of Nonparametric Regression}.
\newblock Springer, New York.

\bibitem[{He et~al.(2016)He, Zhang, Ren and Sun}]{He2016}
He, K., Zhang, X., Ren, S., and Sun, J. (2016).
\newblock Deep residual learning for image recognition.
\newblock \emph{Proceedings of the IEEE conference on computer vision and
  pattern recognition}, pp. 770--778.

\bibitem[{Hu, Shang and Cheng(2020)}]{Hu2020}
Hu, T., Shang, Z., and Cheng, G. (2020).
\newblock Sharp Rate of Convergence for Deep Neural Network Classifiers under
  the Teacher-Student Setting.
\newblock {a}rXiv: 2001.06892.

\bibitem[{Imaizumi and Fukamizu(2019)}]{Imaizumi2019}
Imaizumi, M., and Fukamizu, K. (2019).
\newblock Deep neural networks learn non-smooth functions effectively.
\newblock \emph{Proceedings of the 22nd International Conference on Artificial
  Intelligence and Statistics}. Naha, Okinawa, Japan.

\bibitem[{Kim, Ohn and Kim(2021)}]{Kim2018}
Kim, Y., Ohn, I., and Kim, D. (2021).
\newblock Fast convergence rates of deep neural networks for classification.
\newblock \emph{Neural Networks}, \textbf{138}, pp. 179--197.

\bibitem[{Kohler and Krzy{\.z}ak(2017)}]{KoKr17}
Kohler, M., and Krzy{\.z}ak, A. (2017).
\newblock Nonparametric regression based on hierarchical interaction models.
\newblock \emph{{IEEE} Transactions on Information Theory}, \textbf{63}, pp.
  1620--1630.

\bibitem[{Kohler and Krzy{\.z}ak(2021)}]{KoKr2021}
Kohler, M., and Krzy{\.z}ak, A. (2021).
\newblock Over-parametrized deep neural networks minimizing the empirical risk
  do not generalize well.
\newblock \emph{Bernoulli}, \textbf{27}, pp. 2564--2597.

\bibitem[{Kohler, Krzyzak and Langer(2019)}]{KoKrLa2019}
Kohler, M., Krzyzak, A., and Langer, S. (2019).
\newblock Estimation of a function of low local dimensionality by deep neural
  networks.
\newblock {a}rXiv: 1908.11140.

\bibitem[{Kohler, Krzy{\.z}ak and Walter(2020)}]{KoKrWa2020}
Kohler, M., Krzy{\.z}ak, A., and Walter, B. (2020).
\newblock On the rate of convergence of image classifiers based on
  convolutional neural networks.
\newblock {a}rXiv: 2003.01526.

\bibitem[{Kohler and Langer(2020)}]{KoLa2020}
Kohler, M., and Langer, S. (2020).
\newblock Statistical theory for image classification using deep convolutional
  neural networks with cross-entropy loss.
\newblock {a}rXiv: 2011.13602.

\bibitem[{Kohler and Langer(2021)}]{KoLa2021}
Kohler, M., and Langer, S. (2021).
\newblock On the rate of convergence of fully connected very deep neural
  network regression estimates.
\newblock \emph{{A}nnals of {S}tatistics}, \textbf{49}, pp. 2231--2249.

\bibitem[{Langer(2021)}]{Langer2021}
Langer, S. (2021).
\newblock Analysis of the rate of convergence of fully connected deep
  neuralnetwork regression estimates with smooth activation function.
\newblock \emph{Journal of Multivariate Analysis}, \textbf{182}, p. 104695.

\bibitem[{Larochelle et~al.(2007)Larochelle, Erhan, Courville, Bergstra and
  Bengio}]{Larochelle2007}
Larochelle, H., Erhan, D., Courville, A., Bergstra, J., and Bengio, Y. (2007).
\newblock An empirical evaluation of deep architectures on problems with many
  factors of variation.
\newblock \emph{Proceedings of the 24th International Conference on Machine
  Learning (ICML)}.

\bibitem[{Lin and Zhang(2019)}]{Lin2019}
Lin, S., and Zhang, J. (2019).
\newblock Generalization bounds for convolutional neural networks.
\newblock {a}rXiv: 1910.01487.

\bibitem[{Liu et~al.(2021)Liu, Chen, Zhao and Liao}]{Liu2021}
Liu, H., Chen, M., Zhao, T., and Liao, W. (2021).
\newblock Besov function approximation and binary classification on
  low-dimensional manifolds using convolutional residual networks.
\newblock \emph{Proceedings of the 38th International Conference on Machine
  Learning (PMLR)}, \textbf{139}, pp. 6770--6780.

\bibitem[{Marcos, Volpi and Tuia(2016)}]{Marcos2016}
Marcos, D., Volpi, M., and Tuia, D. (2016).
\newblock Learning rotation invariant convolutional filters for texture
  classification.
\newblock \emph{International Conference on Pattern Recognition (ICPR)}, pp.
  2012--2017.

\bibitem[{Oono and Suzuki(2019)}]{Oono2019}
Oono, K., and Suzuki, T. (2019).
\newblock Approximation and Non-parametric Estimation of ResNet-type
  Convolutional Neural Networks.
\newblock \emph{In International Conference on Machine Learning}, pp.
  4922--4931.

\bibitem[{Petersen and Voigtlaender(2020)}]{Petersen2020}
Petersen, P., and Voigtlaender, F. (2020).
\newblock Equivalence of approximation by convolutional neural networks and
  fully-connected networks.
\newblock \emph{Proceedings of the American Mathematical Society},
  \textbf{148}, pp. 1567--1581.

\bibitem[{Rawat and Wang(2017)}]{Rawat2017}
Rawat, W., and Wang, Z. (2017).
\newblock Deep Convolutional Neural Networks for Image Classification: A
  Comprehensive Review.
\newblock \emph{Neural Computation}, \textbf{29}, pp. 2352--2449.

\bibitem[{Schmidt-Hieber(2020)}]{SchmidtHieber2020}
Schmidt-Hieber, J. (2020).
\newblock Nonparametric regression using deep neural networks with ReLU
  activation function.
\newblock \emph{Annals of Statistics}, \textbf{48}, pp. 1875--1897.

\bibitem[{Suzuki and Nitanda(2019)}]{Suzuki2019}
Suzuki, T., and Nitanda, A. (2019).
\newblock Deep learning is adaptive to intrinsic dimensionality of model
  smoothness in anisotropic Besov space.
\newblock {a}rXiv: 1910.12799.

\bibitem[{Veeling et~al.(2018)Veeling, Linmans, Winkens, Cohen and
  Welling}]{Veeling2018}
Veeling, B.~S., Linmans, J., Winkens, J., Cohen, T., and Welling, M. (2018).
\newblock Rotation Equivariant CNNs for Digital Pathology.
\newblock {a}rXiv: 1806.03962.

\bibitem[{Walter(2021)}]{Wa2021}
Walter, B. (2021).
\newblock Analysis of convolutional neural network image classifiers in a
  hierarchical max-pooling model with additional local pooling.
\newblock {a}rXiv: 2106.05233.

\bibitem[{Wu, Hu and Kong(2015)}]{Wu2015}
Wu, F., Hu, P., and Kong, D. (2015).
\newblock Flip-Rotate-Pooling Convolution and Split Dropout on Convolution
  Neural Networks for Image Classification.
\newblock {a}rXiv: 1507.08754.

\bibitem[{Yarotsky(2018)}]{Yarotsky2018}
Yarotsky, D. (2018).
\newblock Universal approximations of invariant maps by neural networks.
\newblock {a}rXiv: 1804.10306.

\bibitem[{Zhou(2020)}]{Zhou2020}
Zhou, D.-X. (2020).
\newblock Universality of deep convolutional neural networks.
\newblock \emph{Applied and Computational Harmonic Analysis}, \textbf{48}, pp.
  787--794.

\end{thebibliography}
\setcitestyle{authoryear} 

\newpage
\begin{center}
	
	{\LARGE \bf
		Supplementary material to ``Analysis of convolutional neural network image classifiers in a rotationally symmetric model''
	}
\end{center}
\appendix

\noindent
The supplement contains additional material concerning the simulation studies from Section 5, results from the literature used in the proof of Lemma 1 and Theorem 1, the proofs of Lemma 3 and Lemma 4, as well as a bound on the covering number.

\section{Additional material for Section 5}
\subsection{Creating the synthetic image data sets}
\label{seA1}
In order to generate a random image with an appropriate label, we use the Python package \textit{Shapely} to theoretically define a continuous image as follows: 
Firstly, the gray scale value of the background of the image area $C_1$ is set to 1 and for each of the three squares it is randomly (independently) determined whether a quarter is removed or not. The probability that a quarter is removed from a square is given by $p=1-0.5^{1/3}$, which implies that the
class $Y$ of an image is discrete and uniformly distributed on $\{0,1\}$. Secondly, the area, rotation, and gray scale value of each geometric object are determined. The area is determined for each object (independently) by a uniform distribution on the interval $[0.02,0.08]$ for complete squares and on the interval $[0.02,0.06]$ for squares missing a quarter (the second interval is smaller to avoid too large side lengths of these objects). The angle by which an object is rotated is determined (independently) by a uniform distribution on the interval $[0,2\pi]$. The gray scale values of the three objects are determined by randomly permuting the list $(0,1/3,2/3)$ of three gray scale values. Finally, the positions of the objects are determined one after the other as follows: We choose the position of the first object according to a uniform distribution on the restricted image area so that the object is completely within the image area. We repeat the positioning of the second object in the same way until the second object covers only a maximum of five percent of the area of the first object. For the placement of the third object, we use the same method until the third object covers only a maximum of five percent of the area of the first and second object, respectively.
We then use the Python package \textit{Pillow} to discretize the continuous image on $G_{\lambda}$.
\subsection{Rotation by nearest neighbor interpolation}
\label{seA2}
In this section, we define the rotation function $f^{(\alpha)}_{rot}$, which is used in Section 5 for the network architecture $\F_4$. We use a nearest neighbor interpolation here to implement rotation by arbitrary angles for two reasons: Firstly, a nearest neighbor interpolation can be easily implemented using the \textit{Keras backend} library as a layer of a CNN, so the corresponding classifier can be trained using the \textit{Adam} optimizer. Secondly, our theory could be easily extended to such an estimator, since the nearest neighbor interpolation can be traced back to a self-mapping of $G_{\lambda}$ (cf., equation \eqref{seA2eq3} below), which swaps the image positions accordingly, and thus we can obtain a necessary bound for covering number without much effort.

Since we may rotate parts out of the image area by rotating the input image by arbitrary angles, we first introduce a zero padding function $f_{z}:[0,1]^{G_{\lambda}}\rightarrow[0,1]^{G_{\lambda+2\cdot z}}$ that symmetrically adds $z\in\N_0$ rows and columns of zeros on all four sides of the image. %
The output of the function $f_{z}$ is given by
\begin{equation}
	\big(f_{z}(\bx)\big)_{\left(\frac{i-1/2}{\lambda+2\cdot z}-\frac{1}{2},\frac{j-1/2}{\lambda+2\cdot z}-\frac{1}{2}\right)}=
	\begin{cases}
		x_{\left(\frac{i-z-1/2}{\lambda}-\frac{1}{2},\frac{j-z-1/2}{\lambda}-\frac{1}{2}\right)}&,\text{ if }z+1\leq i,j\leq z+\lambda\\
		0&,\text{ elsewhere}
	\end{cases}
\end{equation}
for $i,j\in\{1,\dots,\lambda+2\cdot z\}$. We choose
\begin{equation}
	z_{\lambda}=\left\lceil\frac{\sqrt{2}\cdot\lambda-\lambda}{2}\right\rceil
\end{equation}
to ensure that a rotated version of the image entirely contains the original image. 
To rotate the images by a nearest neighbor interpolation, we define the function
$g^{(\alpha)}:G_{\lambda'}\rightarrow G_{\lambda'}$ that rotates the image positions with a resolution $\lambda'\in\N$ by an angle $\alpha\in[0,2\pi)$.
The output of the function is given by
\begin{equation}
	\label{seA2eq3}
	g^{(\alpha)}(\bv)=\argmin_{\bu\in G_{\lambda'}}\|\bu-rot^{(\alpha)}(\bv)\|_2\quad\big(\bv\in G_{\lambda'}\big),
\end{equation}
where we choose the smallest index in case of ties (we use a bijection which maps $G_{\lambda'}$ to $\{1,\dots,\lambda'^2\}$ to obtain a corresponding order on the indices).
The rotation function $f_{rot}^{(\alpha)}:[0,1]^{G_{\lambda}}\rightarrow[0,1]^{G_{\lambda+2\cdot z_{\lambda}}}$ which rotates an image by the angle $\alpha\in[0,2\pi)$ is then defined by
\[
\big(f_{rot}^{(\alpha)}(\bx)\big)_{\bu}=(f_{z_{\lambda}}(\bx))_{g^{(\alpha)}(\bu)}\quad\big(\bx\in[0,1]^{G_{\lambda}}\big)
\]
for $\bu\in G_{\lambda+2\cdot z_{\lambda}}$.
 \section{Auxiliary results}
In the following section, we present some results from the literature which we have used in the proof of Lemma 1 and Theorem 1.
Our first auxiliary result relates
the misclassification error of our plug-in estimate
to the $L_2$ error of the corresponding least squares estimates.

\begin{lemma}
	\label{le5}
	Define $(g_{\lambda}(\Phi),Y)$, $(g_{\lambda}(\Phi_1),Y_1)$, \dots, $(g_{\lambda}(\Phi_n),Y_n)$, and $\D_n$,
	$\eta$, $f^*$ and $f_n$ as in Section 1.1.
	Then
	\begin{eqnarray*}
		\PROB\{f_n(g_{\lambda}(\Phi))\neq Y\}-\PROB\{f^*(g_{\lambda}(\Phi))\neq Y\}
		&\leq&
		2 \cdot
		\int |\eta_n(x)-\eta(x)| \, \PROB_{g_{\lambda}(\Phi)}(dx)
		\\
		&\leq&
		2 \cdot
		\sqrt{
			\int |\eta_n(x)-\eta(x)|^2  \PROB_{g_{\lambda}(\Phi)}(dx)
		}
	\end{eqnarray*}
	holds.
\end{lemma}

\noindent
{\bf Proof.}
See Theorem 1.1 in \cite{Gyoerfi2002}.
\hfill $\Box$

Our next result
bounds the error of the least squares estimate
via empirical process theory.

\begin{lemma}
	\label{le6}
	Let
	$(X,Y)$, $(X_1,Y_1)$, \dots, $(X_n,Y_n)$
	be independent and identically distributed $\R^d \times \R$-valued
	random variables.
	Assume that the distribution of $(X,Y)$ satisfies
	\begin{align*}
		\E\{\exp(\nconst\cdot Y^2)\} < \infty
	\end{align*}
	for some constant $\const> 0$ and that the regression function
	$m(\cdot)=\EXP\{ Y |X=\cdot \}$
	is bounded in absolute value. Let $\tilde{m}_n$ be the least squares estimate
	\begin{align*}
		\tilde{m}_n(\cdot) = \arg \min_{f \in \mathcal{F}_n} \frac{1}{n} \sum_{i=1}^n |Y_i - f(X_i)|^2
	\end{align*}
	based on some function space $\mathcal{F}_n$
	consisting of functions $f:\R^d \rightarrow \R$
	and set $m_n = T_{\nconst\cdot \log(n)} \tilde{m}_n$ for some constant
	$\const> 0$.
	Then $m_n$ satisfies
	\begin{align*}
		& \mathbf E \int |m_n(x) - m(x)|^2 {\PROB}_X (dx)\notag\\
		&\leq \frac{\nconst\cdot (\log(n))^2 \cdot \sup_{x_1^n \in (\R^d)^n} \left(\log\left(
			\mathcal{N}_1 \left(\frac{1}{n\cdot\mmconst{1}\cdot\log(n)},  T_{c_{4} \log(n)} \mathcal{F}_n, x_1^n\right)
			\right)+1\right)}{n}\notag\\
		&\quad + 2 \cdot \inf_{f \in \mathcal{F}_n} \int |f(x)-m(x)|^2 {\PROB}_X (dx)
	\end{align*}
	for $n > 1$ and some constant $\const> 0$, which does not depend on
	$n$ or the parameters of the estimate.
\end{lemma}

\noindent
{\bf Proof.}
This result follows in a straightforward way from the proof of Theorem 1 in
\cite{Bagirov2009}. A complete proof can be found in the supplement of Bauer and Kohler (2019).
\hfill $\Box$

The next result is an approximation result for
$(p,C)$--smooth
functions by very deep feedforward neural networks.
\begin{lemma}
	\label{le7}
	Let $d \in \N$,
	let $f:\Rd \rightarrow \R$ be $(p,C)$--smooth for some $p=q+s$,
	$q \in \N_0$  and $s \in (0,1]$, and $C>0$. Let $M \in \N$ with $M\geq2$ sufficiently large, where 
	\[M^{2p}\geq
	\nconst\cdot\left(\max\left\{2,\sup_{\substack{\bx\in[-2,2]^d \\
			(l_1,\dots,l_d)\in\N^d \\ l_1+\dots+l_d\leq q}
	}\left|\frac{\partial^{l_1+\dots +l_d}f}{\partial^{l_1}x^{(1)}\dots\partial^{l_d}x^{(d)}}(\bx)\right|\right\}\right)^{4(q+1)}
	\]
	must hold for some sufficiently large constant $\const\geq1$.
	Let $\sigma: \R \to \R$ be the ReLU activation function
	\[
	\sigma(x)= \max\{x,0\}
	\]
	and let $L,r\in\N$ such that
	\begin{enumerate}[label=(\roman*)]
		\item 
		\begin{align*}
			L\geq&5M^d+\left\lceil \log_{4}\left(M^{2p+4\cdot d\cdot(q+1)}\cdot e^{4\dot(q+1)\cdot(M^d-1)}\right) \right\rceil\\
			&\cdot  \lceil \log_2(\max\{d,q\}+2)\rceil
			+\lceil\log_4(M^{2p})\rceil
		\end{align*}
		\item
		\[
		r\geq132\cdot2^d \cdot \lceil e^{d}\rceil\cdot\binom{d+q}{d} \cdot\max\{q+1, d^2\}
		\]
	\end{enumerate}
	hold.
	Then there exists a feedforward neural network 
	\[f_{net}\in\G_d(L,\bk)\]
	with $\bk=(k_1,\dots,k_L)$ and $k_1=\dots=k_L=r$
	such that
	\begin{align*}
		&\sup_{\bx \in [-2,2]^d} | f(\bx)-f_{net}(\bx)|\\
		&\leq
		\nconst
		\cdot\left(\max\left\{2,\sup_{\substack{\bx\in[-2,2]^d \\
				(l_1,\dots,l_d)\in\N^d \\ l_1+\dots+l_d\leq q}
		}\left|\frac{\partial^{l_1+\dots +l_d}f}{\partial^{l_1}x^{(1)}\dots\partial^{l_d}x^{(d)}}(\bx)\right|\right\}\right)^{4(q+1)}
		\cdot M^{-2p}.
	\end{align*}
\end{lemma}

\noindent
{\bf Proof.}
See Theorem 2 b) in \cite{KoLa2021}.
\hfill $\Box$

\section{Proof of Lemma 3 and Lemma 4}
{\bf Proof of Lemma 3.}
Because of inequality \eqref{ple1eq1} it suffices to show that
\begin{align*}
	&\max_{i\in\{1,\dots,t\}}\max_{\bu\in G_{\lambda}~:~\bu+I^{(l)}\subseteq G_{\lambda}}\left|{f}_{l,1}^{(i)}(\bx_{(i,j)+I^{(l)}})-\bar{f}_{l,1}^{(i)}(\bx_{\bu+I^{(l)}})\right|\\
	&
	\leq(C+1)^l\cdot
	\max_{\substack{i\in\{1,\dots,t\},j\in\{1,\dots,4^l\},\\k\in\{1,\dots,l\},s\in\{1,\dots,4^{l-k}\}}}
	\left\{\|g^{(i)}_{0,j}-\bar{g}^{(i)}_{0,j}\|_{[0,1],\infty},\|g_{k,s}^{(i)}-\bar{g}^{(i)}_{k,s}\|_{[0,2]^4,\infty}\right\}.
\end{align*}
This in turn follows from
\begin{equation}
	\label{ple2eq1}
	\begin{split}
		&\left|{f}_{k,s}^{(i)}(\bx)-\bar{f}_{k,s}^{(i)}(\bx)\right|\\
		&\leq (C+1)^{k}\cdot\max_{m\in\{1,\dots,k\},s\in\{1,\dots,4^{l-m}\},j\in\{1,\dots,4^l\}}\left\{\|g_{0,j}^{(i)}-\bar{g}_{0,j}^{(i)}\|_{[0,1],\infty},\|g_{m,s}^{(i)}-\bar{g}^{(i)}_{m,s}\|_{[0,2]^4,\infty}\right\}
	\end{split}
\end{equation}
for all $\bx\in[0,1]^{I^{(k)}}$, $i\in\{1,\dots,t\}$ ,$k\in\{0,\dots,l\}$ and $s\in\{1,\dots,4^{l-k}\}$, which we show by induction on $k$.

For $k=0$, $s\in\{1,\dots,4^l\}$ and $i\in\{1,\dots,t\}$ we have
\begin{align*}
	\left|f_{0,s}^{(i)}(x)-\bar{f}_{0,s}^{(i)}(x)\right|&=\left|g_{0,s}^{(i)}(x)-\bar{g}_{0,s}^{(i)}(x)\right|\leq\left\|g_{0,s}^{(i)}-\bar{g}_{0,s}^{(i)}\right\|_{[0,1],\infty}
\end{align*}
for all $x\in[0,1]$. Assume that equation \eqref{ple2eq1} holds for some $k\in\{0,\dots,l-1\}$. Because of the definition of $\bar{f}^{(i)}_{k,s}$ we have
\[
0\leq\bar{f}^{(i)}_{k,s}(\bx)\leq2
\]
for all $\bx\in[0,1]^{I^{(k)}}$, $i\in\{1,\dots,t\}$, $k\in\{0,\dots,l-1\}$ and $s\in\{1,\dots,4^{l-k}\}$.
Then, the triangle inequality and the Lipschitz assumption on ${g}^{(i)}_{k+1,s}\big|_{[0,2]^2}$ imply
\begin{align*}
	&\left|{f}_{k+1,s}^{(i)}(\bx)-\bar{f}_{k+1,s}^{(i)}(\bx)\right|\\
	&\leq\Big|{g}^{(i)}_{k+1,s}\Big({f}^{(i)}_{k,4\cdot(s-1)+1}(\bx_{\bi^{(i)}_{k,4\cdot(s-1)+1}+I^{(k)}}),{f}^{(i)}_{k,4\cdot(s-1)+2}(\bx_{\bi^{(i)}_{k,4\cdot(s-1)+2}+I^{(k)}}),\\
	&\hspace{2cm}{f}^{(i)}_{k,4\cdot(s-1)+3}(\bx_{\bi^{(i)}_{k,4\cdot(s-1)+3}+I^{(k)}}),{f}^{(i)}_{k,4\cdot s}(\bx_{\bi^{(i)}_{k,4\cdot s}+I^{(k)}})\Big)\\
	&\hspace{0.5cm}-{g}^{(i)}_{k+1,s}\Big(\bar{f}^{(i)}_{k,4\cdot(s-1)+1}(\bx_{\bi^{(i)}_{k,4\cdot(s-1)+1}+I^{(k)}}),\bar{f}^{(i)}_{k,4\cdot(s-1)+2}(\bx_{\bi^{(i)}_{k,4\cdot(s-1)+2}+I^{(k)}}),\\
	&\quad\hspace{2cm}\bar{f}^{(i)}_{k,4\cdot(s-1)+3}(\bx_{\bi^{(i)}_{k,4\cdot(s-1)+3}+I^{(k)}}),\bar{f}^{(i)}_{k,4\cdot s}(\bx_{\bi^{(i)}_{k,4\cdot s}+I^{(k)}})\Big)
	\Big|\\
	&\quad+\Big|{g}^{(i)}_{k+1,s}\Big(\bar{f}^{(i)}_{k,4\cdot(s-1)+1}(\bx_{\bi^{(i)}_{k,4\cdot(s-1)+1}+I^{(k)}}),\bar{f}^{(i)}_{k,4\cdot(s-1)+2}(\bx_{\bi^{(i)}_{k,4\cdot(s-1)+2}+I^{(k)}}),\\
	&\quad\hspace{2cm}\bar{f}^{(i)}_{k,4\cdot(s-1)+3}(\bx_{\bi^{(i)}_{k,4\cdot(s-1)+3}+I^{(k)}}),\bar{f}^{(i)}_{k,4\cdot s}(\bx_{\bi^{(i)}_{k,4\cdot s}+I^{(k)}})\Big)\\
	&\hspace{0.5cm}-\bar{g}^{(i)}_{k+1,s}\Big(\bar{f}^{(i)}_{k,4\cdot(s-1)+1}(\bx_{\bi^{(i)}_{k,4\cdot(s-1)+1}+I^{(k)}}),\bar{f}^{(i)}_{k,4\cdot(s-1)+2}(\bx_{\bi^{(i)}_{k,4\cdot(s-1)+2}+I^{(k)}}),\\
	&\quad\hspace{2cm}\bar{f}^{(i)}_{k,4\cdot(s-1)+3}(\bx_{\bi^{(i)}_{k,4\cdot(s-1)+3}+I^{(k)}}),\bar{f}^{(i)}_{k,4\cdot s}(\bx_{\bi^{(i)}_{k,4\cdot s}+I^{(k)}})\Big)
	\Big|\\
	&\leq C\cdot\max_{j\in\{1,\dots,4\}}\left|{f}^{(i)}_{k,4\cdot(s-1)+m}(\bx_{\bi^{(i)}_{k,4\cdot(s-1)+j}+I^{(k)}})-\bar{f}^{(i)}_{k,4\cdot(s-1)+j}(\bx_{\bi^{(i)}_{k,4\cdot(s-1)+m}+I^{(k)}})\right|\\
	&\quad+\|{g}^{(i)}_{k+1,s}-\bar{g}^{(i)}_{k+1,s}\|_{[0,2]^4,\infty}\\
	&
	\leq
	C\cdot(C+1)^{k}\cdot\max_{m\in\{1,\dots,k\},s\in\{1,\dots,4^{l-m}\},j\in\{1,\dots,4^l\}}\left\{\|g_{0,j}^{(i)}-\bar{g}_{0,j}^{(i)}\|_{[0,1],\infty},\|g_{m,s}^{(i)}-\bar{g}^{(i)}_{m,s}\|_{[0,2]^4,\infty}\right\}\\
	&\quad+\|{g}^{(i)}_{k+1,s}-\bar{g}^{(i)}_{k+1,s}\|_{[0,2]^4,\infty}\\
	&\leq
	(C+1)^{k+1}\cdot\max_{m\in\{1,\dots,k+1\},s\in\{1,\dots,4^{l-m}\},j\in\{1,\dots,4^l\}}\left\{\|g_{0,j}^{(i)}-\bar{g}_{0,j}^{(i)}\|_{[0,1],\infty},\|g_{m,s}^{(i)}-\bar{g}^{(i)}_{m,s}\|_{[0,2]^4,\infty}\right\}
\end{align*}
for all $\bx\in[0,1]^{I^{(k+1)}}$, $i\in\{1,\dots,t\}$ and $s\in\{1,\dots,4^{l-(k+1)}\}$.
\hfill $\Box$

\noindent
In order to prove Lemma 4, we will use the following two auxiliary results.
\begin{lemma}
	\label{le8}
	Let $t\in\N$, set $L_{net}=\lceil\log_2 t\rceil$, set $r_{net}=3\cdot t$ and let $\G_t(L_{net},r_{net})$ be defined as in \eqref{FNNclass}.
	Then there exist $g_{net}\in\G_t(L_{net},r_{net})$ such that
	\[
	g_{net}(\bx)=\max\{x_1,\dots,x_t\}
	\]
	for all $\bx=(x_1,\dots,x_t)\in\R^t$.
\end{lemma}
{\bf Proof.}
W.l.o.g. assume that $t>1$.
In the proof we will use the network
$g_{max}:\R^2\rightarrow\R$ defined by
\[
g_{max}(x_1,x_2)=\sigma(x_2-x_1)+\sigma(x_1)-\sigma(-x_1)\quad(x_1,x_2\in\R)
\]
which satisfies
\begin{align*}
	g_{max}(x_1,x_2)&=\max\{x_2-x_1,0\}+\underbrace{\max\{x_1,0\}-\max\{-x_1,0\}}_{=x_1}=\max\{x_1,x_2\}
\end{align*}
for all $x_1,x_2\in\R$. For $t\in\N\setminus\{1\}$ we set
\[
{r}(t)=3\cdot2^{\lceil\log_2(t)\rceil-1}\quad\text{and}\quad L(t)=\lceil\log_2 t\rceil
\]
and show the assertion by showing the more powerful assertion that for all $t\in\N\setminus\{1\}$ there exists 
\[
g_{net}\in\G_t(L_{net},r(t))\stackrel{r(t)<r_{net}}{\subset}\G_t(L_{net},r_{net})
\]
such that
\[
g_{net}(\bx)=\max\{x_1,\dots,x_t\}
\]
for all $\bx\in\R^t$. We show this by induction on $t$. 

For $t=2$ the assertion follows by using the network $g_{max}$. 
Now let $t>2$ and assume the assertion holds for all natural numbers less than $t$ and greater than one.
Then there exist $g\in\G_{\lceil t/2\rceil}(L(\lceil t/2\rceil),r(\lceil t/2\rceil))$ %
such that
\[
g(\bx)=\max\{x_1,\dots,x_{\lceil t/2\rceil}\}
\]
for all $\bx\in\R^{\lceil t/2\rceil}$.
We then define $g_{net}\in\G_{t}(L(\lceil t/2\rceil)+1,2\cdot r(\lceil t/2\rceil))$ by
\begin{align*}
	g_{net}(\bx)&=g_{max}(g(x_1,\dots,x_{\lceil t/2\rceil}),g(x_{\lfloor t/2\rfloor+1},\dots,x_t))
	=\max\{x_1,\dots,x_t\}.
\end{align*}
It is now sufficient to show that
\[
L(t)=L(\lceil t/2\rceil)+1\quad\text{and}\quad r(t)=2\cdot r(\lceil t/2\rceil)).
\]
Since $2^k<t\leq2^{k+1}$ for some $k\in\N$ we have
\[
\lceil\log_2(2\cdot\lceil t/2\rceil)\rceil\geq\lceil\log_2(t)\rceil=k+1=\left\lceil\log_2\left(2\cdot2^{k}\right)\right\rceil\geq\lceil\log_2(2\cdot\lceil t/2\rceil)\rceil
\]
which implies
\begin{equation}
	\label{dd}
	\lceil\log_2(2\cdot\lceil t/2\rceil)\rceil=\lceil\log_2(t)\rceil.
\end{equation}
By using equation \eqref{dd} we get
\begin{align*}
	L(\lceil t/2\rceil)+1&=\lceil\log_2\lceil t/2\rceil\rceil+1=\lceil\log_2(2\cdot\lceil t/2\rceil)\rceil=\lceil\log_2 t\rceil
	=L(t)
\end{align*}
and
\begin{align*}
	2\cdot r(\lceil t/2\rceil))&
	=2\cdot3\cdot2^{\lceil\log_2(\lceil t/2\rceil)\rceil-1}\\
	&=3\cdot2^{\lceil\log_2(\lceil t/2\rceil)\rceil+1-1}\\
	&=3\cdot2^{\lceil\log_2(2\cdot\lceil t/2\rceil)\rceil-1}\\
	&=3\cdot2^{\lceil\log_2(t)\rceil-1}\\
	&=r(t).
\end{align*}
\hfill $\Box$

\noindent
The next lemma allows us to compute the standard feedforward neural networks \linebreak$\sigma\circ g_{net,k,s}^{(i)}$ from Lemma 4 within a convolutional neural network. Since the input dimension of the standard feedforward neural networks is $d=1$ for $k=0$ and $d=4$ for $k\in\{1,\dots,l\}$ we consider the general case $d\in\N$.
\begin{lemma}
	\label{le9}
	Let $d\in\N$ and $g_{net}\in\G_d(L_{net},r_{net})$ for some $L_{net},r_{net}\in\N$.
	Let
	$\sigma(x)=\max\{x,0\}$ be the ReLU activation function.
	We assume that there is given a convolutional neural network
	$f_{CNN}\in\F^{CNN}_{L,\bk,\bM,B}$
	with $L=r_0+L_{net}+1$ convolutional layers and $k_r=t+r_{net}$ channels in the convolutional layer $r$ $(r=1,\dots,r_0+L_{net}+1)$ for $t\in\N$ and $r_0\in\N_0$, and filter sizes $M_1,\dots,M_{r_0+L_{net}+1}\in\N$ with $M_{r_0+1}=\IND_{\{k>0\}}\cdot2^{k}+3$ for some $k\in\N_{0}$. Let
	\[
	(i_1,j_1),\dots,(i_d,j_d)\in\{-\lfloor2^{k-1}+1\rfloor,\dots,0,\dots,\lfloor2^{k-1}+1\rfloor\}^2,
	\]
	${s}_{0}\in\{1,\dots,t\}$ and $s_1,\dots,s_d\in\{1,\dots,k_{r_0}\}$.
	The convolutional neural network $f_{CNN}$ is given by its weight matrix
	\begin{equation}
		\label{le6eq1}
		\bw
		=
		\left(
		w_{i',j',s,s'}^{(r)}
		\right)_{
			1 \leq i',j' \leq M_r, s \in \{1, \dots, k_{r-1}\}, s' \in \{1, \dots, k_r\}
			r \in \{1, \dots,r_0+L_{net}+1 \}
		},
	\end{equation}
	and its bias weights
	\begin{equation}
		\label{le6eq2}
		\bw_{bias}
		=
		\left(
		w_{s'}^{(r)}
		\right)_{
			s' \in \{1, \dots, k_r\},
			r \in \{1, \dots,r_0+L_{net}+1\}
		}.
	\end{equation}	
	Then we are able to modify the weights \eqref{le6eq1} and \eqref{le6eq2}
	\begin{equation}
		\label{le6eq7}
		w_{t_1,t_2,s,s'}^{(r)}, w_{s'}^{(r)} \quad (s\in\{1,\dots,t+r_{net}\})
	\end{equation}
	in layers $r  \in \{r_0+1, \dots, r_0+L_{net} +1 \}$ and in channels $s'\in\{{s}_{0},t+1,\dots,t+r_{net}\}$ 
	such that
	\begin{equation}
		\label{le2eq1}
		\begin{split}
			&
			o^{(r_0+L_{net}+1)}_{(i',j'),{s}_{0}}
			=
			\sigma\Big(g_{net} \Big(
			o^{(r_0)}_{(i'+i_1,j'+j_1),s_1},o^{(r_0)}_{(i'+i_2,j'+j_2),s_2},
			\dots,
			o^{(r_0)}_{(i'+i_d,j'+j_d),s_d}\Big)\Big)
		\end{split}
	\end{equation}
	for all $(i',j')\in\{1,\dots,\lambda\}^2$, where we set $o^{(r_0)}_{(i',j'),s}=0$ for $(i',j')\notin\{1,\dots,\lambda\}^2$.
	
\end{lemma}
{\bf Proof.}
We assume that the standard feedforward neural network $g_{net}$ is given by
\[
g_{net}(\bx) = \sum_{i=1}^{r_{net}} w_{1,i}^{(L_{net})}g_i^{(L_{net})}(\bx) + w_{1,0}^{(L_{net})},
\]
where $g_i^{(L_{net})}$ is recursively defined by
\[
g_i^{(r)}(\bx) = \sigma\left(\sum_{j=1}^{r_{net}} w_{i,j}^{(r-1)} g_j^{(r-1)}(\bx) + w_{i,0}^{(r-1)} \right)
\]
for
$i \in \{1,\dots,r_{net}\}$,
$r \in \{2, \dots, L_{net}\}$,
and
\[
g_i^{(1)}(\bx) = \sigma \left(\sum_{j=1}^d w_{i,j}^{(0)} x^{(j)} +
w_{i,0}^{(0)} \right)
\quad (i \in \{1, \dots, r_{net}\}).
\]
W.l.o.g. we can assume that $(s_n,i_n,j_n)\neq(s_{m},i_{m},j_{m})$ for distinct $n,m\in\{1,\dots,d\}$ (otherwise one can show the assertion for a accordingly defined $g'_{net}\in\G_{d'}(L_{net},r_{net})$ with $d'<d$).
Since $M_{r_0+1}=2\cdot\lfloor2^{k-1}\rfloor+3$ and $\lceil M_{r_0+1}/2\rceil=\lfloor2^{k-1}\rfloor+2$ we have
\begin{align}
	\label{ple10eq1}
	\begin{split}
		&o^{(r_0+1)}_{(i',j'),t+i}\\
		&=
		\sigma \left(
		\sum_{s=1}^{k_{r_0}}
		\sum_{\substack{t_1,t_2 \in \{1, \dots, M_{r_0+1}\}\\i'+t_1-\lceil M_{r_0+1}/2\rceil\in\{1,\dots,\lambda\}\\j'+t_2-\lceil M_{r_0+1}/2\rceil\in\{1,\dots,\lambda\}}}
		w_{t_1,t_2,s,t+i}^{(r_0+1)}
		\cdot
		o_{(i'+t_1-\lceil M_{r_0+1}/2\rceil,j'+t_2-\lceil M_{r_0+1}/2\rceil),s}^{(r_0)}
		+
		w_{t+i}^{(r_0+1)}
		\right)\\
		&=
		\sigma \left(
		\sum_{s=1}^{k_{r_0}}
		\sum_{\substack{t_1,t_2 \in \{-\lfloor2^{k-1}+1\rfloor, \dots,\lfloor2^{k-1}+1\rfloor\}\\(i'+t_1,j'+t_2)\in\{1,\dots,\lambda\}^2}}
		w_{\lfloor2^{k-1}\rfloor+2+t_1,\lfloor2^{k-1}\rfloor+2+t_2,s,t+i}^{(r_0+1)}
		\cdot
		o_{(i'+t_1,j'+t_2),s}^{(r_0)}
		+
		w_{t+i}^{(r_0+1)}
		\right)
	\end{split}
\end{align}
for all $i\in\{1,\dots,r_{net}\}$ and $(i',j')\in\{1,\dots,\lambda\}^2$.
We aim to choose the weights in \eqref{ple10eq1} such that
\begin{equation*}
	\begin{split}
		o^{(r_0+1)}_{(i',j'),t+i}&=
		\sigma \Bigg(
		\sum_{n=1}^{d}
		w_{i,n}^{(0)}
		\cdot
		o^{(r_0)}_{(i'+i_n,j'+j_n),s_n}
		+
		w_{i,0}^{(0)}
		\Bigg)\\
		&=g_{i}^{(1)}
		\Big(
		o^{(r_0)}_{(i'+i_1,j'+j_1),s_1},o^{(r_0)}_{(i'+i_2,j'+j_2),s_2},
		\dots,
		o^{(r_0)}_{(i'+i_d,j'+j_d),s_d}
		\Big)
	\end{split}
\end{equation*}
for all $i\in\{1,\dots,r_{net}\}$ and $(i',j')\in\{1,\dots,\lambda\}^2$.
Therefore we choose the only non-zero weights by
\[
w_{\lfloor2^{k-1}\rfloor+2+i_n,\lfloor2^{k-1}\rfloor+2+j_n,s_{n},t+i}^{(r_0+1)}=w_{i,n}^{(0)}\quad\text{and}\quad w_{t+i}^{(r_0+1)}=w_{i,0}^{(0)}
\]
for $n\in\{1,\dots,d\}$ and $i\in\{1,\dots,r_{net}\}$ and obtain
\begin{equation}
	\label{ple10eq2}
	\begin{split}
		o^{(r_0+1)}_{(i',j'),t+i}&=
		\sigma \Bigg(
		\sum_{n=1}^{d}
		w_{i,n}^{(0)}
		\cdot
		o^{(r_0)}_{(i'+i_n,j'+j_n),s_n}
		+
		w_{i,0}^{(0)}
		\Bigg)\\
		&=g_{i}^{(1)}
		\Big(
		o^{(r_0)}_{(i'+i_1,j'+j_1),s_1},o^{(r_0)}_{(i'+i_2,j'+j_2),s_2},
		\dots,
		o^{(r_0)}_{(i'+i_d,j'+j_d),s_d}
		\Big)
	\end{split}
\end{equation}
for all $i\in\{1,\dots,r_{net}\}$ and $(i',j')\in\{1,\dots,\lambda\}^2$. For the following layers we have
\begin{equation*}
	\begin{split}
			o^{(r_0+r)}_{(i',j'),t+i}&=
			\sigma \Bigg(
			\sum_{s=1}^{k_{r_0+r-1}}
			\sum_{\substack{t_1,t_2 \in \{1, \dots, M_{r_0+r}\}\\i'+t_1-\lceil M_{r_0+r}/2\rceil\in\{1,\dots,\lambda\}\\j'+t_2-\lceil M_{r_0+r}/2\rceil\in\{1,\dots,\lambda\}}}
			\\
			&\hspace{1.5cm}
			w_{t_1,t_2,s,t+i}^{(r_0+r)}
			\cdot
			o_{(i'+t_1-\lceil M_{r_0+r}/2\rceil,j'+t_2-\lceil M_{r_0+r}/2\rceil),s}^{(r_0+r-1)}
			+
			w_{t+i}^{(r_0+r)}
			\Bigg)\\
		&=
		\sigma \Bigg(
		\sum_{s=1}^{k_{r_0+r-1}}
		\sum_{\substack{t_1,t_2 \in \{1-\lceil M_{r_0+r}/2\rceil, \dots,M_{r_0+r}-\lceil M_{r_0+r}/2\rceil\}\\(i'+t_1,j'+t_2)\in\{1,\dots,\lambda\}^2}}\\
		&\hspace{1.5cm}
		w_{\lceil M_{r_0+r}/2\rceil+t_1,\lceil M_{r_0+r}/2\rceil+t_2,s,t+i}^{(r_0+r)}
		\cdot
		o_{(i'+t_1,j'+t_2),s}^{(r_0+r-1)}
		+
		w_{t+i}^{(r_0+r)}
		\Bigg)
	\end{split}
\end{equation*}
for $r \in \{2, \dots,L_{net}\}$, $i\in\{1,\dots,r_{net}\}$ and $(i',j')\in\{1,\dots,\lambda\}^2$. Here we aim to choose the weights %
such that
\begin{equation}
	\label{ple10eq3}
	o^{(r_0+r)}_{(i',j'),t+i}=\sigma\Bigg(
	\sum_{j=1}^{r_{net}}w_{i,j}^{(r-1)}\cdot o^{(r_0+r-1)}_{(i',j'),t+j}+w_{i,0}^{(r-1)}\Bigg)
\end{equation}
for all $r \in \{2, \dots,L_{net}\}$, $i\in\{1,\dots,r_{net}\}$ and $(i',j')\in\{1,\dots,\lambda\}^2$.
Therefore we choose the only nonzero weights by
\[
w_{\lceil M_{r_0+r}/2\rceil,\lceil M_{r_0+r}/2\rceil,t+j,t+i}^{(r_0+r)}=w_{i,j}^{(r-1)}\quad\text{and}\quad w_{t+i}^{(r_0+r)}=w_{i,0}^{(r-1)}
\]
for $r \in \{2, \dots,L_{net}\}$, $i\in\{1,\dots,r_{net}\}$ and $j \in  \{1, \dots, r_{net}\}$ which implies equation \eqref{ple10eq3}.
In layer $r=r_0+L_{net}+1$ we have
\begin{equation*}
	\begin{split}
		o^{(r_0+L_{net}+1)}_{(i',j'),s_0}&=
		\sigma \Bigg(
		\sum_{s=1}^{k_{r-1}}
		\sum_{\substack{t_1,t_2 \in \{1-\lceil M_{r_0+L_{net}+1}/2\rceil,\dots,M_{r_0+L_{net}+1}-\lceil M_{r_0+L_{net}+1}/2\rceil\}\\(i'+t_1,j'+t_2)\in\{1,\dots,\lambda\}^2}}
		\\&\hspace{1cm}
		w_{\lceil M_{r_0+L_{net}+1}/2\rceil+t_1,\lceil M_{r_0+L_{net}+1}/2\rceil+t_2,s,s_0}^{(r_0+L_{net}+1)}
		\cdot
		o_{(i'+t_1,j'+t_2),s}^{(r_0+L_{net})}
		+
		w_{s_0}^{(r_0+L_{net}+1)}
		\Bigg)
	\end{split}
\end{equation*}
for $(i',j')\in\{1,\dots,\lambda\}^2$ and want to choose the weights such that
\begin{equation}
	\label{ple10eq4}
	o^{(r_0+L_{net}+1)}_{(i',j'),s_0}
	=
	\sigma \left(
	\sum_{i=1}^{r_{net}} w_{1,i}^{(L_{net})}
	\cdot
	o^{(r_0+L_{net})}_{(i',j'),t+i}
	+ w_{1,0}^{(L_{net})}
	\right)
\end{equation}
for all $(i',j')\in\{1,\dots,\lambda\}^2$. For this purpose we choose the only nonzero weights by
\[
w_{\lceil M_{r_0+L_{net}+1}/2\rceil,\lceil M_{r_0+L_{net}+1}/2\rceil,t+i,s_0}^{(r_0+L_{net}+1)}=w_{1,i}^{(L_{net})}\quad\mbox{ and }\quad w_{{s}_0}^{(r_0+L_{net}+1)}=w_{1,0}^{(L_{net})}
\]
for $i\in  \{1, \dots, r_{net}\}$ which implies equation \eqref{ple10eq4}. Combining equations \eqref{ple10eq2}, \eqref{ple10eq3} and \eqref{ple10eq4} then yields the assertion.
\hfill $\Box$

\noindent
{\bf Proof of Lemma 4.}
In the proof we use that for $x\geq0$ we have
\[
\sigma(x)=\max\{x,0\}=x
\]
which enables us to propagate a nonnegative value computed in a layer of a convolutional neural network in channel $s'$ at position $(i',j')$ to the next convolutional layer by
\begin{equation}
	\label{ple3eq2}
	o^{(r)}_{(i',j'),s''}=\sigma\left(o^{(r-1)}_{(i',j'),s'}\right)=o^{(r-1)}_{(i',j'),s'}
\end{equation}	
with corresponding weights in the $r-$th layer in channel $s''$ which are choosen accordingly from the set $\{0,1\}$. 

Firstly,  let $g_{max}\in\G_{t}(L_{t},r_{t})$ be the neural netork from Lemma \ref{le8} such that
\begin{align*}
	\bar{\eta}(\bx)&=\max_{\bu\in G_{\lambda}~:~\bu+I^{(l)}\subseteq G_{\lambda}}\max_{i\in\{1,\dots,t\}}\bar{f}_{l,1}^{(i)}(\bx_{\bu+I^{(l)}})\\
	&=\max_{i\in\{1,\dots,t\}}\max_{\bu\in G_{\lambda}~:~\bu+I^{(l)}\subseteq G_{\lambda}}\bar{f}_{l,1}^{(i)}(\bx_{\bu+I^{(l)}})\\
	&=g_{\max}\left(\max_{\bu\in G_{\lambda}~:~\bu+I^{(l)}\subseteq G_{\lambda}}\bar{f}_{l,1}^{(1)}(\bx_{\bu+I^{(l)}}),\dots,\max_{\bu\in G_{\lambda}~:~\bu+I^{(l)}\subseteq G_{\lambda}}\bar{f}_{l,1}^{(t)}(\bx_{\bu+I^{(l)}})\right)
\end{align*}
for all $\bx\in[0,1]^{G_{\lambda}}$.
Because of the definition of the function class $\F_{\btheta}^{CNN}$, it is thus sufficient to show
that for all $i\in\{1,\dots,t\}$ there exists
$f_{i}\in\F_{L,\bk,\bM,B}^{CNN}$ such that
\begin{equation}
	\label{ple3eq5}
	f_{i}(\bx)=\max_{\bu\in G_{\lambda}~:~\bu+I^{(l)}\subseteq G_{\lambda}}\bar{f}_{l,1}^{(i)}(\bx_{\bu+I^{(l)}})
\end{equation}
for all $\bx\in[0,1]^{G_{\lambda}}$. Therefore, in the remaining of the proof let $i\in\{1,\dots,t\}$ be fixed.
The idea is to successively compute the outputs of the functions
\[
\bar{f}_{0,1}^{(i)},\dots,\bar{f}_{0,4^l}^{(i)},\dots,\bar{f}_{k,1}^{(i)},\dots,\bar{f}_{k,4^{l-k}}^{(i)},\dots,\bar{f}_{l-1,1}^{(i)},\dots,\bar{f}_{l-1,4}^{(i)},\bar{f}_{l,1}^{(i)}
\]
of the discretized hierarchical model $\bar{f}_{l,1}^{(i)}$ by computing the functions $\{\bar{g}_{k,s}^{(i)}\}$ by repeatedly applying Lemma \ref{le9}, where for $k=0$ we apply Lemma \ref{le9} with $d=1$ and for $k=1,\dots,l$ we use $d=4$. We store the outputs of the functions $\bar{f}^{(i)}_{k,s}(\bx_{\bu+I^{(k)}})$ by the above idea of equation \eqref{ple3eq2} in corresponding channels, so that we can use the outputs severals times. For the computation of the maximum in equation \eqref{ple3eq5} we will finally use the global max-pooling layers of our CNN architecture (cf., equation (7)).

A convolutional neural network $f_{i}\in\F_{L,\bk,\bM,B}^{CNN}$ is of the form
\[
f_{i}(\bx)=
\max \Bigg\{
\sum_{s''=1}^{k_L}
w_{s''}\cdot
o_{(i',j'),s''}^{(L)} \,
: \,
(i',j')\in\{1+B,\dots,\lambda-B\}^2
\Bigg\},
\]
with the weight vector
\[
\bw
=
\left(
w_{i',j',s',s''}^{(r)}
\right)_{
	1 \leq i',j' \leq M_r, s' \in \{1, \dots, k_{r-1}\}, s'' \in \{1, \dots, k_r\},
	r \in \{1, \dots,L \}
},
\]
bias weights
\[
\bw_{bias}
=
\left(
w_{s''}^{(r)}
\right)_{
	s'' \in \{1, \dots, k_r\},
	r \in \{1, \dots,L\}
},
\]
and the output weights
\[
\bw_{out}=\big(w_{s}\big)_{s\in\{1,\dots,k_L\}}.
\]

{\it In the first step} we show how to choose the weight vector $\bw$ and the bias weights $\bw_{bias}$ such that
\begin{equation}
	\label{ple3eq8}
	o_{(i',j'),1}^{(L)}=
	\bar{f}_{l,1}^{(i)}(\bx_{\big(\frac{i'-1/2}{\lambda}-\frac{1}{2},\frac{j'-1/2}{\lambda}-\frac{1}{2}\big)+I^{(l)}})
\end{equation}
for all $(i',j')\in\{2^{l-1}+l,\dots,\lambda-2^{l-1}-(l-1)\}^2$.
For $k=0,\dots,l$ we set 
\[
r(k)=\sum_{m=0}^{k}4^{l-m}\cdot(L_{net}+1)
\]
and
show equation \eqref{ple3eq8} by showing via induction on $k$ that
\begin{equation}
	\label{ple3eq1}
	o_{(i',j'),s}^{(r(k))}=\bar{f}_{k,s}^{(i)}\Big(\bx_{\big(\frac{i'-1/2}{\lambda}-\frac{1}{2},\frac{j'-1/2}{\lambda}-\frac{1}{2}\big)+I^{(k)}}\Big)
\end{equation}
for all $(i',j')\in\{\lceil2^{k-1}\rceil+k,\dots,\lambda-\lceil2^{k-1}\rceil-(k-1)\}^2$, $k\in\{0,\dots,l\}$ and $s\in\{1,\dots,4^{l-k}\}$.

We start with $k=0$ and show that
\begin{align*}
	o_{(i',j'),s}^{(r(0))}&=\sigma\left(g^{(i)}_{net,0,s}\left(x_{\frac{i'-1/2}{\lambda}-\frac{1}{2},\frac{j'-1/2}{\lambda}-\frac{1}{2}}\right)\right)\\
	&=\sigma\left(g^{(i)}_{net,0,s}\left(o^{(0)}_{(i',j'),1}\right)\right)
\end{align*}
for all $(i',j')\in\{1,\dots,\lambda\}^2$ and $s\in\{1,\dots,4^l\}$.
The idea is to successively use Lemma \ref{le9} for the computation for each network
\begin{equation}
	\label{ple3eq4}
	\left\{\sigma\left(g^{(i)}_{net,0,s}\left(o^{(0)}_{(i',j'),1}\right)\right)\right\}_{(i',j')\in\{1,\dots,\lambda\}^2}
\end{equation}
for $s\in\{1,\dots,4^l\}$
and store the computed values in the corresponding channels
\[
1,\dots,4^{l}
\]
using equation \eqref{ple3eq2}.
Before we apply Lemma \ref{le9}, we choose the weights in channel
\[
4^{l}+1
\]
as in equation \eqref{ple3eq2} such that 
\[
o_{(i',j'),4^{l}+1}^{(r)}=o_{(i',j'),1}^{(0)}%
\]
for all 
$
r\in\left\{1,\dots,r(0)\right\}
$ and
$(i',j')\in\{1,\dots,\lambda\}^2$.
Next, let us specify how to use Lemma \ref{le9}. We first note that
\[
M_{1},\dots,M_{r(0)}=3.
\]
Now, by using Lemma \ref{le9} with parameters $d=1$,
\[
s_1=
\begin{cases}
	1&,\text{ if }s=1\\
	4^l+1&,\text{ elsewhere}\\
\end{cases}
\]
${s}_0=s$, and $r_0=(s-1)\cdot(L_{net}+1)$ we can calculate the values \eqref{ple3eq4} in layers
\[
r_0+1,\dots,r_0+L_{net}+1
\]
by choosing corresponding weights in channels
\[
s,5\cdot4^{l-1}+1,\dots,5\cdot4^{l-1}+r_{net}
\]
such that we have
\begin{align*}
	o^{(s\cdot(L_{net}+1))}_{(i',j'),s}&=\sigma\left(g^{(i)}_{net,0,s}\left(o^{(0)}_{(i',j'),1}\right)\right)
\end{align*}
for all $(i',j')\in\{1,\dots,\lambda\}^2$ and $s\in\{1,\dots,4^l\}$.
Once a value has been computed in layer $s\cdot(L_{net}+1)$ for $s\in\{1,\dots,4^{l}\}$, it will be propagated to the next layer using equation \eqref{ple3eq2} such that we have
\begin{align*}
	o^{(r(0))}_{(i',j'),s}&=\sigma\left(g^{(i)}_{net,0,s}\left(o^{(0)}_{(i',j'),1}\right)\right)	
\end{align*}
for all $(i',j')\in\{1,\dots,\lambda\}^2$ and $s\in\{1,\dots,4^l\}$, which imply that equation \eqref{ple3eq1} holds for $k=0$.

Now assume that property \eqref{ple3eq1} is true for some $k\in\{0,\dots,l-1\}$ and show that property \eqref{ple3eq1} holds for $k+1$ by choosing the corresponding weights in layers 
\[
r(k)+1,\dots,r(k+1)
\]
such that
\begin{equation}
	\label{ple3eq6}
	\begin{split}
		o_{(i',j'),s}^{(r(k+1))}&=\sigma\Big(g^{(i)}_{net,k+1,s}\Big(\bar{f}^{(i)}_{k,4\cdot(s-1)+1}\Big(\bx_{\big(\frac{i'-1/2}{\lambda}-\frac{1}{2},\frac{j'-1/2}{\lambda}-\frac{1}{2}\big)+\bi_{k,4\cdot(s-1)+1}+I^{(k)}}\Big),
		\\
		&\hspace{2.6cm}
		\bar{f}^{(i)}_{k,4\cdot(s-1)+2}\Big(\bx_{\big(\frac{i'-1/2}{\lambda}-\frac{1}{2},\frac{j'-1/2}{\lambda}-\frac{1}{2}\big)+\bi_{k,4\cdot(s-1)+2}+I^{(k)}}\Big),
		\\
		&\hspace{2.6cm}
		\bar{f}^{(i)}_{k,4\cdot(s-1)+3}\Big(\bx_{\big(\frac{i'-1/2}{\lambda}-\frac{1}{2},\frac{j'-1/2}{\lambda}-\frac{1}{2}\big)+\bi_{k,4\cdot(s-1)+3}+I^{(k)}}\Big),
		\\
		&\hspace{2.6cm}
		\bar{f}^{(i)}_{k,4\cdot s}\Big(\bx_{\big(\frac{i'-1/2}{\lambda}-\frac{1}{2},\frac{j'-1/2}{\lambda}-\frac{1}{2}\big)+\bi_{k,4\cdot s}+I^{(k)}}\Big)
		\Big)\Big)
	\end{split}
\end{equation}
for all $\bx\in[0,1]^{G_{\lambda}}$, $(i',j')\in\{2^{k}+k+2,\dots,\lambda-2^{k}-k-1\}^2$ and $s\in\{1,\dots,4^{l-(k+1)}\}$.
Since
\[
\bi_{k,s}\in\left\{-\frac{\lfloor2^{k-1}\rfloor+1}{\lambda},\dots,0,\dots,\frac{\lfloor2^{k-1}\rfloor+1}{\lambda}\right\}^2
\]
for all $s\in\{1,\dots,4^{l-k}\}$
we have
\[
(i',j')+\lambda\cdot\bi_{k,s}\in\{\lceil2^{k-1}\rceil+k,\dots,\lambda-\lceil2^{k-1}\rceil-(k-1)\}^2
\]
for all $(i',j')\in\{2^{k}+k+1,\dots,\lambda-2^{k}-k\}^2$ and $s\in\{1,\dots,4^{l-k}\}$.
Because of the induction hypothesis
equation \eqref{ple3eq6} then is equivalent to 
\begin{equation*}
	\label{ple3eq3}
	\begin{split}
		o_{(i',j'),s}^{(r(k+1))}&=\sigma\Big(g^{(i)}_{net,k+1,s}\Big(
		o_{(i',j')+\lambda\cdot\bi_{k,4\cdot(s-1)+1},s}^{(r(k))},
		o_{(i',j')+\lambda\cdot\bi_{k,4\cdot(s-1)+2},s}^{(r(k))},
		\\
		&\hspace{3cm}
		o_{(i',j')+\lambda\cdot\bi_{k,4\cdot(s-1)+3},s}^{(r(k))},
		o_{(i',j')+\lambda\cdot\bi_{k,4\cdot s},s}^{(r(k))}		
		\Big)\Big).
	\end{split}
\end{equation*}
Analogous to the induction base case, the idea is to successively use Lemma \ref{le9} for the computation of each network
\begin{equation}
	\label{ple3eq7}
	\begin{split}
		&\sigma\Big(g^{(i)}_{net,k+1,s}\Big(
		o_{(i',j')+\lambda\cdot\bi_{k,4\cdot(s-1)+1},s}^{(r(k))},
		o_{(i',j')+\lambda\cdot\bi_{k,4\cdot(s-1)+2},s}^{(r(k))},
		\\
		&\hspace{3cm}
		o_{(i',j')+\lambda\cdot\bi_{k,4\cdot(s-1)+3},s}^{(r(k))},
		o_{(i',j')+\lambda\cdot\bi_{k,4\cdot s},s}^{(r(k))}		
		\Big)\Big)
	\end{split}
\end{equation}
for $s\in\{1,\dots,4^{l-(k+1)}\}$ and store the computed values in the corresponding channels
\[
1,\dots,4^{l-(k+1)}
\]
using equation \eqref{ple3eq2}.
Before we apply Lemma \ref{le9}, we choose the weights in channels
\[
4^{l-(k+1)}+1,\dots,4^{l-(k+1)}+4^{l-k}
\]
such that 
\[
o_{(i',j'),4^{l-(k+1)}+s}^{(r)}=o_{(i',j'),s}^{(r(k))}%
\]
for all 
$
r\in\left\{r(k)+1,\dots,r(k+1)\right\}
$,
$(i',j')\in\{1,\dots,\lambda\}^2$ and $s=1,\dots,4^{l-k}$ by another application of equation \eqref{ple3eq2}.
Next, let us specify how to use Lemma 9. We first note that
\[
M_{r(k)+1},\dots,M_{r(k+1)}=2\cdot\lfloor2^{k-1}\rfloor+3.
\]
Now, by using Lemma \ref{le9} for $s\in\{1,\dots,4^{l-(k+1)}\}$ with parameters $d=4$,
\[
s_m=
\begin{cases}
	4\cdot(s-1)+m&,\text{ if }s=1\\
	4^{l-(k+1)}+4\cdot(s-1)+m&,\text{ elsewhere}\\
\end{cases}
\]
for $m=1,\dots,4$, $\tilde{s}=s$, and $r_0=r(k)+(s-1)\cdot(L_{net}+1)$ we can calculate the values \eqref{ple3eq7} in layers
\[
r_0+1,\dots,r_0+L_{net}+1
\]
by choosing corresponding weights in channels
\[
s,5\cdot4^{l-1}+1,\dots,5\cdot4^{l-1}+r_{net}
\]
such that we have
\begin{align*}
	o^{(r(k)+s\cdot(L_{net}+1))}_{(i',j'),s}&=\sigma\Big(g^{(i)}_{net,k+1,s}\Big(
	o_{(i',j')+\lambda\cdot\bi_{k,4\cdot(s-1)+1},s}^{(r(k))},
	o_{(i',j')+\lambda\cdot\bi_{k,4\cdot(s-1)+2},s}^{(r(k))},
	\\
	&\hspace{3cm}
	o_{(i',j')+\lambda\cdot\bi_{k,4\cdot(s-1)+3},s}^{(r(k))},
	o_{(i',j')+\lambda\cdot\bi_{k,4\cdot s},s}^{(r(k))}		
	\Big)\Big)	
\end{align*}
for all $(i',j')\in\{2^{k}+k+2,\dots,\lambda-2^{k}-k-1\}^2$ and $s\in\{1,\dots,4^{l-(k+1)}\}$.
Once a value has been saved in layer $r(k)+s\cdot(L_{net}+1)$ for $s\in\{1,\dots,4^{l-(k+1)}\}$, it will be propagated to the next layer using equation \eqref{ple3eq2} such that we have
\begin{align*}
	o^{(r(k+1))}_{(i',j'),s}&=\sigma\Big(g^{(i)}_{net,k+1,s}\Big(
	o_{(i',j')+\lambda\cdot\bi_{k,4\cdot(s-1)+1},s}^{(r(k))},
	o_{(i',j')+\lambda\cdot\bi_{k,4\cdot(s-1)+2},s}^{(r(k))},
	\\
	&\hspace{3cm}
	o_{(i',j')+\lambda\cdot\bi_{k,4\cdot(s-1)+3},s}^{(r(k))},
	o_{(i',j')+\lambda\cdot\bi_{k,4\cdot s},s}^{(r(k))}		
	\Big)\Big)	
\end{align*}
for all $(i',j')\in\{2^{k}+k+2,\dots,\lambda-2^{k}-k-1\}^2$ and $s\in\{1,\dots,4^{l-(k+1)}\}$, which concludes the first step.

{\it In the second step} we choose the output weights $\bw_{out}$ such that \eqref{ple3eq5} holds. Here we simply choose
$w_1=1$ and $w_{s}=0$ for $s\in\{2,\dots,k_L\}$ and together with equation \eqref{ple3eq8} we obtain
\begin{align*}
	f_{i}(\bx)&=\max \Bigg\{
	\sum_{s''=1}^{k_L}
	w_{s''}\cdot
	o_{(i',j'),s''}^{(L)} \,
	: \,
	(i',j')\in\{2^{l-1}+l,\dots,\lambda-2^{l-1}-(l-1)\}^2
	\Bigg\}\\
	&=\max \Bigg\{
	o_{(i',j'),1}^{(L)} \,
	: \,
	(i',j')\in\{2^{l-1}+l,\dots,\lambda-2^{l-1}-(l-1)\}^2
	\Bigg\}\\
	&=\max \Bigg\{
	\bar{f}_{l,1}^{(i)}(\bx_{\big(\frac{i'-1/2}{\lambda}-\frac{1}{2},\frac{j'-1/2}{\lambda}-\frac{1}{2}\big)+I^{(l)}}) \,
	: \,
	(i',j')\in\{2^{l-1}+l,\dots,\lambda-2^{l-1}-(l-1)\}^2
	\Bigg\}\\
	&=\max_{\bu\in G_{\lambda}~:~\bu+I^{(l)}\subseteq G_{\lambda}}\bar{f}_{l,1}^{(i)}(\bx_{\bu+I^{(l)}}),
\end{align*}
where we used that
\begin{align*}
	&\left(\frac{i'-1/2}{\lambda}-\frac{1}{2},\frac{j'-1/2}{\lambda}-\frac{1}{2}\right)+I^{(l)}\\
	&
	=\left\{\frac{i'-2^{l-1}-l+1/2}{\lambda}-\frac{1}{2},\dots,\frac{i'+2^{l-1}+(l-1)+1/2}{\lambda}-\frac{1}{2}\right\}\\
	&\quad\times\left\{\frac{j'-2^{l-1}-l+1/2}{\lambda}-\frac{1}{2},\dots,\frac{j'+2^{l-1}+(l-1)+1/2}{\lambda}-\frac{1}{2}\right\}.
\end{align*}
\hfill $\Box$

\section{A bound on the covering number}
In this Section, we present the following upper bound for the covering number of our convolutional neural network architecture $\F_{\btheta}^{CNN}$ defined as in Section 3. 
\begin{lemma}
	\label{le10}
	Let $n,\lambda\in\N\setminus\{1\}$ and let $\sigma(x)=\max\{x,0\}$ be the ReLU activation function,
	define 
	\[\F\coloneqq\F_{\btheta}^{CNN}\] 
	with $\btheta=(t,L,\bk,\bM,B,L_{net},r_{net})$ as in Section \ref{se3}, and set
	\[
	k_{max}=\max\left\{k_1, \dots, k_{L},t,r_{net}\right\},
	\quad
	M_{max}=\max\{ M_1, \dots, M_{L} \}.\]
	Assume $\nconst\cdot \log n \geq 2$.
	Then we have for any
	$\epsilon \in (0,1)$:
	\begin{eqnarray*}
		&&
		\sup_{\bx_1^n \in {(\R^{G_{\lambda}})}^n} \log\left(
		\mathcal{N}_1 \left(\epsilon,T_{\const \cdot \log n}  \F, \bx_1^n\right) \right)
		\\
		&&
		\leq
		\nconst \cdot L^2 \cdot \log(L\cdot\lambda) \cdot
		\log \left(
		\frac{\mconst\cdot\log n}{\epsilon}
		\right)
	\end{eqnarray*}
	for some constant $\const >0$ which depends only on $L_{net}$, $k_{max}$ and $M_{max}$.
\end{lemma}
The proof of Lemma \ref{le10} is analogous to the proof of Lemma 7 in \cite{KoKrWa2020}. For the sake of completeness, we have adapted the proof below to the slight differences in network architecture (in \cite{KoKrWa2020} asymmetric zero padding is used in the convolutional layers and the output bound in (7) is applied one-sided).
With the aim of proving Lemma \ref{le10}, we first have to study the VC dimension of our function class $\F_{\btheta}^{CNN}$. For a class of subsets of $\R^d$, the VC dimension is defined as follows:
\begin{definition}
	Let $\A$ be a class of subsets of $\R^d$ with $\A\neq\emptyset$ and $m\in\mathbb N$.
	\begin{enumerate}
		\item For $\bx_1,...,\bx_m\in\mathbb R^d$ we define
		\[s(\mathcal A,\left\{\bx_1,...,\bx_m\right\})\coloneqq|\left\{A\cap\{\bx_1,...,\bx_m\}~:~A\in\mathcal A\right\}|.\]
		\item Then the $m$th \textbf{shatter coefficient} $S(\mathcal A,m)$ of $\mathcal A$ is defined by
		\[S(\mathcal A,m)\coloneqq\max_{\{\bx_1,...,\bx_m\}\subset\mathbb R^d}s(\mathcal A,\{\bx_1,...,\bx_m\}).\]
		\item The \textbf{VC dimension} (Vapnik-Chervonenkis-Dimension) $V_{\mathcal A}$ of $\mathcal A$ is defined as
		\[V_{\mathcal A}\coloneqq\sup\{m\in\mathbb N~:~S(\mathcal A,m)=2^m\}.\]
	\end{enumerate}
\end{definition}
For a class of real-valued functions, we define the VC dimension as follows:
\begin{definition}
	Let $\mathcal H$ denote a class of functions from $\R^d$ to $\{0,1\}$ and let $\F$ be a class of real-valued functions.
	\begin{enumerate}
		\item For any non-negative integer $m$, we define the \textbf{growth function} of $H$ as
		\[\Pi_{\mathcal H}(m)\coloneqq\max_{\bx_1,\dots,\bx_m\in\R^d}|\{(h(\bx_1),\dots,h(\bx_m)) : h\in H\}|.\]
		\item The \textbf{VC dimension} (Vapnik-Chervonenkis-Dimension) of $\mathcal H$ we define as
		\[\VC(\mathcal H)\coloneqq\sup\{m\in\N : \Pi_{\mathcal H}(m)=2^m\}.\]
		\item For $f\in\F$ we denote $\sgn(f)\coloneqq\IND_{\{f\geq0\}}$ and $\sgn(\F)\coloneqq\{\sgn(f) : f\in\F\}$. Then the \textbf{VC dimension} of $\F$ is defined as 
		\[\VC(\F)\coloneqq\VC(\sgn(\F)).\]
	\end{enumerate}
\end{definition}
A connection between both definitions is given by the following lemma.
\begin{lemma}
	\label{le11}
	Suppose $\F$ is a class of real-valued functions on $\R^d$.
	Furthermore, we define
	\[\F^+\coloneqq\{\{(\bx,y)\in\R^d\times\R : f(\bx)\geq y\} : f\in\F\}\]
	and define the class $\mathcal H$ of real-valued functions on $\R^d\times\R$ by
	\[\mathcal H\coloneqq\{h((\bx,y))=f(\bx)-y : f\in\F\}.\]
	Then, it holds that
	\[V_{\F^+}=\VC(\mathcal H).\]
\end{lemma}
\noindent
{\bf Proof.}
See Lemma 8 in \cite{KoKrWa2020}.
\hfill $\Box$
~\\
In order to bound the VC dimension of our function class, we need the following auxiliary result about the number of possible sign vectors attained by polynomials of bounded degree.
\begin{lemma}
	\label{le12}
	Suppose $W\leq m$ and let $f_1,...,f_m$ be polynomials of degree at most $D$ in $W$ variables. Define
	\[
	K\coloneqq|\{\left(\sgn(f_1(\ba)),\dots,\sgn(f_m(\ba))\right) : \ba\in\R^{W}\}|.
	\]
	Then we have 
	\[
	K\leq2\cdot\left(\frac{2\cdot e\cdot m\cdot D}{W}\right)^{W}.
	\]
\end{lemma}
\noindent
{\bf Proof.} See Theorem 8.3 in \cite{Anthony1999}.
\hfill $\Box$
~\\~\\
To get an upper bound for the VC dimension of our function class $\F_{\btheta}^{CNN}$ defined as in Section \ref{se3} we will use a modification of Theorem 6 in \cite{Bartlett2019}.
\begin{lemma}
	\label{le13}
	Let $\sigma(x)=\max\{x,0\}$ be the ReLU activation function,
	define 
	\[\F\coloneqq\F_{\btheta}^{CNN}\]
	with $\btheta=(t,L,\bk,\bM,B,L_{net},r_{net})$ as in Section \ref{se3}, and set
	\[
	k_{max}=\max\left\{k_1, \dots, k_{L},t,r_{net}\right\},
	\quad
	M_{max}=\max\{ M_1, \dots, M_{L} \}.\]
	Assume $\lambda>1$. Then, we have
	\[
	V_{\F^+}\leq\nconst\cdot L^2\cdot \log_2(L\cdot\lambda)
	\]
	for some constant $\const>0$ which depends only on
	$L_{net}$, $k_{max}$ and $M_{max}$.
\end{lemma}

\noindent
{\bf Proof.}
We want to use Lemma \ref{le11} to bound $\mathcal V_{\F^+}$ by $\VC(\mathcal H)$, where $\mathcal H$ is the class of real-valued functions on $[0,1]^{G_{\lambda}}\times\R$  defined by
\[\mathcal H\coloneqq\{h((\bx,y))=f(\bx)-y : f\in\F\}.\]
Let $h\in\mathcal H$. Then $h$ depends on $t$ convolutional neural networks 
\begin{equation*}
	f_1,\dots,f_t\in\F^{CNN}(L,\bk,\bM,B)
\end{equation*}
and one standard feedforward neural network $g_{net}\in\G_t(L_{net},r_{net})$ such that
\[h((\bx,y))=g_{net}\circ (f_1,\dots,f_t)(\bx)-y\]
Each one of the convolutional neural networks $f_1,\dots,f_t$ depends on a weight matrix
\[
\bw^{(b)}
=
\left(
w_{i,j,s_1,s_2}^{(b,r)}
\right)_{
	1 \leq i,j \leq M_r, s_1 \in \{1, \dots, k_{r-1}\}, s_2 \in \{1, \dots, k_r\},
	r \in \{1, \dots,L \}
},
\]
the weights
\[
\bw_{bias}^{(b)}
=
\left(
w_{s_2}^{(b,r)}
\right)_{
	s_2 \in \{1, \dots, k_r\},
	r \in \{1, \dots,L\}
}
\]
for the bias in each channel and each convolutional layer,
the output weights
\[
\bw_{out}^{(b)}=(w_{s}^{(b)})_{
	s \in \{1, \dots, k_{L}\}
}
\]
for $b\in\{1,\dots,t\}$.
The standard feedforward neural network $g_{net}$ depends on the inner weigths
\[w_{i,j}^{(r-1)}\]
for $r\in\{2,\dots,L_{net}\}$, $j\in\{0,\dots,r_{net}\}$ and $i\in\{1,\dots,r_{net}\}$ and 
\[w_{i,j}^{(0)}\]
for $j\in\{0,\dots,t\}$, $i\in\{1,\dots,r_{net}\}$
and the outer weights
\[w_i^{(L_{net})}\]
for $i\in\{0,\dots,k_{L_{net}}\}$.

We set 
\[(k_0,\dots,k_{L+L_{net}+1})=(1,k_1,\dots,k_L,t,r_{net},\dots,r_{net})\]
and count the number of weights used up to layer $r\in\{1,\dots,L\}$ in the convolutional part by
\[W_r\coloneqq t\cdot\left(\sum_{s=1}^{r}M_s^2\cdot k_s\cdot k_{s-1}+\sum_{s=1}^{r}k_s\right),\]
for $r\in\{1,\dots,L\}$
(where we set $W_0\coloneqq0$) and 
\begin{equation*}
	W_{L+1}\coloneqq W_{L}+t\cdot k_{L}.
\end{equation*}
We continue in the part of the standard feedforward neural network by counting the weights used up to layer $r\in\{1,\dots,L_{net}\}$ by
\[W_{L+1+r}=W_{L+r}+\left(k_{L+r}+1\right)\cdot k_{L+r+1}\]
and denote the total number of weights by
\begin{equation}
	\begin{split}
		W&=W_{L+L_{net}+2}\\
		&=W_{L+L_{net}+1}+k_{L+L_{net}+1}+1\\
		&\leq L\cdot t\cdot\Big( M_{max}^2\cdot k_{max}^2+k_{max}\Big)+t\cdot k_{max}\\
		&\quad+L_{net}\cdot((k_{max}+1)\cdot k_{max})+k_{max}+1\\
		&\leq L\cdot t\cdot\Big( M_{max}^2\cdot(k_{max}+1)\cdot k_{max}\Big)\\
		&\quad+L_{net}\cdot((k_{max}+1)\cdot k_{max})\\
		&\quad+2\cdot t\cdot(k_{max}+1)\\
		&\leq(L+L_{net}+2)\cdot t\cdot M_{max}^2\cdot(k_{max}+1)\cdot k_{max}\\
		&\leq2\cdot(L+L_{net}+2)\cdot t\cdot M_{max}^2\cdot k_{max}^2.
	\end{split}
	\label{eqW}
\end{equation}
We define $I^{(0)}=\emptyset$ and for $r\in\{1,\dots,L+L_{net}+2\}$ we define the index sets
\[I^{(r)}=\{1,\dots,W_{r}\}.\]
Furthermore, we define a sequence of vectors containing the weights used up to layer $r\in\{1,\dots,L\}$ in the convolutional part by
\begin{align*}
	&\ba_{I^{(r)}}\coloneqq\Big(\ba_{I^{(r-1)}},w_{1,1,1,1}^{(1,r)},\dots,w_{M_r,M_r,k_{r-1},k_{r}}^{(1,r)},w_{1}^{(1,r)},\dots,w_{k_r}^{(1,r)},\\
	&\hspace{3cm}\dots,w_{1,1,1,1}^{(t,r)},\dots,w_{M_r,M_r,k_{r-1},k_{r}}^{(t,r)},w_{1}^{(t,r)},\dots,w_{k_r}^{(t,r)}
	\Big)\in\R^{W_r}
\end{align*}
(where $\ba_{\emptyset}$ denotes the empty vector),
\[\ba_{I^{(L+1)}}\coloneqq(\ba_{I^{(L)}},w_{1}^{(1)},\dots,w_{k_{L}}^{(1)},\dots,w_{1}^{(t)},\dots,w_{k_{L}}^{(t)})\in\R^{W_{L+1}},\]
and by continuing with the part of the standard feedforward neural network we get for $r\in\{1,\dots,L_{net}\}$
\[\ba_{I^{(r+L+1)}}\coloneqq\left(\ba_{I^{(r+L)}},w_{1,0}^{(r-1)},\dots,w_{k_{r+L+1},k_{r+L}}^{(r-1)}\right)\in\R^{W_{r+L+1}}\]
and
\[\ba\coloneqq\left(\ba_{I^{(L+L_{net}+1)}},w_0^{(L_{net})},\dots,w_{{L_{net}}}^{(L_{net})}\right)\in\R^W.\]
With this notation we can write
\[\mathcal H=\{(\bx,y)\mapsto h((\bx,y),\ba) : \ba\in\R^W\}\]
and for $b\in\{1,\dots,t\}$
\[\F^{CNN}(L,\bk,\bM,B)=\{\bx\mapsto f_b(\bx,\ba) : \ba\in\R^W\},\]
where the convolutional networks $f_1,\dots,f_t\in\F^{CNN}(L,\bk,\bM,B)$, as described above, each depends only on $W_{L+1}/t$ variables of $\ba$.
To get an upper bound for the VC-dimension of $\mathcal H$, we will bound the growth function $\Pi_{\sgn(\mathcal H)}(m)$. 
In the following we 
consider first the case where %
\begin{equation}
	m\geq W
	\label{eqV1}
\end{equation}
since this will allow us several uses of Lemma \ref{le12}.
To bound the growth function $\Pi_{\sgn(\mathcal H)}(m)$, we fix the input values \[(\bx_1,y_1),\dots,(\bx_m,y_m)\in [0,1]^{G_{\lambda}}\times\R\] 
and consider $h\in\mathcal H$ as a function of the weight vector $\ba\in\R^{W}$ of $h$ 
\[\ba\mapsto h((\bx_k,y_k),\ba)=g\circ(f_{1},\dots,f_{t})(\bx_k,\ba)-y_k=h_k(\ba)\]
for any $k\in\{1,\dots,m\}$.
Then, an upper bound for
\[K\coloneqq|\{(\sgn(h_1(\ba)),\dots,\sgn(h_m(\ba))) : \ba\in\R^{W}\}|\]
implies an upper bound for the growth function $\Pi_{\sgn(\mathcal H)}(m)$.
For any partition  
\[\mathcal S=\{S_1,\dots,S_M\}\] of $\R^W$ it holds that
\begin{eqnarray}
	K\leq\sum_{i=1}^{M}|\{(\sgn(h_1(\ba)),\dots,\sgn(h_m(\ba)) : \ba\in S_i\}|.
	\label{sum}
\end{eqnarray}
We will construct a partition $\mathcal S$ of $\R^W$ such that within each region $S\in\mathcal S$ , the functions $h_k(\cdot)$		
are all fixed polynomials of bounded degree for $k\in\{1,\dots,m\}$,
so that each summand of equation \eqref{sum} can be bounded via Lemma \ref{le12}. We do this in two steps.

\textit{In the first step} we construct a partition $\mathcal S^{(1)}$ of $\R^W$ such that within each $S\in\mathcal S^{(1)}$ the $t$ convolutional neural networks $f_{1,k}\left(\ba\right),\dots,f_{t,k}\left(\ba\right)$ are all fixed polynomials with dergee of at most $L+1$ for all $k\in\{1,\dots,m\}$, where we denote  
\[f_{b,k}\left(\ba\right)=f_{b}\left(\bx_k,\ba\right)\]
for $b\in\{1,\dots,t\}$. 
For $b\in\{1,\dots,t\}$ we have
\begin{align*}
	&f_{b,k}\left(\ba\right)=\max\Bigg\{\sum_{s=1}^{k_{L}}w_s^{(b)}\cdot o^{\left(L\right)}_{(i,j),b,s,\bx_k}(\ba_{I^{(L)}}) : (i,j)\in\{1+B,\dots,\lambda-B\}^2\Bigg\},
\end{align*}
where $o_{(i,j),b,s_2,\bx}^{(L)}(\ba_{I^{(L)}})$ is
recursively defined by
\begin{align*}
	&o_{(i,j),b,s_2,\bx}^{(r)}(\ba_{I^{(r)}})\\
	&=
	\sigma \left(
	\sum_{s_1=1}^{k_{r-1}}
	\sum_{\substack{t_1,t_2 \in \{1, \dots, M_r\}\\i+t_1-\lceil M_r/2\rceil\in\{1,\dots,\lambda\}\\j+t_2-\lceil M_r/2\rceil\in\{1,\dots,\lambda\}}}
	w_{t_1,t_2,s_1,s_2}^{(b,r)}
	\cdot
	o_{(i+t_1-\lceil M_r/2\rceil,j+t_2-\lceil M_r/2\rceil),b,s_1,\bx}^{(r-1)}(\ba_{I^{(r-1)}})
	+
	w_{s_2}^{(b,r)}
	\right)
\end{align*}
for $(i,j)\in\{1,\dots,\lambda\}^2$
and
$r \in \{1, \dots, L\}$, and by
\[
o_{(i,j),b,1,\bx }^{(0)}(\ba_{I^{(0)}}) = x_{\left(\frac{i-1/2}{\lambda}-\frac{1}{2},\frac{j-1/2}{\lambda}-\frac{1}{2}\right)}
\quad \mbox{for }
(i,j)\in \{1, \dots, \lambda\}^2.
\]
Firstly, we construct a partition $\mathcal S_{L}=\{S_1,\dots,S_M\}$ of $\R^{W}$ such that within each $S\in\mathcal S_{L}$ 				
\[o_{(i,j),b,s,\bx_k}^{(L)}(\ba_{I^{(L)}})\]
is a fixed polynomial for all $k\in\{1,\dots,m\}$, $s\in\{1,\dots, k_L\}$, $b\in\{1,\dots,t\}$ and $(i,j)\in D$ with degree of at most $L$ in the $W_{L}$ variables $\ba_{I^{(L)}}$ of $\ba\in S$.
We construct the partition $\mathcal S_{L}$ iteratively layer by layer, by creating a sequence $\mathcal S_0,\dots,\mathcal S_{L}$, where each $\mathcal S_r$ is a partition of $\R^{W}$ with the following properties:
\begin{enumerate}
	\item We have $|\mathcal S_0|=1$ and, for each $r\in\{1,\dots,L\}$,
	\begin{equation}
		\frac{|\mathcal S_r|}{|\mathcal S_{r-1}|}\leq2\left(\frac{2\cdot e\cdot t\cdot k_r\cdot\lambda^2\cdot m\cdot r}{W_r}\right)^{W_r},
		\label{prob2}
	\end{equation}
	\item For  each $r\in\{0,\dots,L\}$, and each element $S\in\mathcal S_{r}$, each $(i,j) \in\{1,\dots,\lambda\}^2$, each $s\in\{1,\dots,k_r\}$, each $k\in\{1,\dots,m\}$, and each $b\in\{1,\dots,t\}$ when $\ba$ varies in $S$,
	\[o^{(r)}_{(i,j),b,s,\bx_k}(\ba_{I^{(r)}})\]
	is a fixed polynomial function in the $W_r$ variables $\ba_{I^{(r)}}$ of $\ba$, of total degree no more than $r$.
\end{enumerate}
We define $\mathcal S_0\coloneqq\{\R^{W}\}$. Since 
\[o^{(0)}_{(i,j),b,s,\bx_k}(\ba_{I^{(0)}})=(x_k)_{\left(\frac{i-1/2}{\lambda}-\frac{1}{2},\frac{j-1/2}{\lambda}-\frac{1}{2}\right)}\]
is a constant polynomial, property 2 above is satisfied for $r=0$. 
Now suppose that $\mathcal S_0,\dots,\mathcal S_{r-1}$ have been defined, and we want to define $\mathcal S_{r}$. For $S\in\mathcal S_{r-1}$ let 
\[p_{(i,j),b,s_1,\bx_k,S}(\ba_{I^{(r-1)}})\]
denote the function $o_{(i,j),b,s_1,\bx_k}^{(r-1)}(\ba_{I^{(r-1)}})$, when $\ba\in S$. By induction hypothesis 
\[p_{(i,j),b,s_1,\bx_k,S}(\ba_{I^{(r-1)}})\]
is a polynomial with total degree no more than $r-1$, and depends on the $W_{r-1}$ variables $\ba_{I^{(r-1)}}$ of $\ba$ for any $b\in\{1,\dots,t\}$, $k\in\{1,\dots,m\}$, $(i,j)\in\{1,\dots,\lambda\}^2$ and $s_1\in\{1,\dots,k_{r-1}\}$.
Hence for any $b\in\{1,\dots,t\}$ $k\in\{1,\dots,m\}$, $(i,j)\in\{1,\dots,\lambda\}^2$ and $s_2\in\{1,\dots,k_{r}\}$
\[\sum_{s_1=1}^{k_{r-1}}
\sum_{\substack{t_1,t_2 \in \{1, \dots, M_r\}\\i+t_1-\lceil M_r/2\rceil\in\{1,\dots,\lambda\}\\j+t_2-\lceil M_r/2\rceil\in\{1,\dots,\lambda\}}}
w_{t_1,t_2,s_1,s_2}^{(b,r)}
\cdot
p_{(i+t_1-\lceil M_r/2\rceil,j+t_2-\lceil M_r/2\rceil),b,s_1,\bx_k,S}(\ba_{I^{(r-1)}})
+
w_{s_2}^{(b,r)}\] 
is a polynomial in the $W_r$ variables $\ba_{I^{(r)}}$ of $\ba$ with total degree no more than $r$.
Because of condition \eqref{eqV1} we have $t\cdot k_r\cdot m\cdot \lambda^2\geq W_r$.
Hence, by Lemma \ref{le12}, the collection of polynomials 
\begin{equation}
	\label{ple13eq1}
	\begin{split}
			&\left\{\sum_{s_1=1}^{k_{r-1}}
		\sum_{\substack{t_1,t_2 \in \{1, \dots, M_r\}\\i+t_1-\lceil M_r/2\rceil\in\{1,\dots,\lambda\}\\j+t_2-\lceil M_r/2\rceil\in\{1,\dots,\lambda\}}}
		w_{t_1,t_2,s_1,s_2}^{(b,r)}
		\cdot
		p_{(i+t_1-\lceil M_r/2\rceil,j+t_2-\lceil M_r/2\rceil),b,s_1,\bx_k,S}(\ba_{I^{(r-1)}})
		+
		w_{s_2}^{(b,r)} :\right.
		\\
		&\left.\hspace{3cm}b\in\{1,\dots,t\}, k\in\{1,\dots,m\}, (i,j)\in\{1,\dots,\lambda\}^2,s_2\in\{1,\dots,k_r\}\vphantom{\sum_{\substack{t_1,t_2 \in \{1, \dots, M_r\}\\(i+t_1-1,j+t_2-1)\in D}}}\right\}
	\end{split}
\end{equation}	
attains at most
\[\Pi\coloneqq2\left(\frac{2\cdot e\cdot t\cdot k_{r}\cdot m\cdot\lambda^2\cdot r}{W_r}\right)^{W_r}\]
distinct sign patterns when $\ba\in S$. %
Therefore, we can partition $S\subset\R^W$ into $\Pi$ subregions, such that all the polynomials don't change their signs within each subregion. Doing this for all regions $S\in\mathcal S_{r-1}$ we get our required partition $\mathcal S_r$ by assembling all of these subregions. In particular, property 1 (inequality \eqref{prob2}) is then satisfied.

Fix some $S'\in\mathcal S_{r}$. Notice that, when $\ba$ varies in $S'$, all the polynomials in \eqref{ple13eq1}
don't change their signs, hence 
\begin{align*}
	&o_{(i,j),b,s_2,\bx_k}^{(r)}(\ba_{I^{(r)}})\\
	&=
	\sigma \left(
	\sum_{s_1=1}^{k_{r-1}}
	\sum_{\substack{t_1,t_2 \in \{1, \dots, M_r\}\\i+t_1-\lceil M_r/2\rceil\in\{1,\dots,\lambda\}\\j+t_2-\lceil M_r/2\rceil\in\{1,\dots,\lambda\}}}
	w_{t_1,t_2,s_1,s_2}^{(b,r)}
	\cdot
	o_{(i+t_1-\lceil M_r/2\rceil,j+t_2-\lceil M_r/2\rceil),b,s_1,\bx}^{(r-1)}(\ba_{I^{(r-1)}})
	+
	w_{s_2}^{(b,r)}
	\right)\\
	&=\max \Bigg\{
	\sum_{s_1=1}^{k_{r-1}}
	\sum_{\substack{t_1,t_2 \in \{1, \dots, M_r\}\\i+t_1-\lceil M_r/2\rceil\in\{1,\dots,\lambda\}\\j+t_2-\lceil M_r/2\rceil\in\{1,\dots,\lambda\}}}
	w_{t_1,t_2,s_1,s_2}^{(b,r)}
	\cdot
	o_{(i+t_1-\lceil M_r/2\rceil,j+t_2-\lceil M_r/2\rceil),b,s_1,\bx}^{(r-1)}(\ba_{I^{(r-1)}})\\
	&\hspace{2cm}
	+
	w_{s_2}^{(b,r)}
	,0\Bigg\}
\end{align*}
is either a polynomial of degree no more than $r$ in the $W_r$ variables $\ba_{I^{(r)}}$ of $\ba$ or a constant polynomial with value $0$ for all $(i,j)\in\{1,\dots,\lambda\}^2$, $b\in\{1,\dots,t\}$, $s_2\in\{1,\dots,k_r\}$ and $k\in\{1,\dots,m\}$. Hence, property 2 is also satisfied and we are able to construct our desired partition $\mathcal S_{L}$. Because of inequality \eqref{prob2} of property 1 it holds that 
\[|\mathcal S_{L}|\leq\prod_{r=1}^{L}2\left(\frac{2\cdot e\cdot t \cdot k_r\cdot\lambda^2\cdot m\cdot r}{W_r}\right)^{W_r}.\]
For any $(i,j)\in\{1,\dots,\lambda\}^2$, $b\in\{1,\dots,t\}$ and $k\in\{1,\dots,m\}$, we define
\[f_{(i,j),b,\bx_k}(\ba_{I^{(L+1)}})\coloneqq\sum_{s_2=1}^{k_{L}}
w_{s_2}^{(b)} \cdot o_{(i,j),b,s_2,\bx_k}^{(L)}(\ba_{I^{(L)}}).\]	
For any fixed $S\in\mathcal S_{L}$, let $p_{(i,j),b,S,\bx_k}(\ba_{I^{(L+1)}})$ denote the function $f_{(i,j),b,\bx_k}(\ba_{I^{(L+1)}})$, when $\ba\in S$. By construction of $\mathcal S_{L}$ this is a polynomial of degree no more than $L+1$ in the $W_{L+1}$ variables $\ba_{I^{(L+1)}}$ of $\ba$. 
Because of condition \eqref{eqV1} we have $t\cdot\lambda^4\cdot m\geq W_{L+1}$.
Hence, by Lemma \ref{le12}, the collection of polynomials
\begin{align*}
	&\Big\{p_{(i_1,j_1),b,S,\bx_k}(\ba_{I^{(L+1)}})-p_{(i_2,j_2),b,S,\bx_k}(\ba_{I^{(L+1)}}) : \\
	&\quad(i_1,j_1),(i_2,j_2)\in\{1,\dots,\lambda\}^2, (i_1,j_1)\neq(i_2,j_2), b\in\{1,\dots,t\}, k\in\{1,\dots,m\}
	\Big\}
\end{align*}
attains at most
\[\Delta\coloneqq2\left(\frac{2\cdot e\cdot t\cdot\lambda^4\cdot m\cdot (L+1)}{W_{L+1}}\right)^{W_{L+1}}\]
distinct sign patterns when $\ba\in S$. %
Therefore, we can partition $S\subset\R^W$ into $\Delta$ subregions, such that all the polynomials don't change their signs within each subregion. Doing this for all regions $S\in\mathcal S_{L}$ we get our required partition $\mathcal S^{(1)}$ by assembling all of these subregions. For the size of our partition $\mathcal S^{(1)}$ we get
\[
|\mathcal S^{(1)}|\leq\prod_{r=1}^{L}2\cdot\left(\frac{2\cdot t\cdot e\cdot k_r\cdot\lambda^2\cdot m\cdot r}{W_r}\right)^{W_r}\cdot2\cdot\left(\frac{2\cdot e\cdot t\cdot\lambda^4\cdot m\cdot(L+1)}{W_{L+1}}\right)^{W_{L+1}}.
\]
Fix some $S'\in\mathcal S^{(1)}$. Notice that, when $\ba$ varies in $S'$, all the polynomials 
\begin{align*}
	&\Big\{p_{(i_1,j_1),b,S,\bx_k}(\ba_{I^{(L+1)}})-p_{(i_2,j_2),b,S,\bx_k}(\ba_{I^{(L+1)}}) :\\
	&(i_1,j_1),(i_2,j_2)\in\{1,\dots,\lambda\}^2, (i_1,j_1)\neq(i_2,j_2), b\in\{1,\dots,t\}, k\in\{1,\dots,m\}
	\Big\}
\end{align*}
don't change their signs. Hence, there is a permutation $\pi^{(b,k)}$ of the set 
\[
\{1+B,\dots,\lambda-B\}^2
\]
for any $b\in\{1,\dots,t\}$ and $k\in\{1,\dots,m\}$ such that
\[f_{\pi^{(b,k)}((1+B,1+B)),b,\bx_k}(\ba_{I^{(L+1)}})\geq\dots\geq f_{\pi^{(b,k)}((\lambda-B,\lambda-B)),b,\bx_k}(\ba_{I^{(L+1)}})\]
for $\ba\in S'$ and any $k\in\{1,\dots,m\}$ and $b\in\{1,\dots,t\}$. Therefore, it holds that
\begin{align*}
	f_{b,k}(\ba)&=\max\left\{f_{(1+B,1+B),b,\bx_k}\left(\ba_{I^{(L+1)}}\right),\dots,f_{\left(\lambda-B,\lambda-B\right),b,\bx_k}\left(\ba_{I^{(L+1)}}\right)\right\}\\
	&=f_{\pi^{(b,k)}((1+B,1+B)),b,\bx_k}(\ba_{I^{(L+1)}}),
\end{align*}
for $\ba\in S'$. Since $f_{\pi^{(b,k)}((1+B,1+B)),b,\bx_k}(\ba_{I^{(L+1)}})$ is a polynomial within $S'$, also $f_{b,k}(\ba)$ is a polynomial within $S'$ with degree no more than $L+1$ and in the $W_{L+1}$ variables $\ba_{I^{(L+1)}}$ of $\ba\in\R^W$.

\textit{In the second step} we construct the partition $\mathcal S$ starting from partition $\mathcal S^{(1)}$ such that within each region $S\in\mathcal S$ the functions $h_k(\cdot)$ are all fixed polynomials of degree of at most $L+L_{net}+2$ for $k\in\{1,\dots,m\}$. We have 
\[h_k(\ba)=\sum_{i=1}^{k_{L+L_{net}+1}}w_i^{(L_{net})}\cdot g_{i,k}^{(L_{net})}\left(\ba_{I^{(L+L_{net}+1)}}\right)+w_0^{(L_{net})}-y_k\]
where the $g_{i,k}^{(L_{net})}$ are recursively defined by
\[g_{i,k}^{(r)}\left(\ba_{I^{(L+r+1)}}\right)=\sigma\left(\sum_{j=1}^{k_{L+r}}w_{i,j}^{(r-1)}g_{j,k}^{(r-1)}(\ba_{I^{(L+r)}})\right)\]
for $r\in\{1,\dots,L_{net}\}$ and 
\[g_{i,k}^{(0)}(\ba_{I^{(L+1)}})=f_{i,k}(\ba)\]
for $i\in\{1,\dots,k_{L+1}\}$ ($k_{L+1}=t$).
As above we construct the partition $\mathcal S$ iteratively layer by layer, by creating a sequence $\mathcal S_0,\dots,\mathcal S_{L_{net}}$, where each $\mathcal S_r$ is a partition of $\R^{W}$ with the following porperties:
\begin{enumerate}
	\item We set $\mathcal S_0=\mathcal S^{(1)}$ and, for each $r\in\{1,\dots,L_{net}\}$,
	\begin{equation}
		\frac{|\mathcal S_r|}{|\mathcal S_{r-1}|}\leq2\left(\frac{2\cdot e\cdot k_{L+r+1}\cdot m\cdot (L+r+1)}{W_{L+r+1}}\right)^{W_{L+r+1}},
		\label{prob1}
	\end{equation}
	\item For  each $r\in\{0,\dots,L_{net}\}$, and each element $S\in\mathcal S_{r}$, each $i \in \{1,\dots,k_{L+r+1}\}$, and each $k\in\{1,\dots,m\}$ when $\ba$ varies in $S$,
	\[g_{i,k}^{(r)}(\ba_{I^{(L+r+1)}})\]
	is a fixed polynomial function in the $W_{L+r+1}$ variables $\ba_{I^{(L+r+1)}}$ of $\ba$, of total degree no more than $L+r+1$.
\end{enumerate}		
As we have already shown in step 1, property 2 above is satisfied for $r=0$. Now suppose that $\mathcal S_0,\dots,\mathcal S_{r-1}$ have been defined, and we want to define $\mathcal S_r$. For $S\in\mathcal S_{r-1}$ and $j \in \{1,\dots,k_{L+r}\}$ let $p_{j,k,S}(\ba_{I^{(L+r)}})$ denote the function $g_{j,k}^{(r-1)}(\ba_{I^{(L+r)}})$, when $\ba\in S$. By induction hypothesis $p_{j,k,S}(\ba_{I^{(L+r)}})$ is a polynomial with total degree no more than $L+r$, and depends on the $W_{L+r}$ variables $\ba_{I^{(L+r)}}$ of $\ba$. Hence for any $k\in\{1,\dots,m\}$ and $i\in\{1,\dots,k_{L+r+1}\}$ 
\begin{align*}
	\sum_{j=1}^{k_{L+r}}w_{(i,j)}^{(r-1)}\cdot p_{j,k,S}(\ba_{I^{(L+r)}})+w_{i,0}^{(r-1)}
\end{align*}
is a polynomial in the $W_{L+r+1}$ variables $\ba_{I^{(L+r+1)}}$ variables of $\ba$ with total degree no more than $L+r+1$. Because of condition \eqref{eqV1} we have $k_{L+r+1}\cdot m\geq W_{L+r+1}$. Hence, by Lemma \ref{le12}, the collection of polynomials
\begin{align*}
	\left\{\sum_{j=1}^{k_{L+r}}w_{(i,j)}^{(r-1)}\cdot p_{j,k,S}(\ba_{I^{(L+r)}})+w_{i,0}^{(r-1)} : k\in\{1,\dots,m\}, i\in\{1,\dots,k_{L+r+1}\}\right\}
\end{align*}
attains at most
\[\Pi\coloneqq 2\left(\frac{2\cdot e\cdot k_{L+r+1}\cdot m\cdot (L+r+1)}{W_{L+r+1}}\right)^{W_{L+r+1}}\]
distinct sign patterns when $\ba\in S$.
Therefore, we can partition $S\subset\R^W$ into $\Pi$ subregions, such that all the polynomials don't change their signs within each subregion. Doing this for all regions $S\in\mathcal S_{r-1}$ we get our required partition $\mathcal S_r$ by assembling all of these subregions. In particular property 1 is then satisfied. In order to see that condition 2 is also satisfied, we can proceed analogously to step 1. Hence, when $\ba$ varies in $S\in\mathcal S$ the function
\[h_k(\ba)=\sum_{i=1}^{k_{L+L_{net}+1}}w_i^{(L)}\cdot g_{i,k}^{(L_{net})}\left(\ba_{I^{(L+L_{net}+1)}}\right)+w_0^{(L)}-y_k\]
is a polynomial of degree no more than $L+L_{net}+2$ in the $W$ variables of $\ba\in\R^W$ for any $k\in\{1,\dots,m\}$.		
For the size of our partition $\mathcal S$ we get
\begin{align*}
	|\mathcal S|&\leq\prod_{r=1}^{L}2\cdot\left(\frac{2\cdot e\cdot t\cdot k_r\cdot\lambda^2\cdot m\cdot r}{W_r}\right)^{W_r}\cdot2\cdot\left(\frac{2\cdot e\cdot\lambda^4\cdot m\cdot(L+1)}{W_{L+1}}\right)^{W_{L+1}}\\
	&\hspace{0.5cm}\cdot\prod_{r=1}^{L_{net}}2\cdot\left(\frac{2\cdot e\cdot k_{L+r+1}\cdot m\cdot(L+r+1)}{W_{L+r+1}}\right)^{W_{L+r+1}}\\
	&\leq\prod_{r=1}^{L+L_{net}+1}2\cdot\left(\frac{2\cdot e\cdot t\cdot k_r\cdot\lambda^4\cdot m\cdot r}{W_r}\right)^{W_r}
\end{align*}
By condition \eqref{eqV1} and another application of Lemma \ref{le12} it holds for any $S'\in\mathcal S$ that
\begin{align*}
	&|\{(\sgn(h_1(\ba)),\dots,\sgn(h_m(\ba))) : \ba\in S'\}|\\
	&\leq2\cdot\left(\frac{2\cdot e\cdot m\cdot (L+L_{net}+2)}{W}\right)^{W}.
\end{align*}
Now we are able to bound $K$ via equation \eqref{sum} and because $K$ is an upper bound for the growth function we set $k_{L+L_{net}+2}=1$ and get
\begin{align}
	\Pi_{\sgn(\mathcal H)}(m)&\leq\prod_{r=1}^{L+L_{net}+2}2\cdot\left(\frac{2\cdot e\cdot t\cdot k_r\cdot\lambda^4\cdot r\cdot m}{W_r}\right)^{W_r}\nonumber\\
	&{\leq}2^{L+L_{net}+2}\cdot\left(\frac{\sum_{r=1}^{L+L_{net}+2}2\cdot e\cdot t\cdot k_r\cdot\lambda^4\cdot r\cdot m}{\sum_{r=1}^{L+L_{net}+2} W_r}\right)^{\sum_{r=1}^{L+L_{net}+2} W_r}\nonumber\\
	&=2^{L+L_{net}+2}\cdot\left(\frac{R\cdot m}{\sum_{r=1}^{L+L_{net}+2} W_r}\right)^{\sum_{r=1}^{L+L_{net}+2} W_r},\label{In1}
\end{align}
with $R\coloneqq2\cdot e\cdot t\cdot\lambda^4\cdot \sum_{r=1}^{L+L_{net}+2}k_r\cdot r$. In the second row we used the weighted AM-GM inequality (see,
e.g., \cite{Cvetkovski2012}, pp. 74-75). 
Without loss of generality, we can assume that 
$\VC(\mathcal H)\geq\sum_{r=1}^{L+L_{net}+2}W_r$
because in the case $\VC(\mathcal H)<\sum_{r=1}^{L+L_{net}+2}W_r$ we have
\begin{align*}
	\VC(\mathcal H)&~<(L+L_{net}+2)\cdot W\\
	&\stackrel{\eqref{eqW}}{\leq}2\cdot (L+L_{net}+2)^2\cdot t\cdot M_{max}^2\cdot k_{max}^2 \\
	&~\leq\const\cdot L^2
\end{align*}
for some constant $\const>0$ which only depends on $L_{net}$, $M_{max}$ and $k_{max}$ and get the assertion by Lemma \ref{le11}.
Hence we get by the definition of the VC--dimension and inequality \eqref{In1} (which only holds for $m\geq W$)
\[2^{\VC(\mathcal H)}=\Pi_{\sgn(\mathcal H)}(\VC(\mathcal H))\leq2^{L+L_{net}+2}\cdot\left(\frac{R\cdot\VC(\mathcal H)}{\sum_{r=1}^{L+L_{net}+2} W_r}\right)^{\sum_{r=1}^{L+L_{net}+2} W_r}.\]
Since 
\[R\geq2\cdot e\cdot t\cdot\lambda^4\cdot\sum_{r=1}^{1+1+2}r\geq2\cdot e\cdot t\cdot\lambda^4\cdot 10\geq16\]
Lemma \ref{le14} below (with parameters $R$, $m=\VC(\mathcal H)$, $w=\sum_{r=1}^{L+L_{net}+2}W_r$ and $L'=L+L_{net}+2$) implies that
\begin{align*}
	\VC(\mathcal H)&\leq(L+L_{net}+2)+\left(\sum_{r=1}^{L+L_{net}+2} W_r\right)\cdot\log_2(2\cdot R\cdot\log_2(R))\\
	&\leq(L+L_{net}+2)+(L+L_{net}+2)\cdot W\\
	&\hspace{1cm}\cdot \log_2(2\cdot(2\cdot e\cdot t\cdot\lambda^4\cdot (L+L_{net}+2)\cdot k_{max})^2)\\
	&\leq2\cdot(L+L_{net}+2)\cdot W\cdot \log_2\left(\left(2\cdot e\cdot t\cdot(L+L_{net}+2)\cdot k_{max}\cdot\lambda\right)^8\right)\\
	&\stackrel{\eqref{eqW}}{\leq}32\cdot t\cdot(L+L_{net}+2)^2\cdot k_{\text{max}}^2\cdot M_{\text{max}}^2\\
	&\hspace{1cm}\cdot\log_2\left(2\cdot e\cdot t\cdot(L+L_{net}+2)\cdot k_{max}\cdot\lambda\right)\\
	&\leq\const\cdot L^2\cdot \log_2(L\cdot\lambda),
\end{align*}
for some constant $\const>0$ which only depends on $L_{net}$, $k_{\text{max}}$ and $M_{\text{max}}$. In the third row we used equation \eqref{eqW} for the total number of weights $W$. Now we make use of Lemma \ref{le11} and finally get
\[V_{\F^+}\leq \const\cdot L^2\cdot \log_2(L\cdot\lambda).\]	
\hfill $\Box$
\begin{lemma}
	\label{le14}
	Suppose that $2^m\leq2^{L'}\cdot(m\cdot R/w)^w$ for some $R\geq16$ and $m\geq w\geq L'\geq0$. Then,
	\[
	m\leq L'+w\cdot\log_2(2\cdot R\cdot\log_2(R)).
	\]
\end{lemma}
\noindent
{\bf Proof.}
See Lemma 16 in \cite{Bartlett2019}.
\hfill $\Box$
~\\~\\
\noindent
{\bf Proof of Lemma 10.} 
Using Lemma \ref{le13} and
\[
V_{T_{c_{4} \cdot \log n} \F^+}
\leq
V_{\F^+},
\]
we can conclude from this together with Lemma 9.2 and Theorem 9.4
in \cite{Gyoerfi2002}
\begin{eqnarray*}
	&&
	\mathcal{N}_1 \left(\epsilon,   T_{\mmconst{2} \cdot \log n} \F,
	\bx_1^n\right)
	\\
	&&
	\leq
	3 \cdot \left(
	\frac{4 e \cdot\mmconst{2}\cdot \log n}{\epsilon}
	\cdot
	\log
	\frac{6 e \cdot\mmconst{2} \cdot \log n}{\epsilon}
	\right)^{V_{T_{\mmconst{2}\cdot \log n} \F^+}}
	\\
	&&
	\leq
	3 \cdot \left(
	\frac{6 e \cdot\mmconst{2}\cdot \log n}{\epsilon}
	\right)^{
		2 \cdot
		\mconst \cdot L^2 \cdot \log (L\cdot \lambda^2)
	}
	.
\end{eqnarray*}
This completes the proof of Lemma 10.
\hfill $\Box$
\end{document}